\pdfoutput=1

\documentclass[11pt]{article}

\usepackage[preprint]{acl}

\usepackage{times}
\usepackage{latexsym}

\usepackage[T1]{fontenc}

\usepackage[utf8]{inputenc}

\usepackage{microtype}

\usepackage{inconsolata}

\usepackage{graphicx}
\usepackage{enumitem}
\usepackage{subcaption}
\usepackage{booktabs}
\usepackage{caption}
\usepackage{amsmath}

\usepackage{comment}
\usepackage{amsfonts}

\usepackage{xcolor,colortbl}
\usepackage{multirow}
\usepackage{tikz}

\definecolor{lightred}{RGB}{255, 204, 204}
\definecolor{lightgreen}{RGB}{204, 255, 204}

\newcommand{\mee}[1]{\textcolor{red}{ #1}}

\title{\emph{
  \textcolor{black}{``Lost}%
  -\textcolor{gray!60!black}{in}%
  -\textcolor{white!20!gray}{the}%
  -\textcolor{white!60!gray}{Later''}
}: Framework for Quantifying Contextual Grounding in Large Language Models}

\author{
Yufei Tao$^{\diamond,1}$ \quad
Adam Hiatt$^{\diamond,2}$ \quad
Rahul Seetharaman$^{\ddagger,3}$ \quad
Ameeta Agrawal$^{\diamond,4}$ \\
$^{\diamond}$Portland State University, USA \quad
$^{\ddagger}$University of Massachusetts Amherst, USA \\
\texttt{\{yutao$^1$, ahiatt$^2$, ameeta$^4$\}@pdx.edu} \quad  
\texttt{rahulseetharaman$^3$@gmail.com}
}

\begin{document}
\maketitle
\begin{abstract}
Large language models are capable of leveraging both contextual and parametric knowledge but how they prioritize and integrate these sources remains underexplored. We introduce CoPE\footnote{\url{https://github.com/PortNLP/CoPE}}, a novel evaluation framework that systematically measures  contextual knowledge (CK) and parametric knowledge (PK) across models and languages. Using our MultiWikiAtomic dataset in English, Spanish, and Danish, we analyze how large language models (LLMs) integrate context, prioritize information, and incorporate PK in open-ended question answering. {Our analysis uncovers a phenomenon we call \emph{lost-in-the-later}, where LLMs tend to overlook or deprioritize information that appears later in a given context, revealing a strong positional bias that affects contextual grounding.}  {We further find that reasoning models, as well as non-reasoning models prompted with chain-of-thought (CoT), use context even less than non-reasoning models without CoT and fail to mitigate the lost-in-the-later effect. CoT prompting, in particular, results in lower recall and shorter responses, leading to degraded contextual grounding.} Based on these insights, we design prompt-based methods to effectively leverage input context. A case study applying CoPE to summarization demonstrates that CK-informed prompting improves factual grounding and reduces hallucination. 
\end{abstract}


\section{Introduction}

\begin{figure*}[t]
    \centering
    \includegraphics[width=0.95\textwidth]{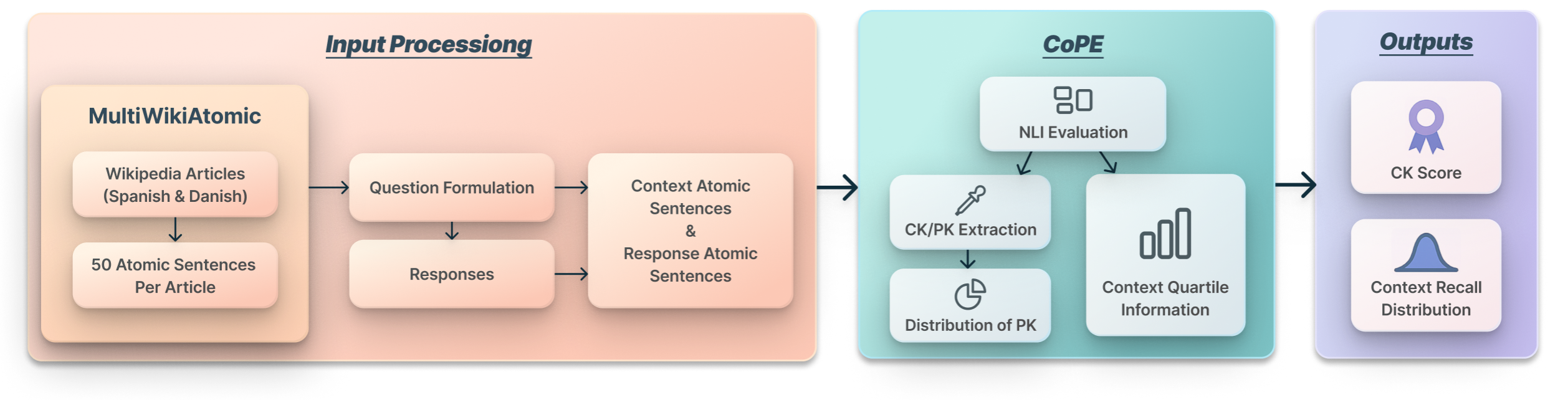}
    \caption{Overview of the MultiWikiAtomic dataset creation and CoPE framework.}
    \label{fig:pipeline}\vspace{-0.2cm}
\end{figure*}

Large language models (LLMs) have demonstrated impressive capabilities in processing contextual information to generate task-relevant and coherent responses across a variety of applications.
By conditioning outputs on provided input, LLMs can adapt to different domains, ranging from technical to legal \cite{bayarriplanas2024boostinghealthcarellmsretrieved, hu2025contextalignmentactivatingenhancingllm, chen2025clearcontextualllmempoweredprivacy}.

{However, recent studies highlight  that LLMs do not fully utilize all available context, even when sufficient contextual cues are provided \cite{ghosh2024quantifyingrelianceexternalinformation, zhang2024evaluatingexternalparametricknowledge}. 
Following prior work, we adopt a behavioral definition: \textbf{Contextual Knowledge} (CK) refers to content that is entailed by the input context, while \textbf{Parametric Knowledge} (PK) includes any output that is not entailed by the context \cite{tao2024contextleadsparametricmemory,kortukov2024studyinglargelanguagemodel,xu2024knowledgeconflictsllmssurvey,bi2025parametersvscontextfinegrained}. PK may include memorised facts, spontaneous generalisations, or new inferences; thus, our classification is behavioral only. This framing allows us to study whether outputs are grounded in the provided input, without assuming anything about how that knowledge is stored internally. 

{Context grounding is also distinct from factual correctness: a model can faithfully follow context that is itself inaccurate, or it can produce correct answers that are not grounded in the context at all.} Understanding how LLMs balance CK and PK is critical, especially in high-stakes domains such as medicine or law, where relying too much on PK can potentially increase the risk of factual inaccuracies and hallucination.} Yet existing frameworks to precisely quantify CK and PK are lacking. Furthermore, since models predominantly trained on English may exhibit different CK-PK balances when processing other languages, multilingual evaluation remains an important yet largely unexplored area.  

{In parallel, reasoning-oriented prompting strategies such as Chain-of-Thought (CoT) have gained popularity \cite{wei2022chain,sprague2024cot}. CoT is often assumed to help models stay aligned with input context by encouraging more deliberate reasoning. Yet it remains unclear whether CoT also improves contextual grounding or mitigates positional biases. Our findings challenge this assumption. We show that CoT prompting not only fails to reduce reliance on parametric knowledge but also results in degraded recall from later context segments, further exacerbating the “lost-in-the-later” effect. Moreover, reasoning models, which produce multi-step outputs by design, demonstrate the lowest contextual knowledge (CK) scores across all settings, highlighting a potential trade-off between reasoning and grounding that persists even without explicit CoT prompting.}

To systematically quantify the balance between CK and PK, we introduce the Context and Parametric Evaluation (CoPE) framework. Figure~\ref{fig:pipeline} presents a schematic overview of our study, from dataset construction to CoPE-based evaluation and final insights. The process begins with the creation of the MultiWikiAtomic dataset, where Wikipedia articles are segmented into atomic sentences to serve as context. We then formulate questions based on these contexts and collect model responses, which are also atomized. Both context and response atomic units are analyzed through CoPE, which attributes the information in the response to contextual  or parametric knowledge sources. 
{Some key findings include: }

\begin{itemize}
\setlength\itemsep{0.1em}
\item LLMs do not fully utilize the context, with CK usage peaking at approximately 70\%, across all models and languages studied. 
\item {Reasoning models use context even less. Chain-of-Thought (CoT) prompting does not mitigate this and yields the lowest overall CK usage.}
\item Models prioritize earlier context, incorporating progressively less information from the later portions of the context, even when relevant—a phenomenon we term \emph{\textbf{“lost-in-the-later”}}. 
\item PK is more likely to appear at the end of responses.
\item In contradiction rich setting, CK drops as expected, but not completely.
\item Higher percentage of CK  is correlated with a reduced likelihood of hallucination.
\end{itemize}

Informed by this analysis, we then developed a simple and adaptable prompting strategy to increase CK reliance.
Our contributions are as follows:
\begin{enumerate}
\setlength\itemsep{0.1em}
\item A framework -- CoPE -- for systematically evaluating LLMs’ ability to leverage provided context versus relying on PK.
\item A multilingual dataset of 15,000 atomic sentences in English, Spanish, and Danish to study LLMs' use of CK and PK.
\item An empirical analysis using six LLMs and a simple yet effective prompting strategy for increasing context grounding.
\item {A case study showing that CK prompting, developed based on insights from CoPE, improves factual grounding in multidocument summarization while maintaining the quality.}
\end{enumerate}

\section{Context and Parametric Evaluation (CoPE) Framework}
{CoPE is model- and task-agnostic, requiring only model outputs for evaluation. Unlike prior work {limited to specific tasks, contexts, or model accessibility}, CoPE can be applied to both open- and closed-source models, and across different generation tasks. This flexibility makes it useful for studying knowledge attribution patterns in different settings.}

\begin{figure}[!t]
    \centering
\includegraphics[width=\linewidth]{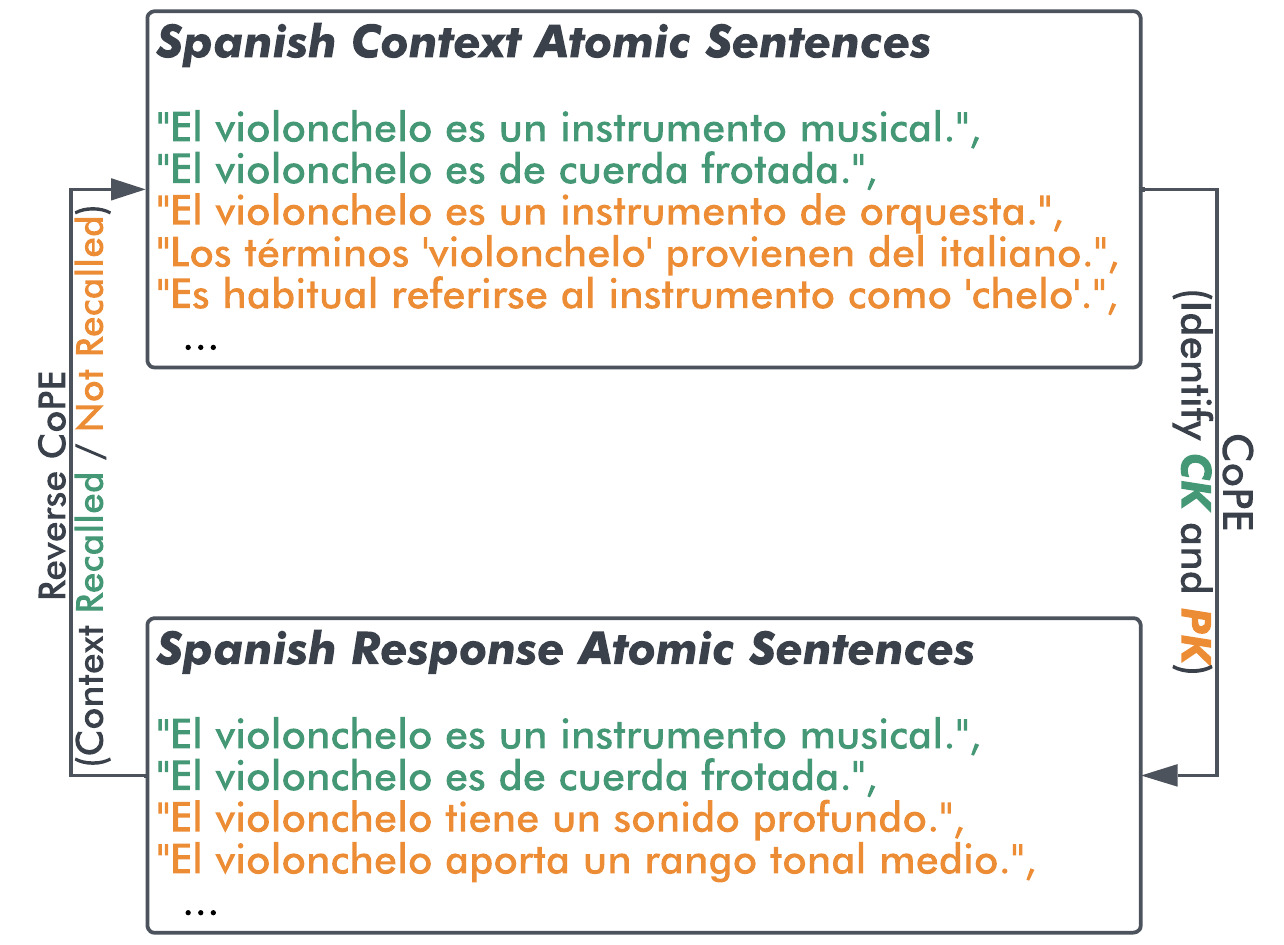}
    \caption{Example of CoPE identifying CK, PK, and context recall in a model response.} \label{fig:copeval_example}\vspace{-0.3cm}
\end{figure}

It provides a structured methodology for assessing how LLMs utilize CK and PK. Figure~\ref{fig:copeval_example} illustrates an example CoPE usage process. Given an input context and a corresponding model-generated response, CoPE computes two scores: contextual knowledge score and context recall distribution. 

\paragraph{\em Contextual Knowledge score} This metric quantifies the proportion of a  response that is directly entailed by the provided context  versus information not directly entailed from the context. Formally, given a model-generated response $R$ consisting of {$n$} atomic sentences, the CK score is computed as:
    \[
\text{CK} = \frac{\sum_{i=1}^{n} \mathbb{I} (S_i \in C)}{n} \times 100
\]
where \( C \) represents the set of entailed context sentences and \( \mathbb{I} \) is an indicator function that assigns 1 if a response atomic sentence \( S_i \) is entailed by the context and 0 otherwise. A higher CK score indicates that the model is effectively grounding its response in context. Since CK and PK are complementary, $PK = 100 - CK$.

\paragraph{\em Context Recall distribution} This metric examines how models recall information from different segments of the input context (early, middle, or late). To quantify this, the input context is divided into $k$ segments $\{Q_1, Q_2, ..., Q_k \}$, and the context recall for each segment is computed as:
\vspace{-2pt}
\[
\text{CR}_q =  \frac{\sum_{i=1}^{n} \mathbb{I} (S_i \text{ is entailed by } Q_q)}{|Q_q|}
\]
where $CR_q$ represents the recall percentage from the $q^{th}$ segment of the context, \( \mathbb{I} \) assigns 1 if atomic sentence \( S_i \) in the response is entailed by $Q_q$ and 0 otherwise, and $|Q_q|$ is the number of sentences in the $q^{th}$ segment.

\begin{figure*}[t!]
    \centering
    \begin{subfigure}[t]{0.32\textwidth}
    \includegraphics[width=\textwidth]{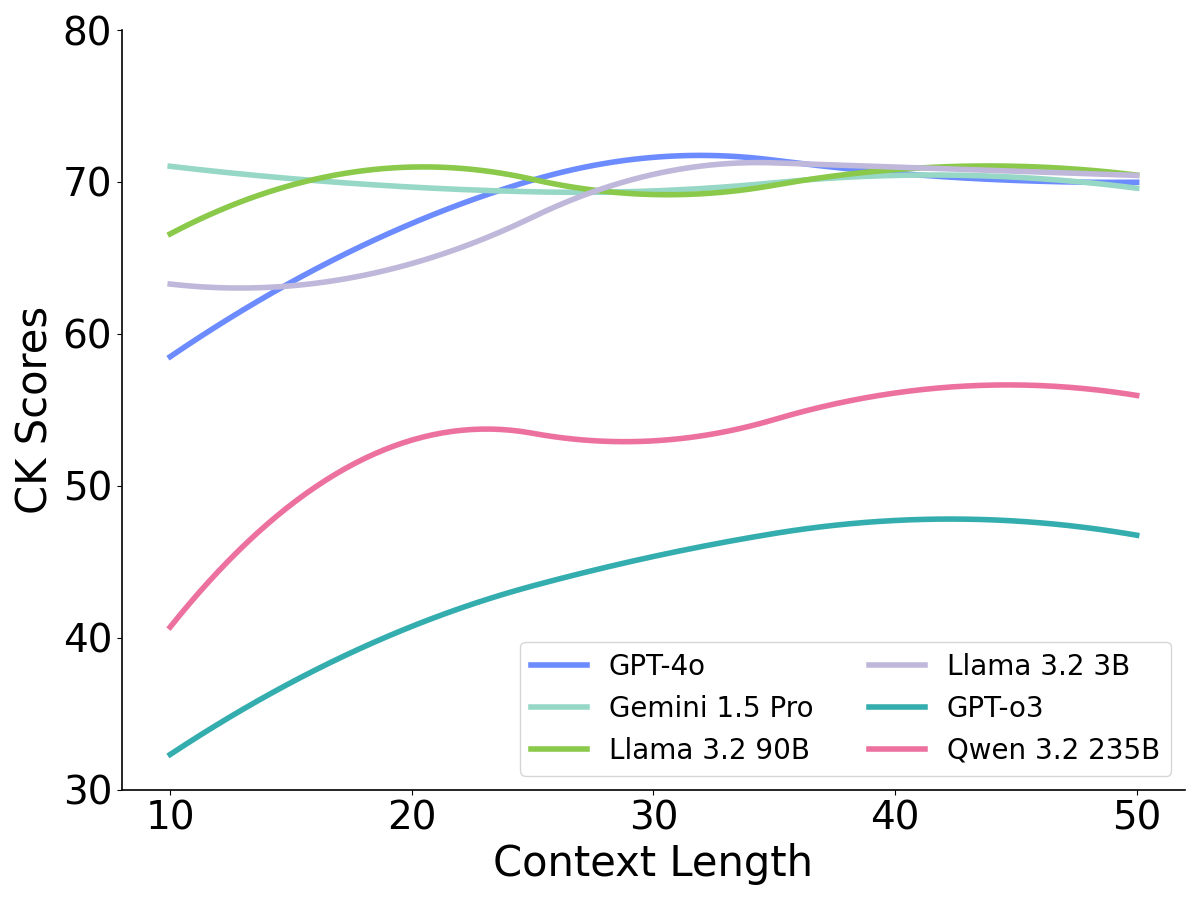}
    \caption{English}
    \end{subfigure}
    \begin{subfigure}[t]{0.32\textwidth}
    \includegraphics[width=\textwidth]{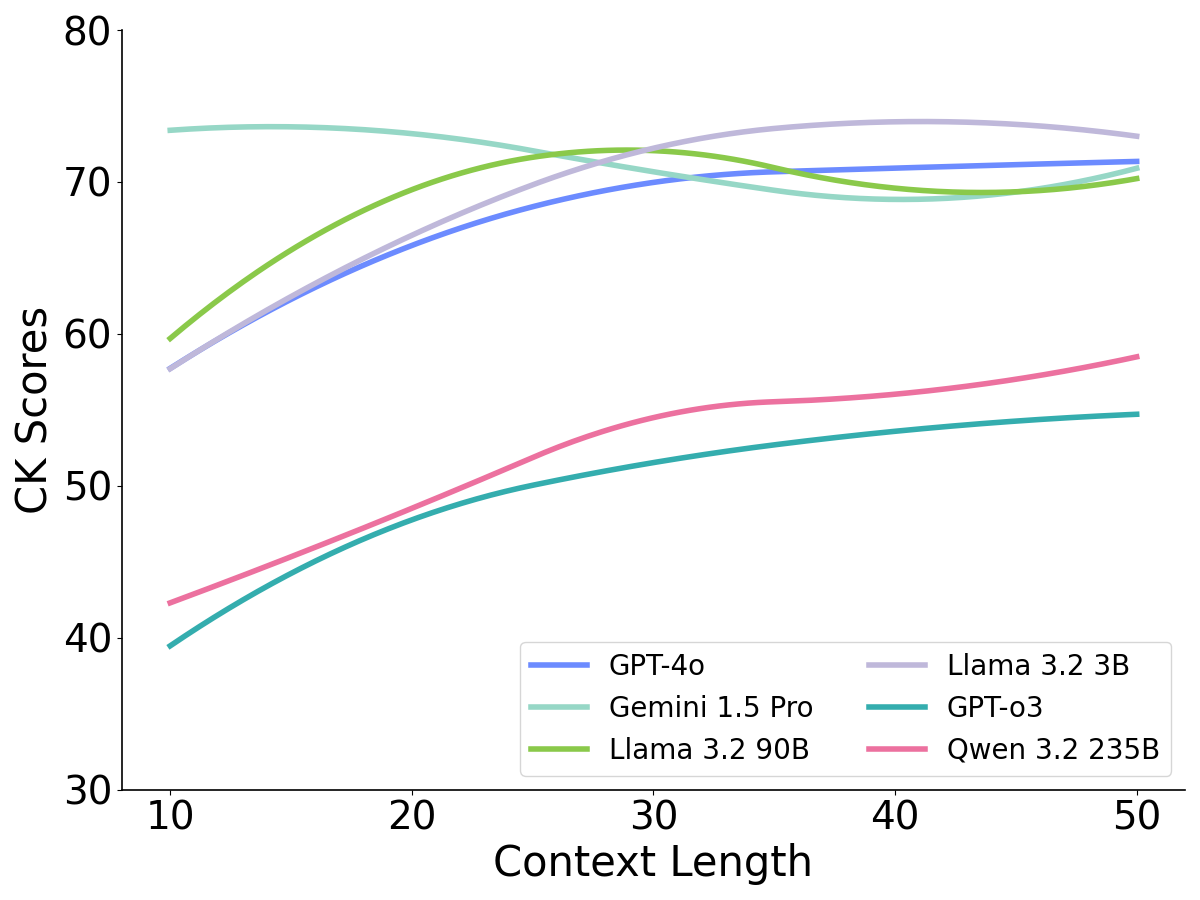}
    \caption{Spanish}
    \end{subfigure}
    \begin{subfigure}[t]{0.32\textwidth}
    \includegraphics[width=\textwidth]{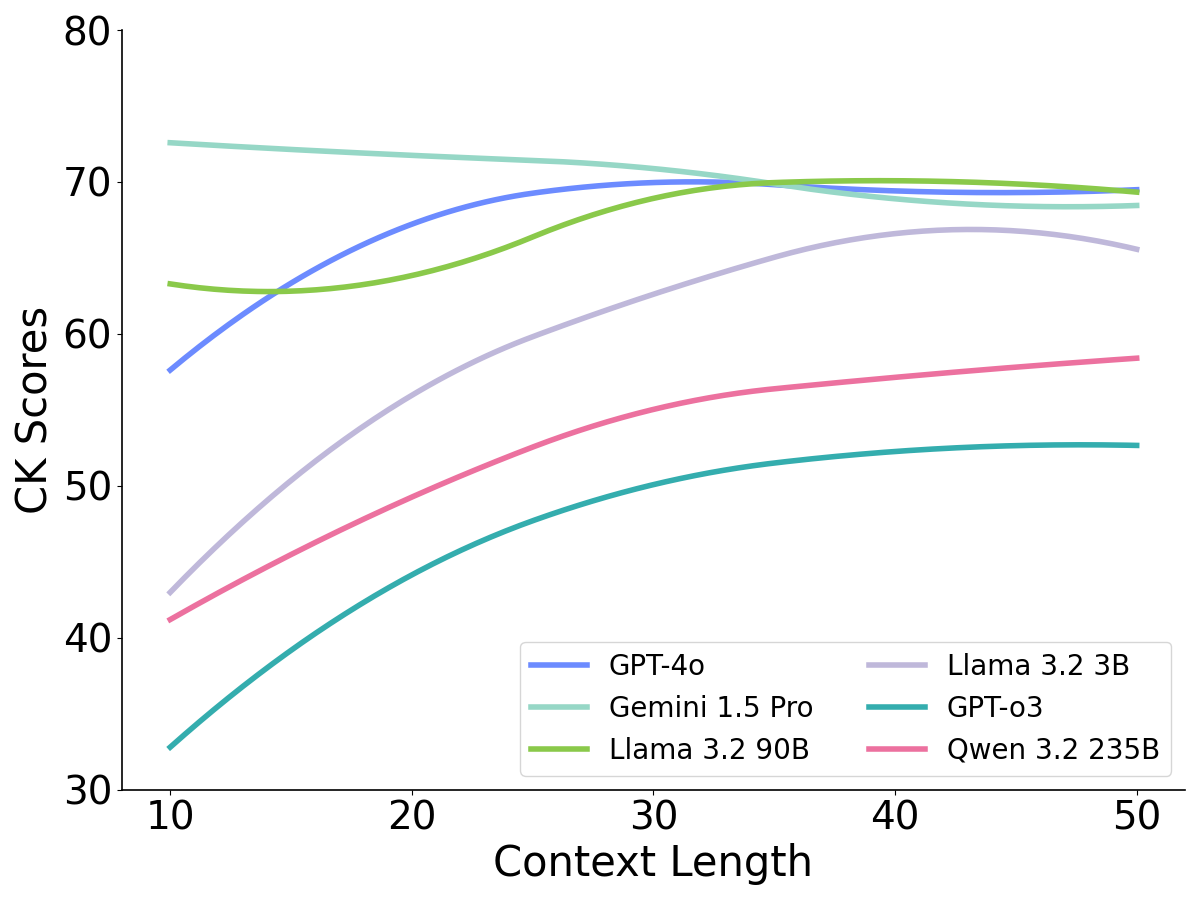}
    \caption{Danish}
    \end{subfigure}
    \caption{Contextual Knowledge (CK) scores across context lengths for different models and languages.}
\label{fig:main_analysis_combined_languages}
\end{figure*}

\paragraph{Implementation Details} To facilitate fine-grained analysis, both the input context and model responses are broken down into atomic sentences. {Atomic sentences} are minimal, standalone factual propositions that cannot be split further without losing semantic integrity. To ensure consistency across languages, we adopt the similar prompt-based sentence atomization method introduced in prior work~\citep{min2023factscorefinegrainedatomicevaluation}, which has been shown to reliably segment text into atomic sentences (see Appendix~\ref{app:breaking_atomic_prompts} for prompt details). This segmentation promotes uniformity in CK-PK classification by reducing semantic overlap and ambiguity. 

To classify atomic sentences as CK or PK, \textsc{CoPE} uses a bidirectional natural language inference (NLI) approach. 
Specifically, we adapted the \textsc{INFUSE} framework~\cite{zhang-etal-2024-fine} to aggregate entailment scores in both directions—context-to-output and output-to-context. This helps capture weaker forms of entailment and improves attribution reliability, especially in cases where rephrasings or indirect references are present. Each atomic sentence $S_i$ receives a CK-PK probability score $p \in [0,1]$, and is classified as CK if $p > t$, where $t$ is a threshold. We use \textsc{INFUSE} with the multilingual NLI model \textit{mDeBERTa-v3-base-xnli-multilingual-nli}\footnote{\url{https://huggingface.co/MoritzLaurer/mDeBERTa-v3-base-xnli-multilingual-nli-2mil7}}.

{\noindent\textbf{Threshold Calibration} \quad
We selected $t=0.7$ as the threshold for CK-PK classification based on three empirical checks. We created controlled sets in English, Spanish and Danish, with each context containing 15 CK and 5 PK atomic sentences, and found that CoPE’s classification error was less than 0.5\%. Third, we filtered out atomic sentences with scores close to the threshold, replotted CK/PK trends, and observed that shifts were minimal and the main trends held across languages and models. Full details and figures are provided in Appendix~\ref{app:implementation_detail}.}

\section{MultiWikiAtomic Dataset}
To run a controlled investigation under multilingual settings, we create \textbf{MultiWikiAtomic}, an extension of WikiAtomic \cite{tao2024contextleadsparametricmemory}, to assess how LLMs utilize context in a knowledge-consistent, open-ended question answering task across different languages. Specifically, we add 100 articles in Spanish and Danish. For each topic, we create contexts of six sizes (incrementing from 0 up to 50 sentences). In total, the dataset comprises 5,000 atomic sentences each in Spanish and Danish, which combined with existing English data brings the total to 15,000 sentences. 

\begin{figure*}[t!]
    \centering
    \begin{subfigure}[t]{0.32\textwidth}
    \includegraphics[width=\textwidth]{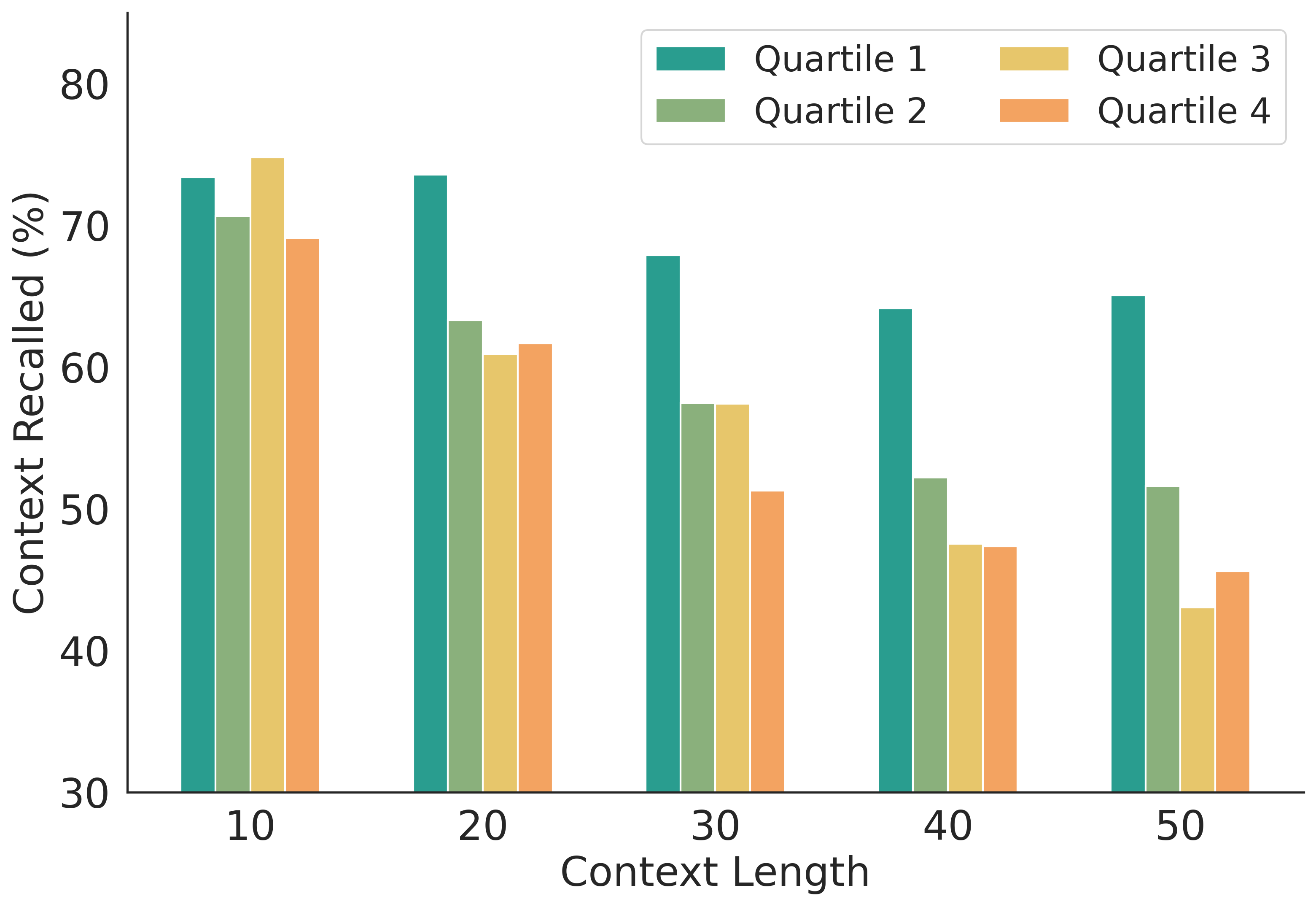}
    \caption{English}
    \label{fig:context_recall_en_llama3290b}
    \end{subfigure}
    \hfill
    \begin{subfigure}[t]{0.32\textwidth}
    \includegraphics[width=\textwidth]{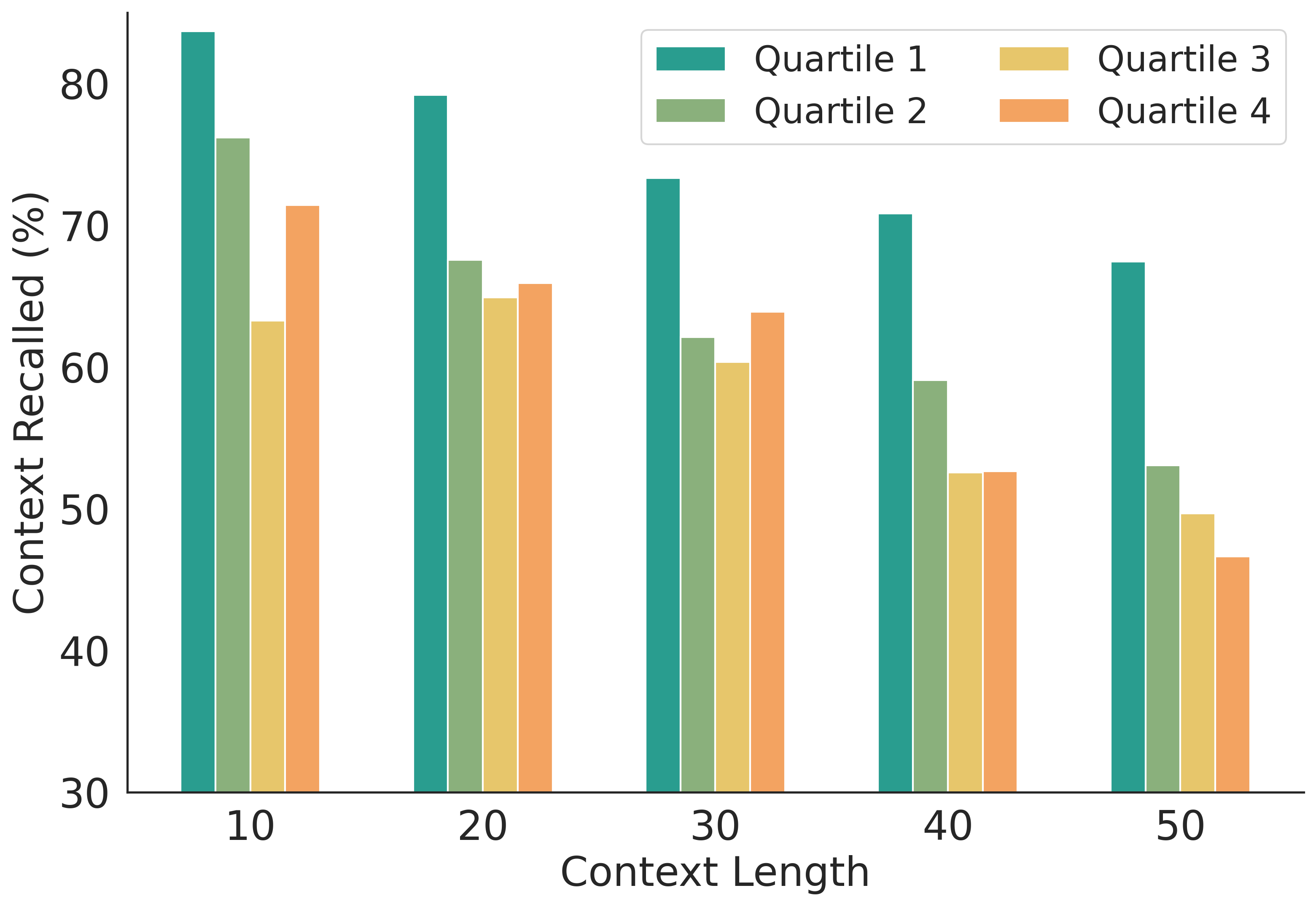}
    \caption{Spanish}
    \label{fig:context_recall_es_llama3290b}
    \end{subfigure}
    \hfill
    \begin{subfigure}[t]{0.32\textwidth}
    \includegraphics[width=\textwidth]{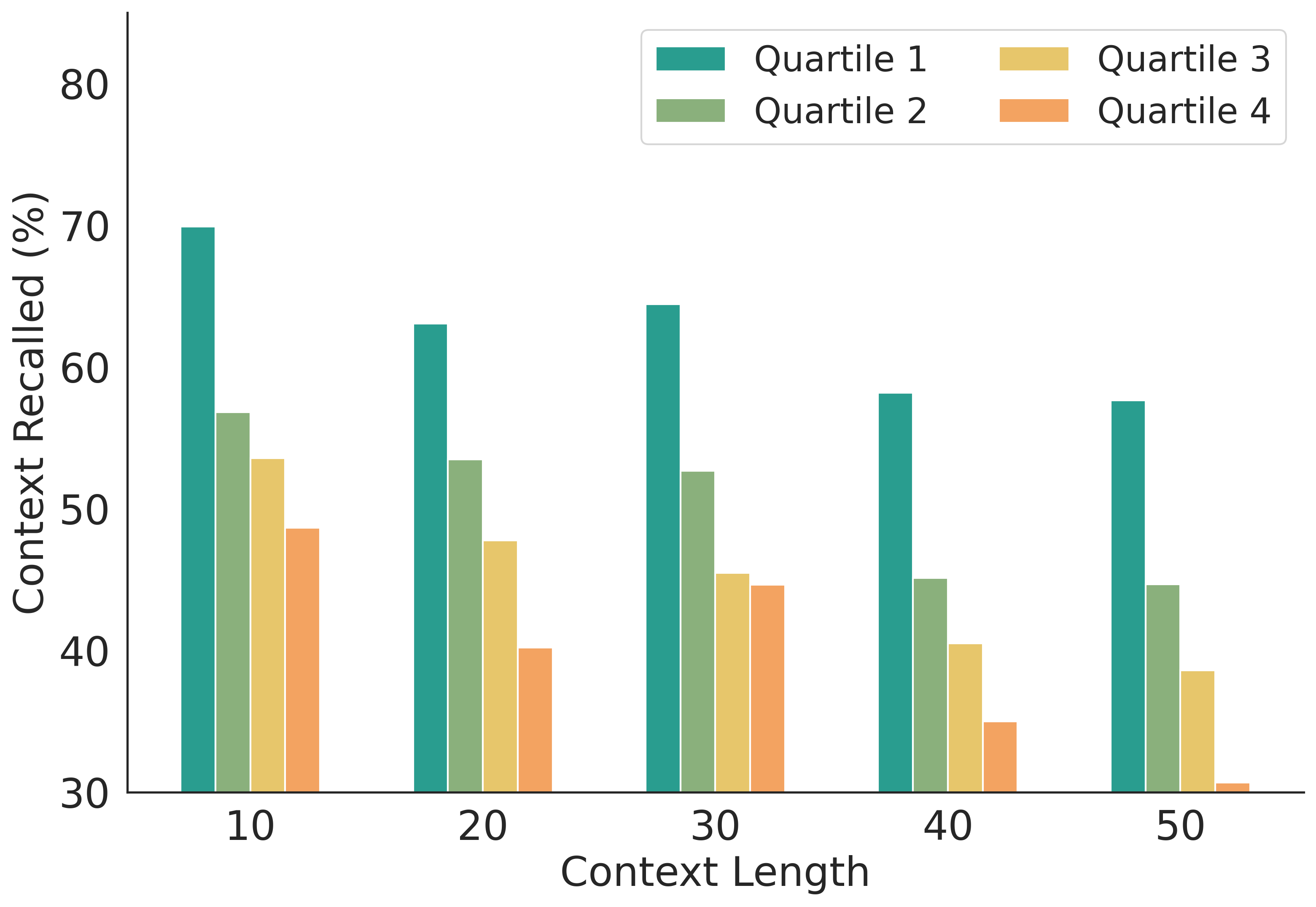}
    \caption{Danish}
    \label{fig:context_recall_da_llama3290b}
    \end{subfigure}
    \caption{Context Recall (CR) distribution of Llama 3.2 3B across English, Spanish, and Danish.}
    \label{fig:context_recall_llama323b_all_lang}\vspace{-0.3cm}
\end{figure*}

\section{Experiments and Results}
In this section we discuss model selection for our experiments and the resulting analyses using CoPE.

\subsection{Models}
To ensure the robustness of our evaluation, we include a diverse selection of both proprietary and open-weights LLMs, spanning large (GPT-4o and Gemini 1.5 Pro) and small models (Llama 3.2 90B, and Llama 3.2 3B) along with two reasoning models (GPT-o3 and Qwen 3 235B \cite{yang2025qwen3technicalreport}). For each topic, we ask an open-ended question in the form: \textit{``With this information, tell me about [Topic].''} With six models and three languages, this process generated 10,800 question–response pairs for comprehensive analysis.

\subsection{How do Models Balance CK {\em vs.} PK?}
Figure~\ref{fig:main_analysis_combined_languages} presents the CK score across varying context sizes for six models across three languages. CK scores reflect the proportion of a model’s response grounded in the provided context; hence, higher CK scores indicate stronger contextual grounding and reduced reliance on PK.

For English, most models' CK scores plateau around 70–75, indicating strong contextual grounding with increasing context. Similar trends persist for Spanish and Danish but with greater divergence among models. Overall, Gemini 1.5 Pro performs well, followed by GPT-4o, whereas smaller models (LLaMA 3.2 3B) lag behind. Interestingly, reasoning models like GPT-o3 and Qwen 3 235B show persistently low CK socres, reaching only about 55 CK even after additional context. This may be due to their longer, step-by-step answers likely drift from the original input, which makes it harder to stay grounded and may make them more prone to hallucinations.

All models improve their CK scores as context length increases, indicating that additional context generally boosts grounding. Larger models achieve higher CK scores, implying better integration of input context and less reliance on PK. CK scores are generally lower for Danish, reflecting the challenges of contextual grounding in less-resourced languages.


\subsection{How is Context Recalled?}
Next we examine how models recall information from different parts of the context. Figure~\ref{fig:context_recall_llama323b_all_lang} (full set of results in Appendix \ref{app:full_context_recall_results}) shows a consistent pattern across all three languages: models strongly favor the beginning of the input. Specifically, the first quartile of the context receives the highest attention and grounding, while the final quartile contributes the least. This imbalance becomes more pronounced with increased context length, and as we move from English to Danish.

Crucially, this pattern holds across all tested models, including both general-purpose and reasoning-focused ones. {While prior work has documented the {{“lost in the middle”}} \cite{liu2023lostmiddlelanguagemodels} issue, our results reveal a distinct and persistent \textbf{\em{“lost-in-the-later”}} effect: models increasingly underutilize later parts of the input, despite their relevance. }

{Although our inputs include only 50 atomic sentences, which is well below typical long-context thresholds, we still observe a consistent drop in recall for later context segments across models and languages. This suggests that this {\em{`lost-in-the-later'}} phenomenon is \emph{not} due to long context limitations. Instead, it points to a structural bias in how models prioritize earlier input, raising concerns even in short- to mid-length settings.}

While our Wikipedia-derived dataset preserves the natural flow of information, we examined whether disrupting this order affects models’ tendency to focus more on early context. We selected 450 questions, randomized the order of their context sentences, and used the CoPE framework to analyze model responses. The resulting context recall distributions showed only minor shifts (\textasciitilde5\%), suggesting that LLMs are relatively insensitive to the original ordering of information within the context, and the {\em `lost-in-the-later'} effect persists even when contextual ordering is removed. {This rules out surface-level data order as a primary cause. Since we also observe the effect across a range of model architectures, a potential explanation is that pretraining introduces a bias toward earlier tokens in input sequences. Future work should explore this direction in more depth.}


\subsection{Where Does PK Appear? }
\label{app:where_does_pk_appear}

Our analysis confirms that models never fully utilize all provided context and consistently integrate around 30\% PK. While models tend to prioritize earlier context, PK is still introduced systematically, raising the question of where exactly it appears in responses.

\begin{figure}[!t]
    \centering
    \includegraphics[width=0.8\linewidth]{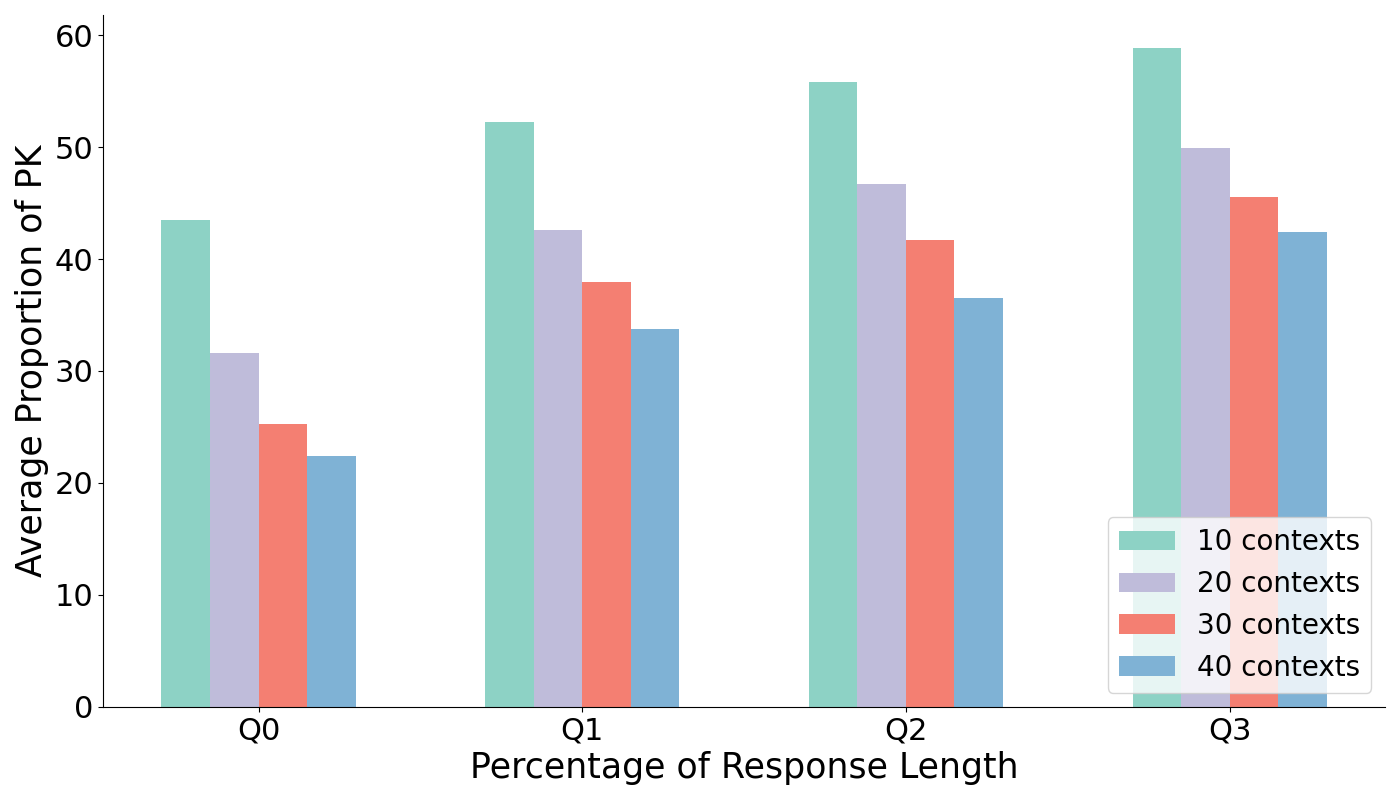}
    \caption{PK distribution across response quartiles in Spanish (Llama 3.2 3B). }
    \label{fig:es_llam323b_position_0}
\end{figure}

To investigate this, we divided each response into quartiles and measured the proportion of PK appearing in each part. Figure~\ref{fig:es_llam323b_position_0} illustrates a representative pattern observed across models and languages, showing that PK tends to increase progressively toward the final quartile (see Appendix \ref{app:full_results_position_of_0} for detailed results). This trend is particularly pronounced when fewer context sentences are available (e.g., 10 or 20),  where PK proportions remain high throughout the response. As the context grows (30--50 sentences), the rise in PK by the final quartile is more gradual yet noticeable. 


\subsection{How does Contradiction in Context Affect CK Scores and Context Recall?}
To assess whether our framework generalizes to contradiction-rich settings, we conduct further experiments by generating counterfactual examples for our  dataset of Wikipedia articles to match the structure and style of our original task. We selected 20 topics, each with 20 atomic context sentences, {and created four context conditions for each topic as follows:}
\begin{itemize}
\item \textbf{All factual:} all 20 context sentences contain factual, original information. 
\item \textbf{All counterfactual:} all 20 context sentences are made non-factual by swapping named entities (e.g., people or locations). 
\item \textbf{True-first (TF):} factual content appears in the first half, followed by counterfactuals. 
\item \textbf{False-first (FF):} counterfactuals appear first, followed by factual content.
\end{itemize}

{To avoid internal inconsistencies, we manually verified all examples to ensure logical continuity.}



\begin{table}[!t]
\centering
\small
\definecolor{headerblue}{HTML}{e0edf0}
\begin{tabular}{lcccc}
\specialrule{1pt}{0pt}{0pt}  
\rowcolor{headerblue}
\textbf{\rule{0pt}{2.6ex}\rule[-1.2ex]{0pt}{0pt}Model} & \textbf{Factual} & \textbf{CF} & \textbf{TF} & \textbf{FF} \\
\specialrule{0.5pt}{0pt}{5pt}
GPT-4o         & \textbf{72.11}  & 69.40 & 71.22 & 68.51 \\
Gemini 1.5 Pro & \textbf{76.29}  & 64.97 & 71.97 & 67.28 \\
LLaMA 3.2 90B  & \textbf{76.32}  & 66.91 & 74.44 & 64.40 \\
LLaMA 3.2 3B   & \textbf{73.13}  & 68.27 & 72.97 & 68.48 \\
GPT-o3         & 49.58  & 46.34 & \textbf{55.75} & 54.37 \\
Qwen 3 235B    & 55.30  & 52.85 & \textbf{58.02} & 55.62 \\
\bottomrule
\end{tabular}
\caption{CK scores for the English dataset under four context conditions: \textbf{Factual} (all factual), \textbf{CF} (all counterfactual), \textbf{TF} (true-first), and \textbf{FF} (false-first).}
\label{tab:contradiction_summary}
\end{table}

\begin{figure*}[ht]
    \centering
    \begin{subfigure}[t]{0.329\textwidth}
    \includegraphics[width=\linewidth]{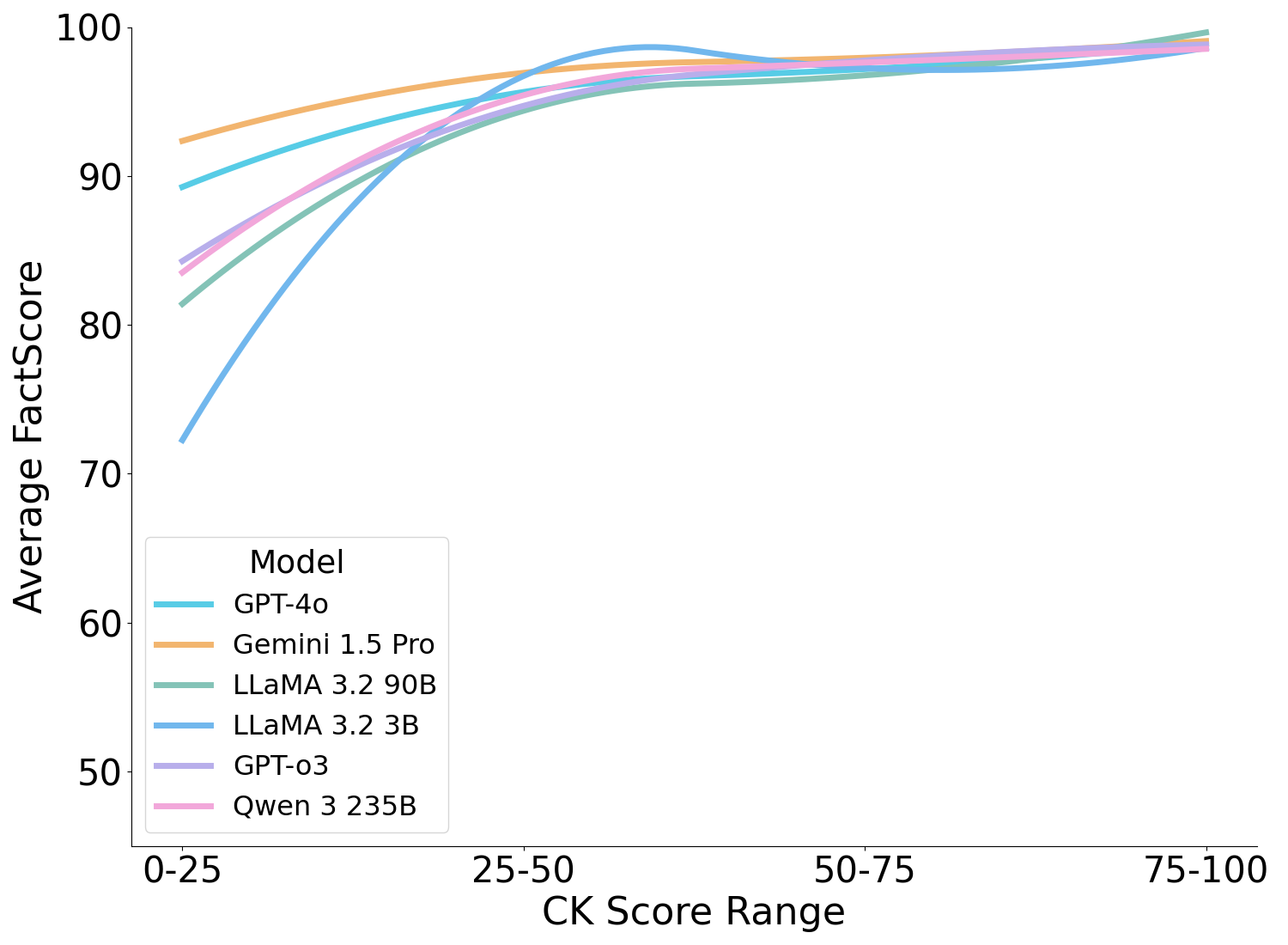}
    \caption{English}
    \end{subfigure}
    \begin{subfigure}[t]{0.329\textwidth}
    \includegraphics[width=\linewidth]{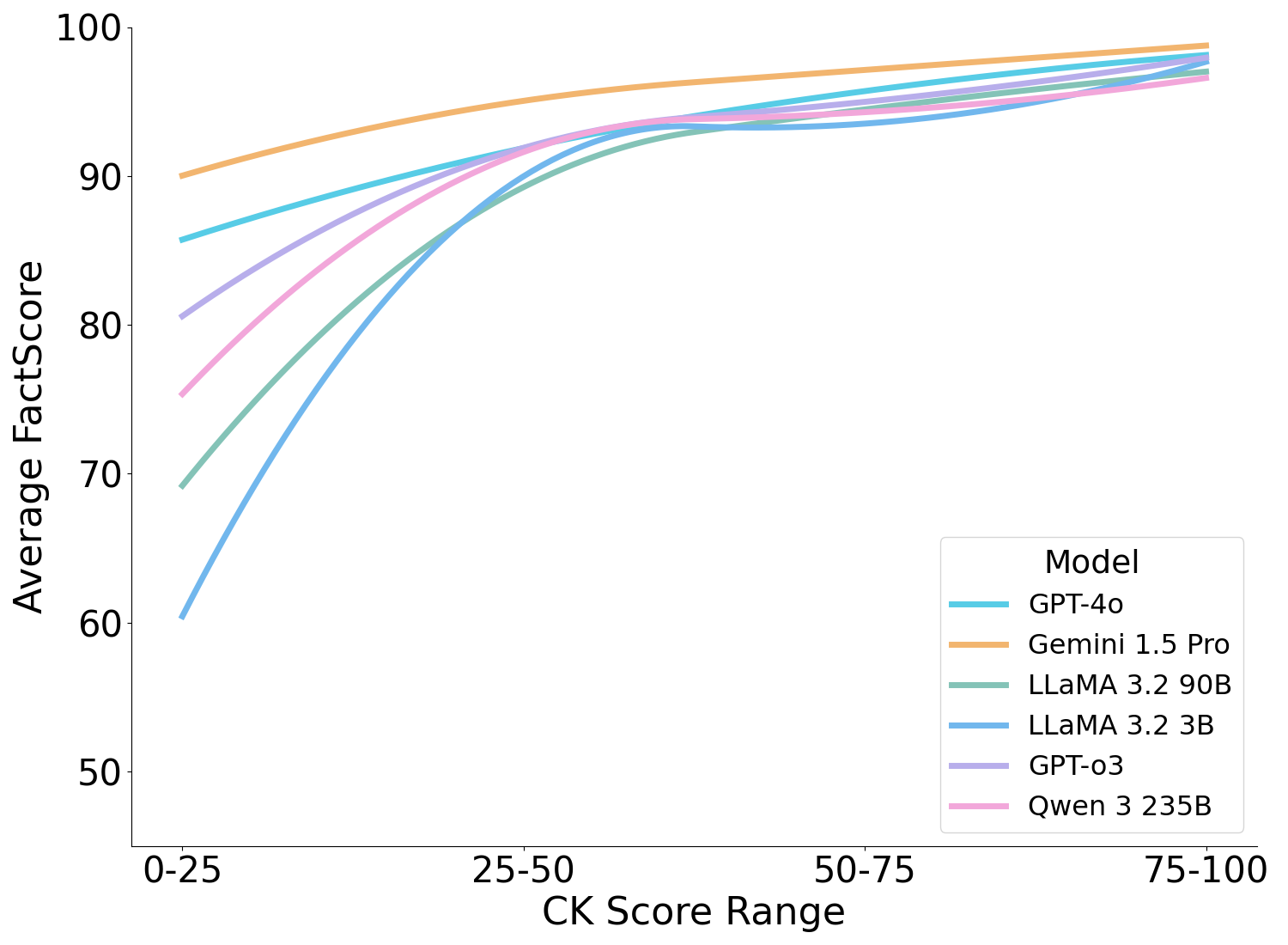}
    \caption{Spanish}
    \end{subfigure}
    \begin{subfigure}[t]{0.329\textwidth}
    \includegraphics[width=\linewidth]{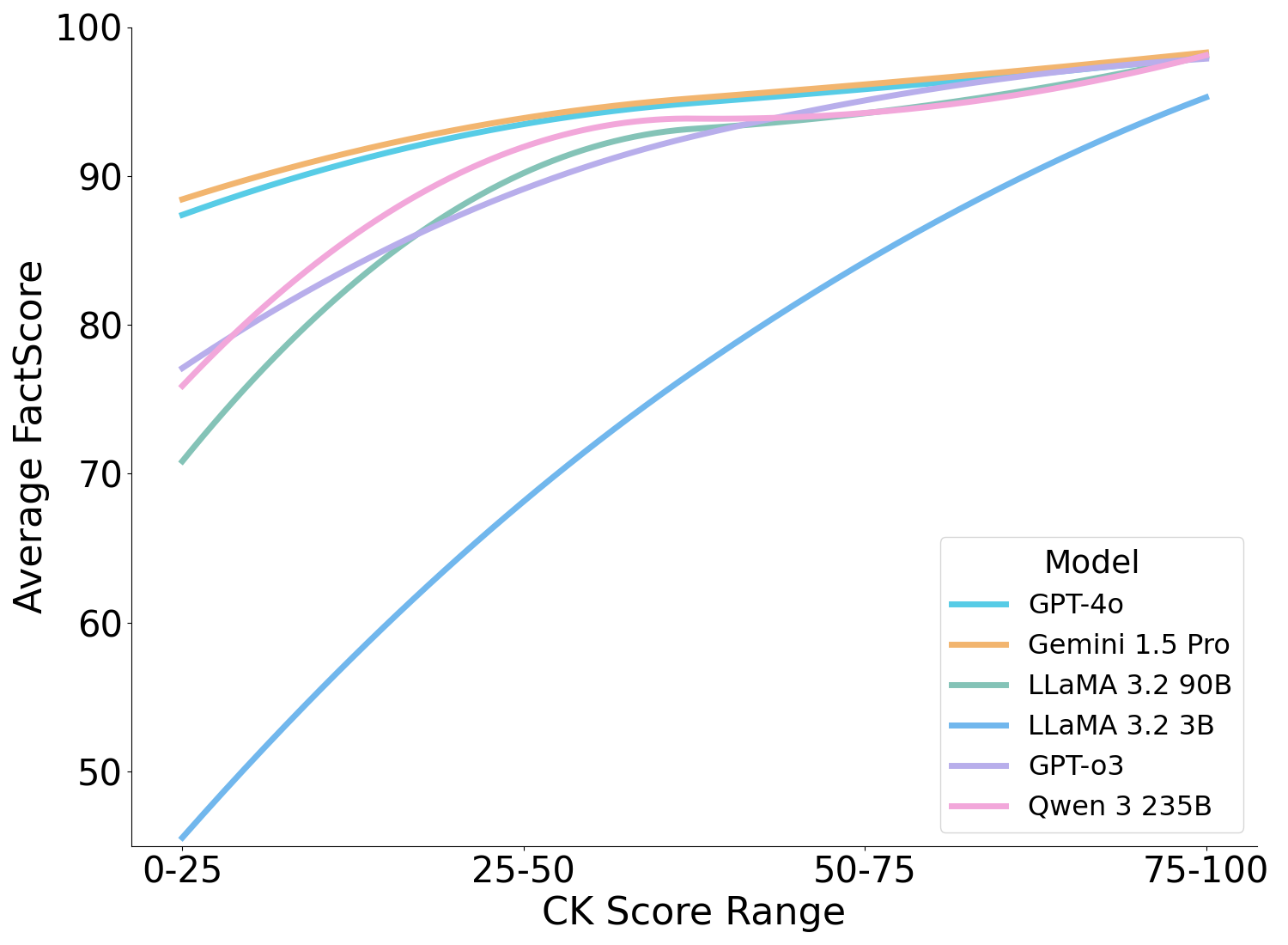}
    \caption{Danish}
    \end{subfigure}
    \caption{Average FActScore for each model in English, Spanish, and Danish, grouped by different CK score ranges. Higher FActScore indicates lower hallucination.}
    \label{fig:ckscore_factscore_plots}
\end{figure*}

Table \ref{tab:contradiction_summary} presents CK scores for six models evaluated under four context conditions (see Appendix \ref{app:contradiction_setting} for full results). All models achieve highest CK scores under `factual' condition, except for reasoning models which obtain highest CK scores under `true-first' condition. All models experience a drop in CK score under fully `counterfactual' contexts. For example, the steepest drops are observed for Gemini 1.5 Pro from 76.29 to 64.97 and LLaMA 3.2 90B drops from 76.32 to 66.91. The score drop reflects reduced contextual grounding when the context contradicts prior model knowledge. It suggests models detect and hesitate to fully anchor to false context. However, although CK scores drop in `counterfactual' setting, they do not drop to zero. This distinction is important: CK reflects grounding to input, not real-world truth, means models are following context even if that context is factually wrong. 

{Second, in mixed settings (FF and TF), CK scores typically fall between those of the fully factual and fully counterfactual conditions, confirming that contradiction weakens contextual grounding. Input order also matters: CK scores are often higher in the `true-first' setting than in `false-first', indicating that models make better use of early factual content than late-arriving corrections. This may reflect a continuation of the {\em `lost-in-the-later'} effect, now amplified by contradiction.
}

\subsection{How Does CK Score Correlate to Hallucination?}
Hallucination is a well-documented issue in LLMs where generated content deviates from the provided context. Building on the hypothesis that increased reliance on CK correlates with a lower likelihood of generating hallucinated content, we measure hallucination rates of the responses using modified FActScore framework \cite{min2023factscorefinegrainedatomicevaluation} (see Appendix \ref{app:rest_fact_scores} for details). 

{To investigate the relationship between CK score and hallucination, we computed the average FActScore for each response and grouped these values according to CK score range. 
{As shown in Figure \ref{fig:ckscore_factscore_plots}, across all languages and non-reasoning models, we observe a clear trend: higher CK scores are associated with higher FActScores. This supports our hypothesis that grounding responses in context reduces hallucination risk and improves factual accuracy.}



{Larger models maintain high FActScores across all CK ranges. Smaller and reasoning models start with lower FActScores, but their factual consistency improves quickly as CK increases, allowing them to nearly match the larger models when CK is high. At comparable CK scores, large reasoning models like GPT-o3 and Qwen 3 235B achieve FActScores similar to LLaMA 3.2 90B but remain slightly lower than the largest non-reasoning models. The smallest model, LLaMA 3.2 3B, starts with the lowest FActScores but also shows steepest improvement, ultimately achieving high factual consistency at maximal CK. Notably, as CK approaches its highest range, FActScores for all models converge, regardless of size or architecture.}

{Language resource level also plays a role. Models generally start with higher FActScores in English. In Spanish and Danish, which have fewer resources than English, initial scores are lower but improve substantially as CK increases. Although the maximum FActScore in these languages remains slightly below English, all models show the same sharp trajectory of improvement as responses become more grounded in context.}

{Taken together, these results demonstrate that increasing contextual grounding is a universally effective way to mitigate hallucination. When CK is high, all models, regardless of size, architecture, or language resource, are able to achieve strong factual consistency in their responses.}

\section{Contextual Knowledge Prompts}
{Building on our detailed analysis, we explore whether whether better prompting strategies can improve models’ ability to ground responses in context. Our goals are threefold: to increase CK scores, reduce the lost-in-the-later effect by encouraging recall from later parts of the context, and promote more even context utilization overall. We experiment with six prompt variants: a baseline (\textbf{Original}), a prompt that restricts the model to only the input (\textbf{Strict}), one that encourages drawing evenly from all context segments (\textbf{Balanced}), a combination of both constraints (\textbf{CK Prompt}), a chain-of-thought version (\textbf{CoT}), and a merged version that integrates CoT with CK (\textbf{CoT + CK Prompt}). Full prompt texts are provided in Appendix~\ref{app:ck_prompts_with_cot}.}

{Results from LLaMA 3.2 90B in English are shown in Figure~\ref{fig:difference_instruction_context_recall_all} and Table~\ref{tab:context_score_50_en} (for similar results from other models and languages, see Appendix \ref{app:prompt_solutions}). The strict prompt addressed the uneven recall distribution to a certain extent but the gap between the first and last quartiles persisted. The balanced and CK prompts further reinforced even context utilization, leading to a more uniform recall distribution. The CK prompt offers the most consistent gains, improving CK, reducing PK, and leading to a much more even recall distribution. This makes it the most effective strategy for addressing lost-in-the-later issue.}


{We also evaluated \texttt{CoT} and \texttt{CoT + CK} prompts. Although they were able to increase the CK scores, both led to substantially lower context recall than all other prompt styles—including \texttt{original}, \texttt{balanced}, and \texttt{CK Prompt}. This drop is likely due to two factors: (1) CoT encourages reflection and synthesis beyond the given context, and (2) CoT outputs are often over 50\% shorter, as reasoning steps consume much of the token budget (see Appendix \ref{app:cot_length} for details).} {Still, \texttt{CoT + CK Prompt} significantly outperforms CoT alone, improving CK usage by 10–15 points. However, it remains below the best non-CoT prompts. Notably, smaller models like LLaMA 3.2 3B often failed to produce coherent CoT responses, further reducing CK. 

\begin{figure}[!t]
    \centering
    \includegraphics[width=0.95\linewidth]{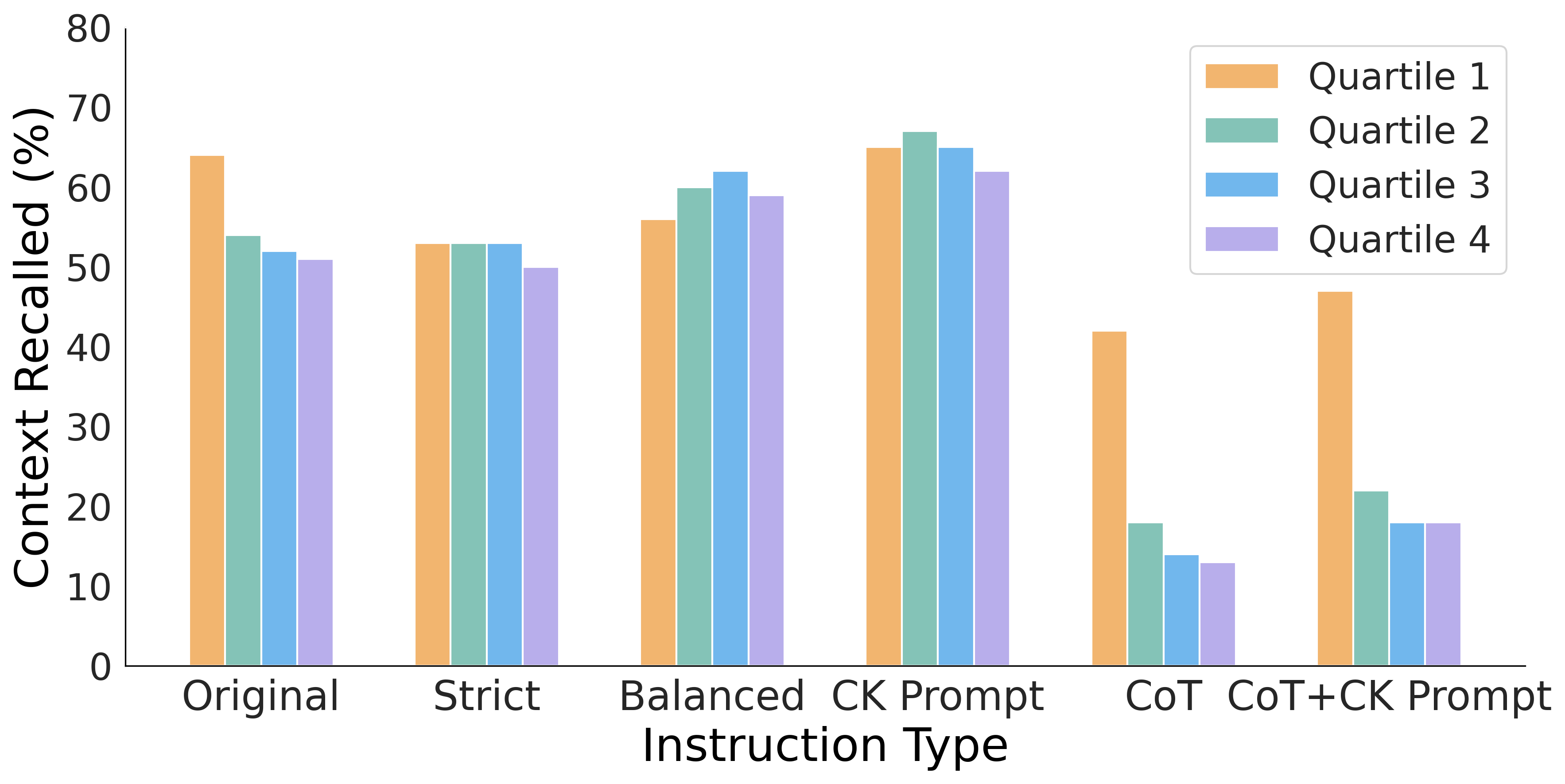}
    

    
    
    \caption {Context recall (at 50 contexts) across different instructions in English using \texttt{Llama 3.2 90B}, illustrating the impact of prompting strategies on recall distribution across quartiles.}
    \label{fig:difference_instruction_context_recall_all}
\end{figure}

{Overall, Table~\ref{tab:context_score_50_en} confirms that \texttt{CK prompt} consistently yields the highest CK scores across English, Spanish, and Danish. Though no prompt fully solves all challenges, this simple strategy provides a practical way to mitigate CK underuse and recall imbalance in current models.}

\begin{table}[t]
\centering
\small
\definecolor{headerblue}{HTML}{e0edf0}
\begin{tabular}{lccc}
\specialrule{1pt}{0pt}{0pt}  
\rowcolor{headerblue}
\textbf{\rule{0pt}{2.6ex}\rule[-1.2ex]{0pt}{0pt}Prompt} & \textbf{English} & \textbf{Spanish} & \textbf{Danish} \\ \\
\specialrule{0.5pt}{-10pt}{5pt}
Original            & 70.45  & 70.22  & 69.33 \\
Strict              & 75.86  & 75.86  & 76.67 \\
Balanced            & 61.54  & 70.21  & 64.00 \\
CK Prompt           & \textbf{78.72} & \textbf{78.95} & \textbf{77.50} \\
CoT  + CK Prompt              & 73.10  & 74.05  & 75.22 \\
CoT + CK Prompt     & 75.10  & 78.28  & 76.91 \\
\bottomrule
\end{tabular}
\caption{CK scores at 50-context setting across languages and prompting strategies using LLaMA 3.2 90B.}
\label{tab:context_score_50_en}
\end{table}

\section{Case Study: Summarization with CK Prompt}
We apply our best-performing CK prompt to multi-document summarization to evaluate its effectiveness. Given that relevant information can appear anywhere within the context, multi-document summarization provides an ideal test case. 

\subsection{Datasets and Metrics}
We use two abstractive summarization datasets: QMSum \cite{zhong2021qmsumnewbenchmarkquerybased} and DivSum \cite{olabisi-etal-2022-analyzing}. QMSum consists of 232 meeting transcriptions with corresponding summaries, while DivSum includes 25 topic inputs and human-written summaries. 


We generated summaries using \texttt{LLaMA 3.2 90B} under two prompting strategies: Base and CK prompt (see Appendix \ref{app:summ_details} for implementation details). The model-generated summaries are evaluated using standard summarization metrics such as {ROUGE-L} \cite{lin2004rouge}, {BERTScore} \cite{zhang2020bertscoreevaluatingtextgeneration}, and entailment-based NLI method to compare the generated summaries to reference summaries. Additionally, we use LLM-based evaluator {G-Eval} \cite{liu2023gevalnlgevaluationusing} which measures coherence, consistency, fluency, and relevance. 
Beyond comparing the generated summaries to gold references, we also report the overall CK score, which evaluates how well the summaries are grounded in the input documents. 


\subsection{Results}
The results in Table~\ref{tab:evaluation_results} exhibit little to no differences in ROUGE-L and BERTScore. However, the NLI metric confirms that the CK prompt produces summaries that better align with human-written references. G-Eval further indicates improved consistency. The CK prompt also significantly increases CK scores, about 6-7\% higher than the base prompt. Benefiting from CoPE’s insights, CK prompt is simple and effective in  generating more factually grounded summaries with less reliance on parametric memory and reducing hallucination risks. { While we primarily report quantitative metrics, we also manually reviewed a subset of summaries and found that the CK prompt produced coherent outputs.}


\definecolor{headerblue}{HTML}{e0edf0}
\begin{table}[!t]
\centering
\setlength{\tabcolsep}{7pt}
\small
\begin{tabular}{lcc|cc}
\specialrule{1pt}{0pt}{0pt}

\rowcolor{headerblue}
\textbf{\rule{0pt}{2.6ex}Metrics} & \multicolumn{2}{c|}{\textbf{QMSum}} & \multicolumn{2}{c}{\textbf{DivSum}} \\
\rowcolor{headerblue}
 & {Base} & {CK} & {Base} & {CK} \\

\specialrule{0.5pt}{0pt}{5pt}
ROUGE-L      & 0.09 & \textbf{0.10} & \textbf{0.16} & 0.15 \\
BERTScore    & 0.43 & 0.43          & 0.86          & 0.86 \\
NLI - Gold   & 23.29 & \textbf{27.69} & 33.77 & \textbf{44.38} \\
\midrule
\multicolumn{5}{l}{\textbf{G-Eval}} \\
Coherence    & 3.49 & \textbf{3.51} & \textbf{4.43} & 4.41 \\
Consistency  & 3.20 & \textbf{3.29} & 4.44 & \textbf{4.48} \\
Fluency      & 3.00 & 3.00          & \textbf{2.97} & 2.92 \\
Relevance    & \textbf{3.31} & 3.17 & 3.86 & \textbf{4.15} \\
\midrule
CK Score     & 83.22 & \textbf{90.51} & 77.41 & \textbf{83.08} \\
\bottomrule
\end{tabular}
\caption{Traditional summarization metrics and context overlap for QMSum and DivSum datasets. CK scores are computed between generated summaries and original documents. }
\label{tab:evaluation_results}
\end{table}


\section{Related Works}

Several studies have explored how LLMs balance context and parametric knowledge, revealing that models often revert to stored knowledge even when relevant context is available, sometimes leading to inconsistencies. This challenge is further compounded by positional biases—models tend to prioritize beginning or end of the input context while neglecting details in the middle, a phenomenon known as the ``lost in the middle'' effect \cite{liu2023lostmiddlelanguagemodels, gao-etal-2024-insights}. Recent studies show that models struggle with retrieving information from both the middle and end of documents, especially in fine-tuning scenarios \cite{saito2024answerinvestigatingpositionalbias}. Beyond input order effects, research has also examined how models segment and prioritize input information, while retrieval-augmented frameworks shed light on the competition between parametric and contextual knowledge \citep{ravaut2024contextutilizationsummarizationlarge, farahani2024decipheringinterplayparametricnonparametric}. 

{Recent work has explored how LLMs balance contextual and parametric knowledge. \citet{tao2024contextleadsparametricmemory} introduce a dataset and evaluation framework to systematically measure how LLMs rely on context versus parametric memory in knowledge-consistent scenarios.  \citet{bi2025parametersvscontextfinegrained} introduces a plug-in mechanism to steer decoding away from parametric content, but requires access to token-level logits and cannot be applied to closed-source models. \citet{fu2025harnessingunseenhiddeninfluence} propose a hybrid needle-in-a-haystack QA benchmark to test reliance on intrinsic knowledge under long-context settings. \citet{liu2025olmotracetracinglanguagemodel} introduce OLMoTrace, a tracing system that identifies verbatim overlaps between model outputs and pretraining data, enabling post-hoc inspection of memorization. }


Counterfactual datasets are widely used to assess model faithfulness by testing whether models override PK in favor of provided CK when contradictions arise \cite{liu2023recallbenchmarkllmsrobustness, gat2023faithfulexplanationsblackboxnlp}. These datasets introduce modified facts or contradictions to evaluate whether models update their responses accordingly. While effective for detecting PK over-reliance, these approaches primarily assess conflict scenarios, making them less suited for measuring CK-PK balance in knowledge-consistent settings.


{In contrast, our work focuses on knowledge-consistent settings where contextual and parametric content are aligned. CoPE supports general tasks such as summarization and QA, operates at the sentence level using semantic entailment, and is compatible with both open- and closed-source models. By providing multilingual evaluation across English, Spanish, and Danish, our framework offers broader coverage than prior approaches. It also reveals a new positional bias “{\em lost in the later}” where models underutilize relevant information that appears later in context, adding a novel perspective to the study of contextual grounding.}


\section{Conclusion and Future Work}
We introduced a new dataset (MultiWikiAtomic) and  a framework (CoPE)  for systematically evaluating how  LLMs balance contextual knowledge (CK) and parametric knowledge (PK). Our framework provides detailed insights across three distinct languages -- English, Spanish, and Danish,  into models’ struggles to fully utilize newly added information, their persistent reliance on PK even when discouraged, 
{and a positional bias we term as {\em "lost-in-the-later"}—a tendency to over-prioritize early context while neglecting or failing to integrate relevant information that appears later.} 
Future work may explore more complex multilingual settings, including crosslingual scenarios, to further refine CK-PK analysis. Additionally, while our current summarization solution leverages a prompting strategy informed by CoPE, future research might utilize CoPE to develop more effective solutions in other tasks.

\section*{Limitations}
While CoPE provides a structured approach to evaluating the balance between CK and PK in multilingual LLMs, several limitations remain:

{First, CoPE has not been extensively tested on conversational or social media data, such as tweets. These types of data are often noisy and informal, which makes it challenging to break content down into reliable atomic sentences and to accurately distinguish contextual from parametric knowledge.}

Another important limitation is that in real applications, context can be inaccurate or contain errors. Our framework measures how well models ground to the context, not whether the answer is factually correct. Better grounding does not always mean better factuality. Handling both grounding and factual accuracy, especially when context is imperfect, is an open challenge for future work.

Additionally, FActScore, used to measure hallucination, was originally designed for English. While we extended it with additional knowledge sources for Spanish and Danish, its effectiveness in fully capturing factuality in non-English responses was beyond the scope of this paper. 

\section*{Acknowledgments} This research was supported in part by the National Science Foundation under Grant No. HNDS-R 2242205. We gratefully acknowledge their support.

\bibliography{custom}

\appendix

\section{Atomic Sentence Breaking Down}
\label{app:breaking_atomic_prompts}
The prompt we used to break down normal context sentences into Atomic Sentences is:

\begin{quote}
\small
(No list, not bullet points, everything should be on the same line)
Definition of Atomic: An atomic sentence is a type of declarative sentence which is either true or false, also referred to as a proposition, statement, or truth-bearer. It cannot be broken down into simpler sentences without losing its meaning.

You will do multiple passes for first 60 sentences, using appropriate NLP method if needed, for each sentence:

The first pass: remove comma in the sentence, rewrite the sentence into multiple smaller sentences if needed.

The second pass: remove 'and' and 'or' in the sentence, rewrite the sentence into multiple smaller sentences if needed.

The third pass: replace indirect references with direct references (topic word) to maintain clarity and focus on the text’s main topic.
The fourth pass: separate temporal information (dates, times) from the main action into distinct sentences.

The final pass: make sure each sentence contains exactly only one information. Nothing more than one information, even the information that are dependent on each other.

The goal of these passes is to break down each long sentence into very small atomic sentences that contain one single inseparable information.

Output Format:

A JSON object with the following elements:

atomic\_sentences: A list of 60 atomic sentences.

count: The number of sentences in the atomic\_sentences.

Not in JSON output but you need to think:

The process of each pass, are you sure you removed all commas, are you sure you removed all 'and' and 'or', are you sure you replaced all indirect references with the actual topic word.    
\end{quote}

\section{Implementation Detail}
\label{app:implementation_detail}

This section outlines the experimental setup, model parameters, and CK-PK classification method. We used default parameters across all models, setting temperature and top\_p to 1, and both presence\_penalty and frequency\_penalty to 0. 

For atomic sentence extraction—used for both Wikipedia articles and model responses—we employed GPT-4o with a `max\_tokens` limit of 2048, sufficient for returning a JSON object containing up to 50 atomic sentences. For response generation across models, we generally set `max\_tokens` to 1024 to reflect real-world usage constraints. However, for reasoning-heavy models (e.g., GPT-o3 and Qwen 3 235B), we doubled this limit to 2048 to accommodate the longer responses required for multi-step reasoning.

API calls to OpenAI and Google AI Studio were used to obtain responses from GPT-4o and Gemini 1.5 Pro, completing within 4 hours. For open-source models, Llama 3.2 models and Qwen 3 model, we used a third-party inference service, which took 3 hours. The full CoPE evaluation of responses was performed on one Nvidia 4090 GPU and completed in approximately 20 hours.

\vspace{0.5em}
\noindent \textbf{Threshold Calibration for CK-PK Classification} \quad
To determine the optimal threshold ($t$) for CK-PK classification, we conducted a preliminary evaluation using the same dataset and experimental configuration as \citet{tao2024contextleadsparametricmemory}. Their original work used \textit{ALBERT-xlarge-VitaminC-MNLI}, an English-only NLI model. To enable multilingual evaluation, we substituted it with a publicly available multilingual model: \textit{mDeBERTa-v3-base-xnli-multilingual-nli}\footnote{\url{https://huggingface.co/MoritzLaurer/mDeBERTa-v3-base-xnli-multilingual-nli-2mil7}}.

To validate this adaptation, we first computed CK-PK distributions using their setup and then compared these with outputs from our multilingual model under varying thresholds. We selected $t = 0.7$ because it yielded distributions most aligned with the original evaluation, preserving continuity while supporting Spanish and Danish.

\textbf{CoPE Accuracy Evaluation} \quad To verify CoPE’s accuracy, we tested its results on English ($n$ = 109), Spanish ($n$ = 98) and Danish ($n$ = 98) instances. We synthetically constructed responses containing 10 actual CK sentences and 5 from unrelated topics, targeting a CK score of 66.66\%. CoPE yielded an average CK score of 66.94\% (std = 1.32) for English, 66.97\% (std = 1.39) for Spanish and 66.89\% (std = 1.27) for Danish, confirming its reliability in distinguishing CK and PK.

\vspace{0.5em}
\noindent \textbf{Threshold Sensitivity Analysis} \quad
To evaluate the robustness of our CK-PK classification, we performed an ablation by filtering out all atomic sentences with entailment scores between 0.4 and 0.8.These sentences typically fall into borderline or ambiguous cases, where the model’s confidence in semantic entailment is unclear. The filtered results reflect a stricter view of what qualifies as CK.

Figures~\ref{fig:threshold_ablation_all} show CK/PK trends before and after filtering. Despite the removal of borderline cases, the trends across context length, language, and model type remain stable, with only minor variations. This supports the reliability of our framework and indicates that our findings are not highly sensitive to the specific entailment threshold chosen.

\begin{figure*}[ht]
    \centering
    \begin{subfigure}[t]{0.32\textwidth}
        \centering
        \includegraphics[width=\linewidth]{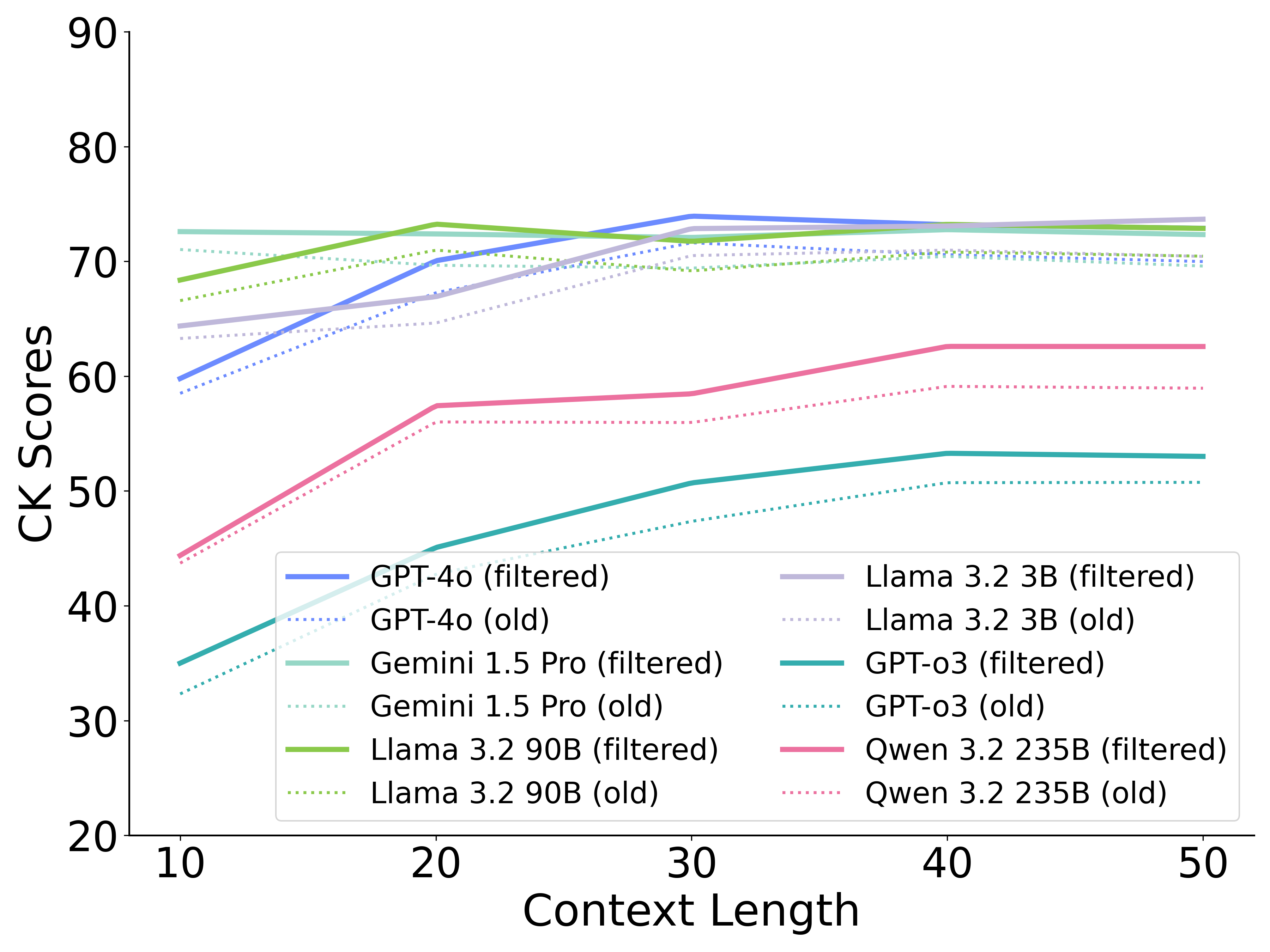}
        \caption{English}
        \label{fig:threshold_gpt4o}
    \end{subfigure}
    \begin{subfigure}[t]{0.32\textwidth}
        \centering
        \includegraphics[width=\linewidth]{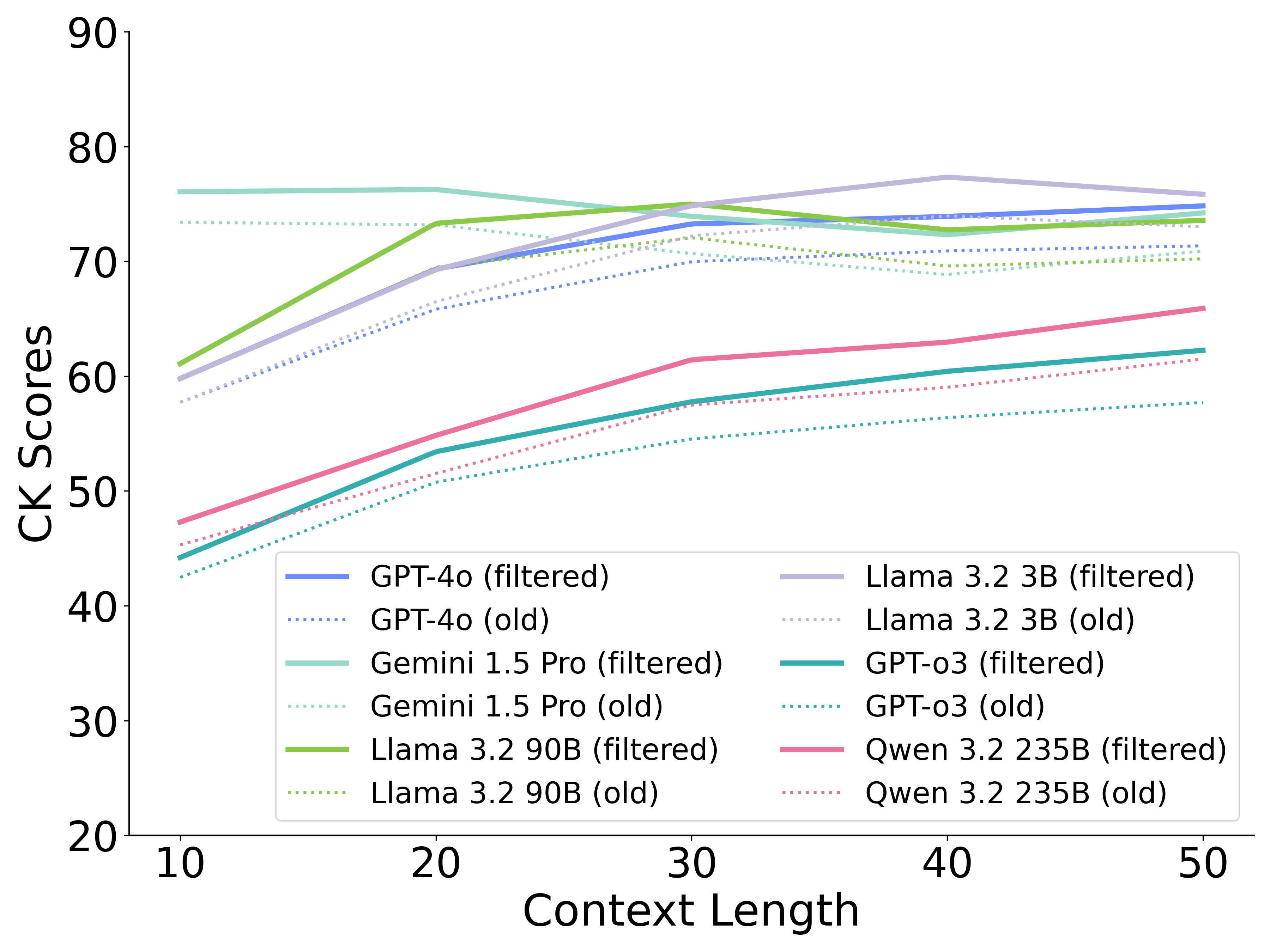}
        \caption{Spanish}
        \label{fig:threshold_gemini}
    \end{subfigure}
    \begin{subfigure}[t]{0.32\textwidth}
        \centering
        \includegraphics[width=\linewidth]{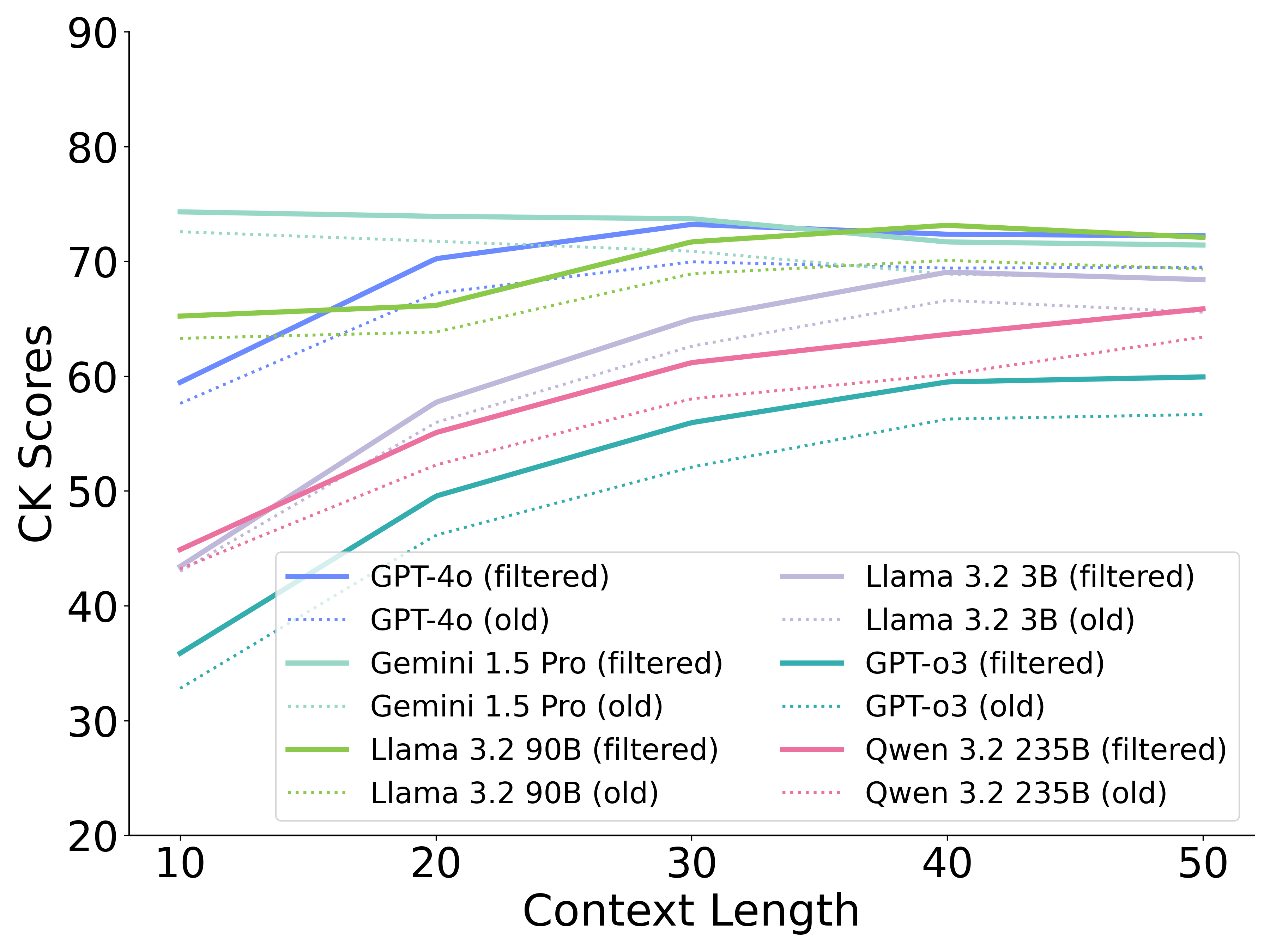}
        \caption{Danish}
        \label{fig:threshold_llama3290b}
    \end{subfigure}

    \caption{Threshold ablation results across all languages and models. Each plot shows CK/PK score trends with and without borderline entailment filtering.}
    \label{fig:threshold_ablation_all}
\end{figure*}

\section{Full Results of Context Recall}
\label{app:full_context_recall_results}
Figure \ref{fig:context_recallanalysis_english}, \ref{fig:context_recall_analysis_spanish} and \ref{fig:full_context_recall_analysis_danish} are full results of context recall in English, Spanish, and Danish, respectively, for each model.

\begin{figure*}[t!]
    \centering
    \begin{subfigure}[t]{0.32\textwidth}
    \includegraphics[width=1\textwidth]{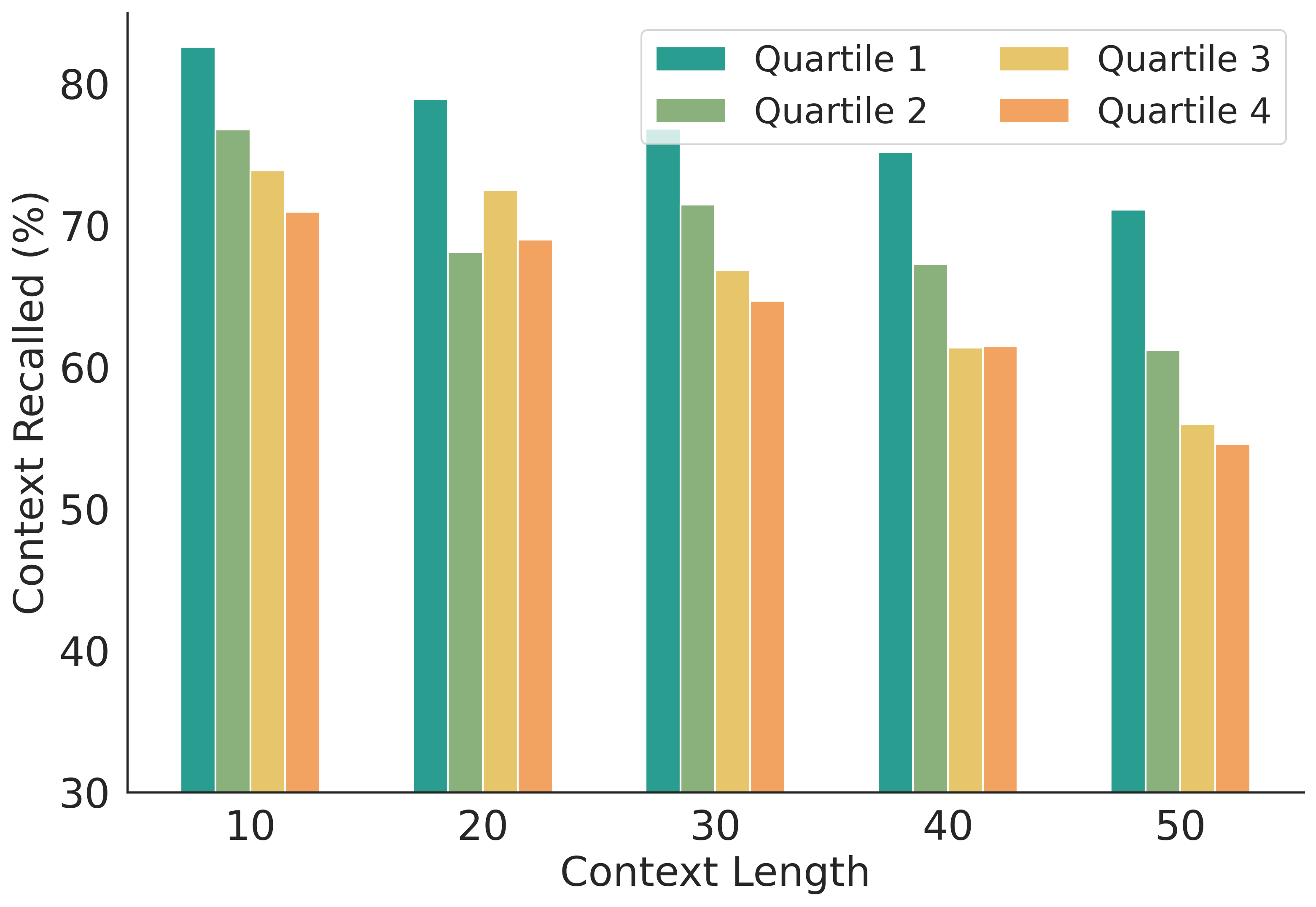}
    \caption{GPT-4o}
    \end{subfigure}
    \hfill
    \begin{subfigure}[t]{0.32\textwidth}
    \includegraphics[width=\textwidth]{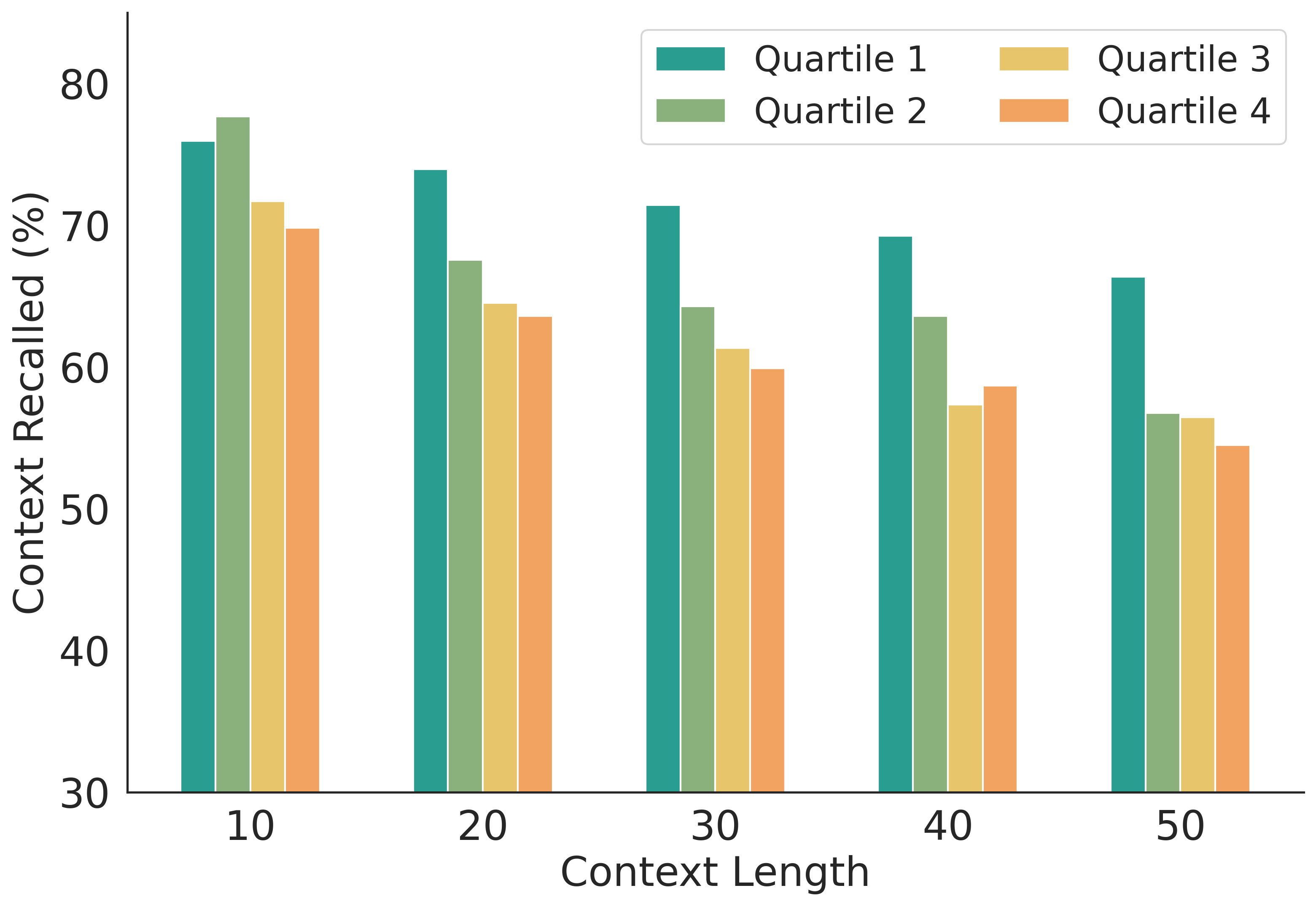}
    \caption{Gemini 1.5 Pro}
    \end{subfigure}
    \hfill
    \begin{subfigure}[t]{0.32\textwidth}
    \includegraphics[width=\textwidth]{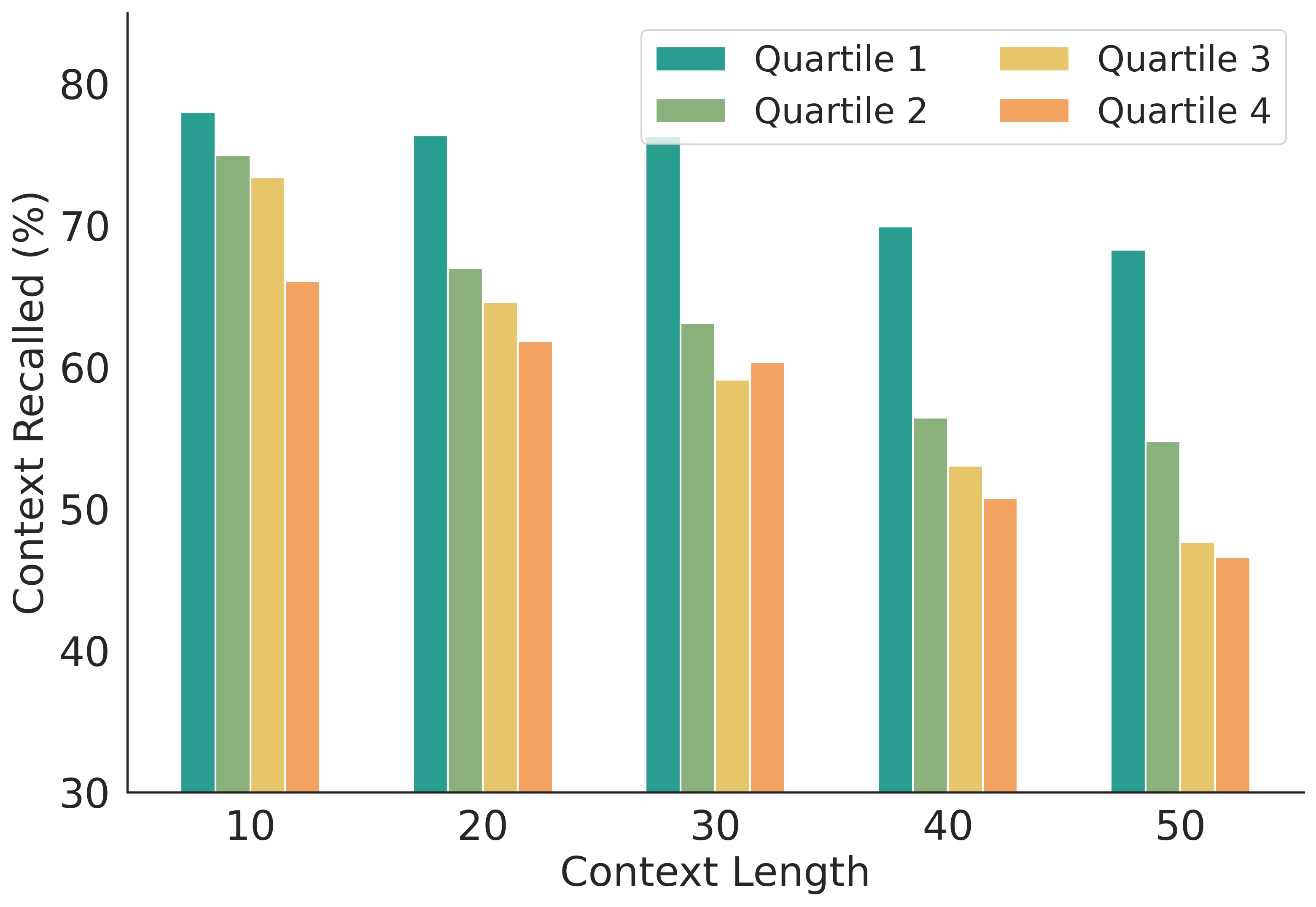}
    \caption{Llama 3.2 90B}
    \end{subfigure}
    \hfill
    \begin{subfigure}[t]{0.32\textwidth}
    \includegraphics[width=\textwidth]{figures/context_recall/fixed_y_range/en_llama323b.png}
    \caption{Llama 3.2 3B}
    \end{subfigure}
    \hfill
    \begin{subfigure}[t]{0.32\textwidth}
    \includegraphics[width=\textwidth]{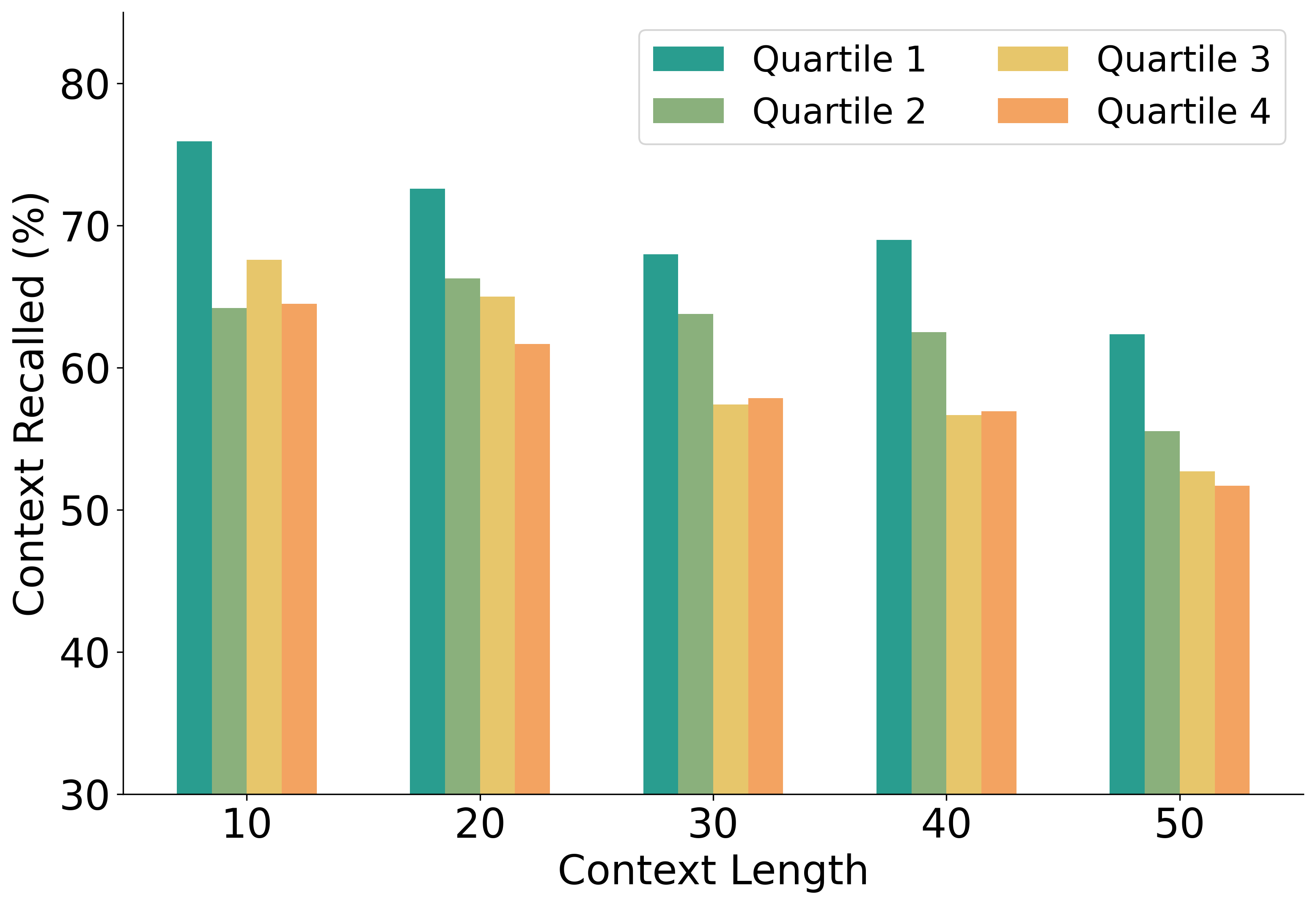}
    \caption{GPT-o3}
    \end{subfigure}
    \hfill
    \begin{subfigure}[t]{0.32\textwidth}
    \includegraphics[width=\textwidth]{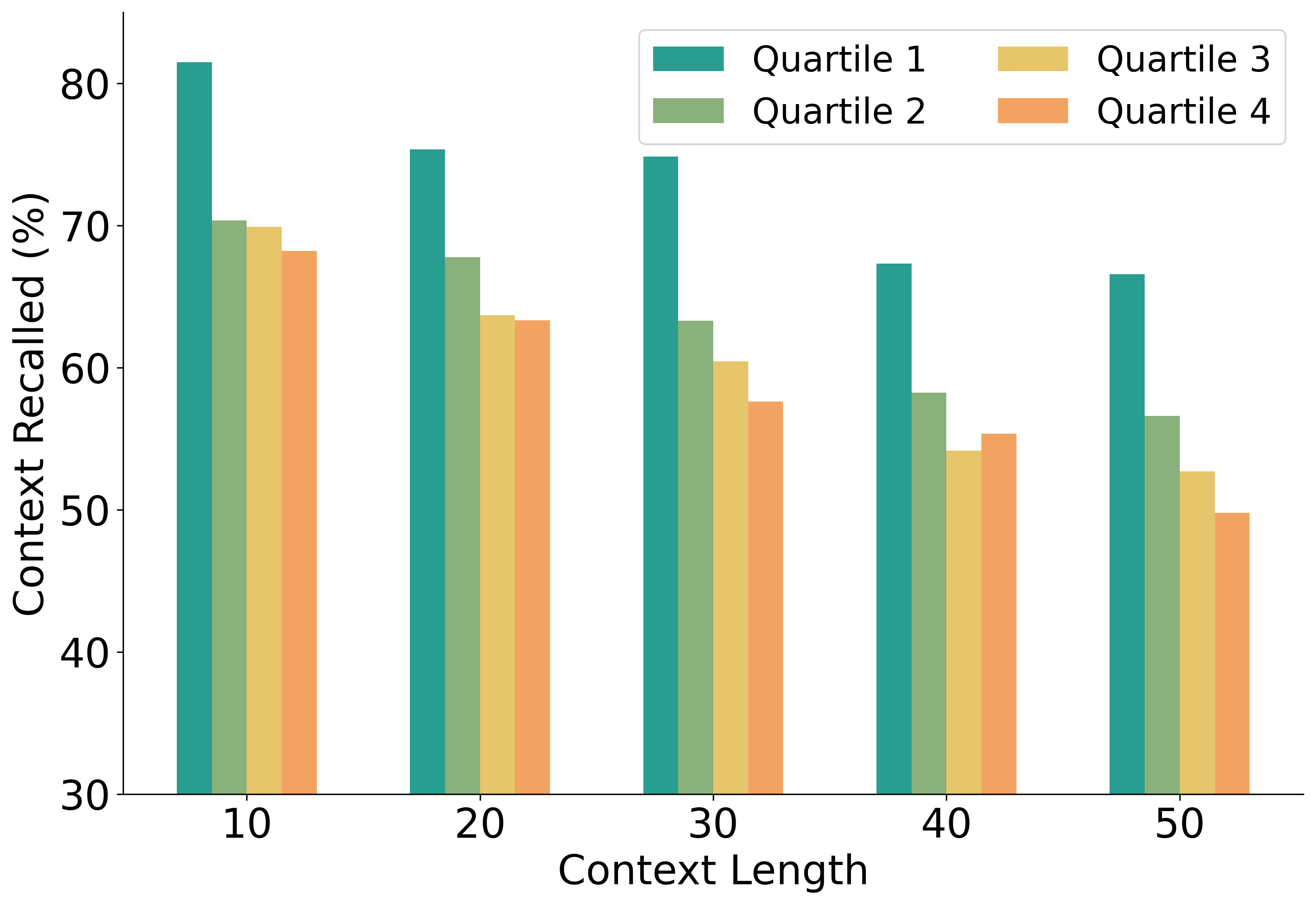}
    \caption{Qwen 3 235B}
    \end{subfigure}
    \caption{Context recall results for English}
    \label{fig:context_recallanalysis_english}
\end{figure*}

\begin{figure*}[t!]
    \centering
    \begin{subfigure}[t]{0.32\textwidth}
    \includegraphics[width=\textwidth]{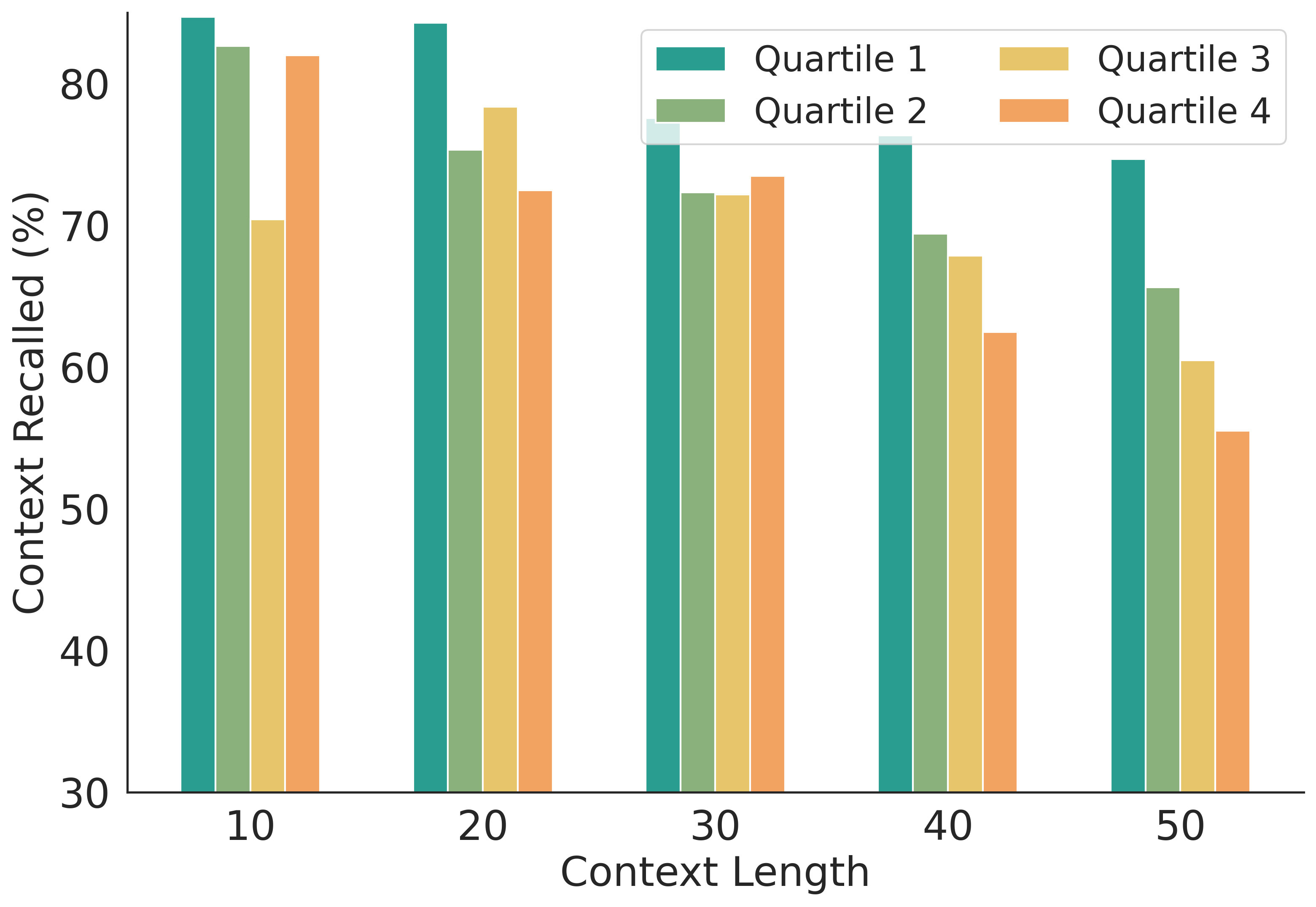}
    \caption{GPT-4o}
    \end{subfigure}
    \hfill
    \begin{subfigure}[t]{0.32\textwidth}
    \includegraphics[width=\textwidth]{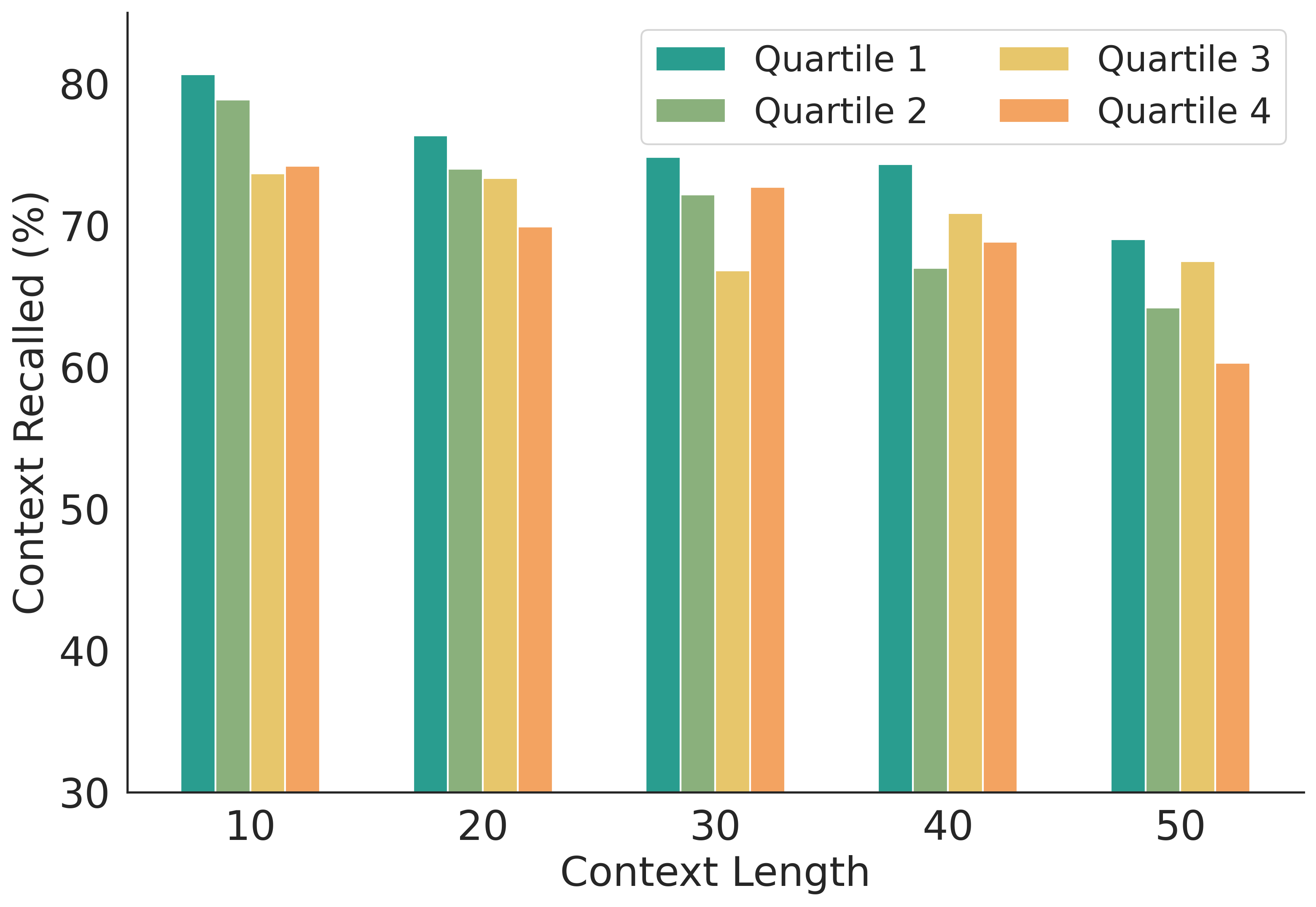}
    \caption{Gemini 1.5 Pro}
    \end{subfigure}
    \hfill
    \begin{subfigure}[t]{0.32\textwidth}
    \includegraphics[width=\textwidth]{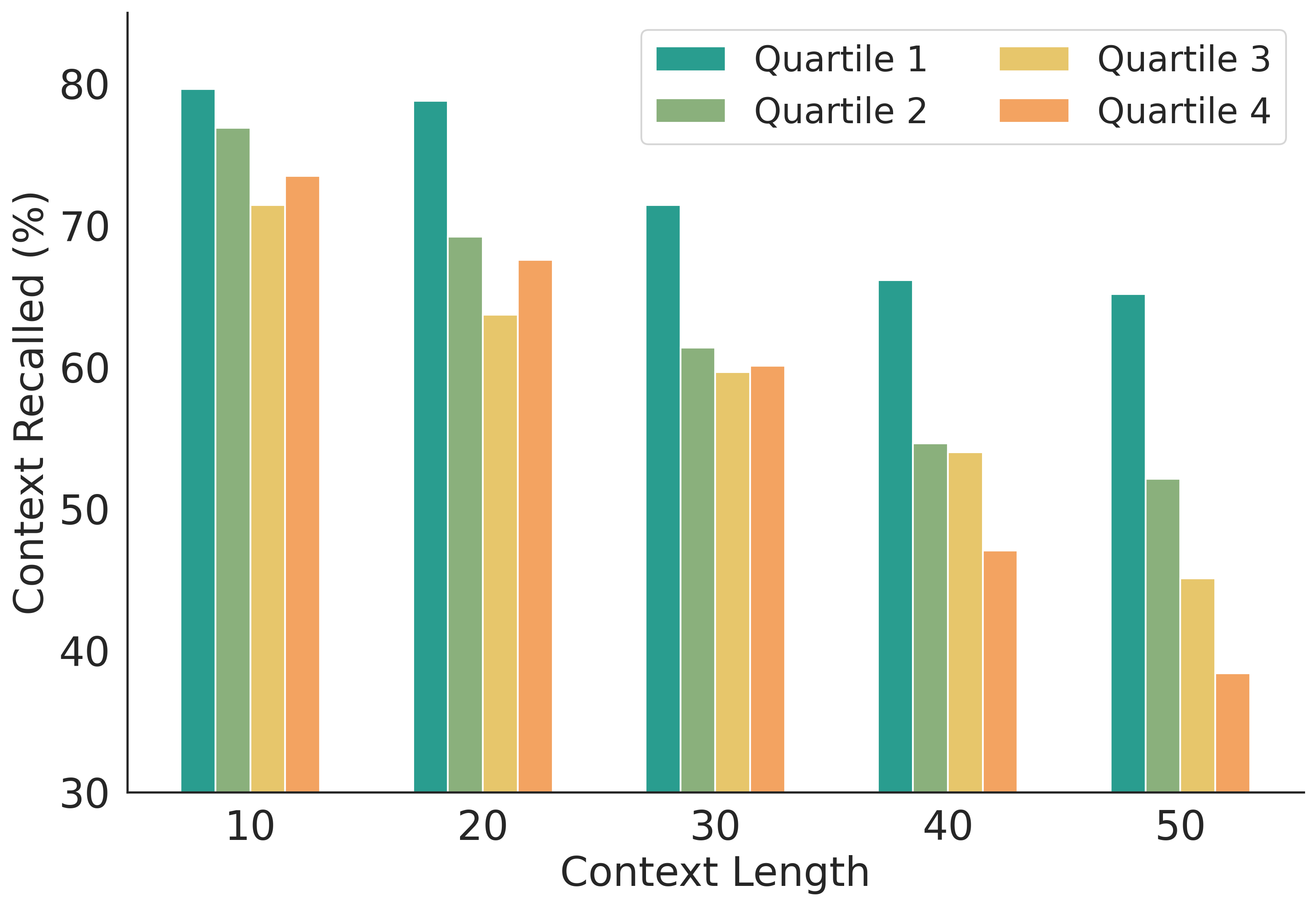}
    \caption{Llama 3.2 90B}
    \end{subfigure}
    \hfill
    \begin{subfigure}[t]{0.32\textwidth}
    \includegraphics[width=\textwidth]{figures/context_recall/fixed_y_range/es_llama323b.png}
    \caption{Llama 3.2 3B}
    \end{subfigure}
    \hfill
    \begin{subfigure}[t]{0.32\textwidth}
    \includegraphics[width=\textwidth]{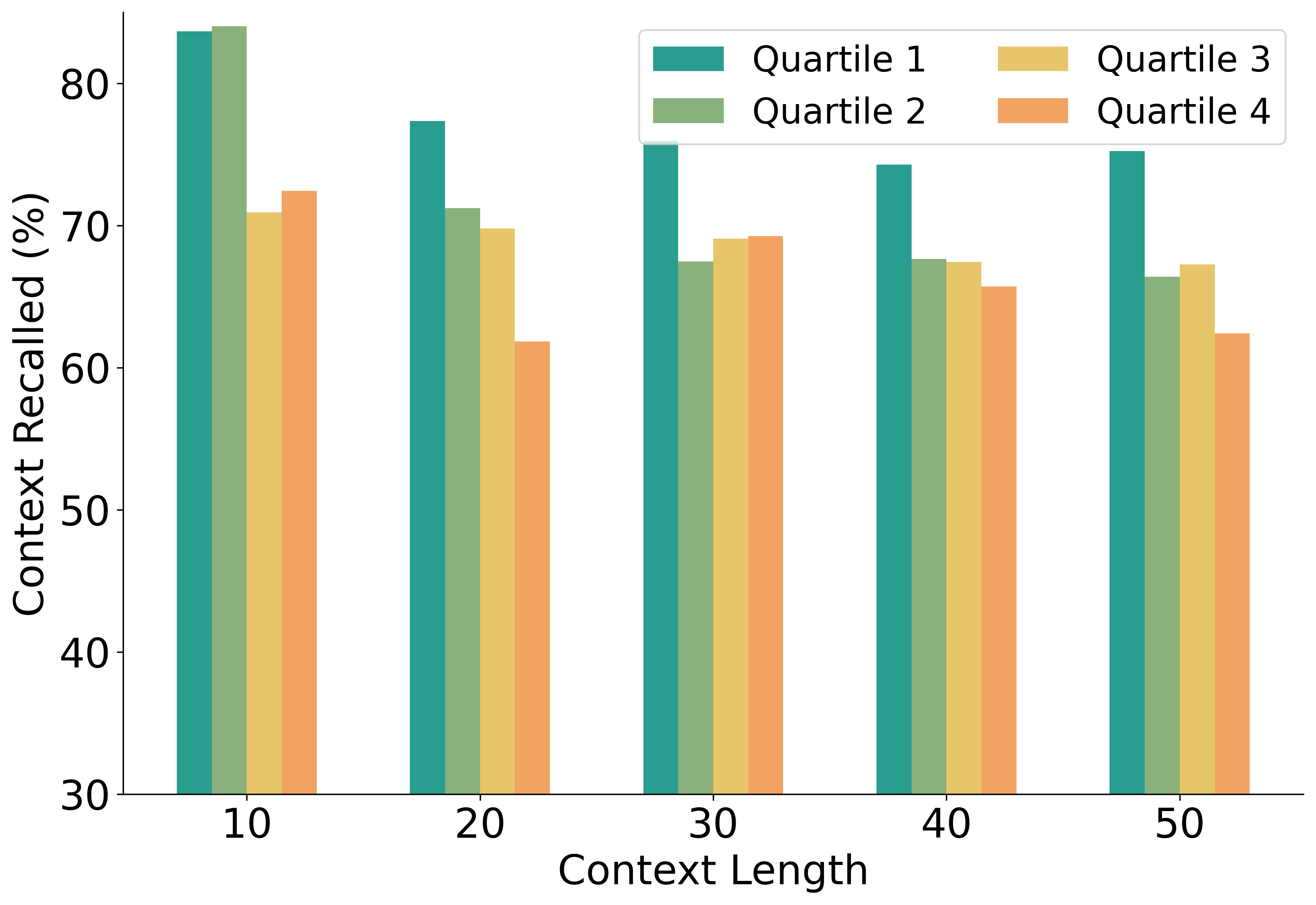}
    \caption{GPT-o3}
    \end{subfigure}
    \hfill
    \begin{subfigure}[t]{0.32\textwidth}
    \includegraphics[width=\textwidth]{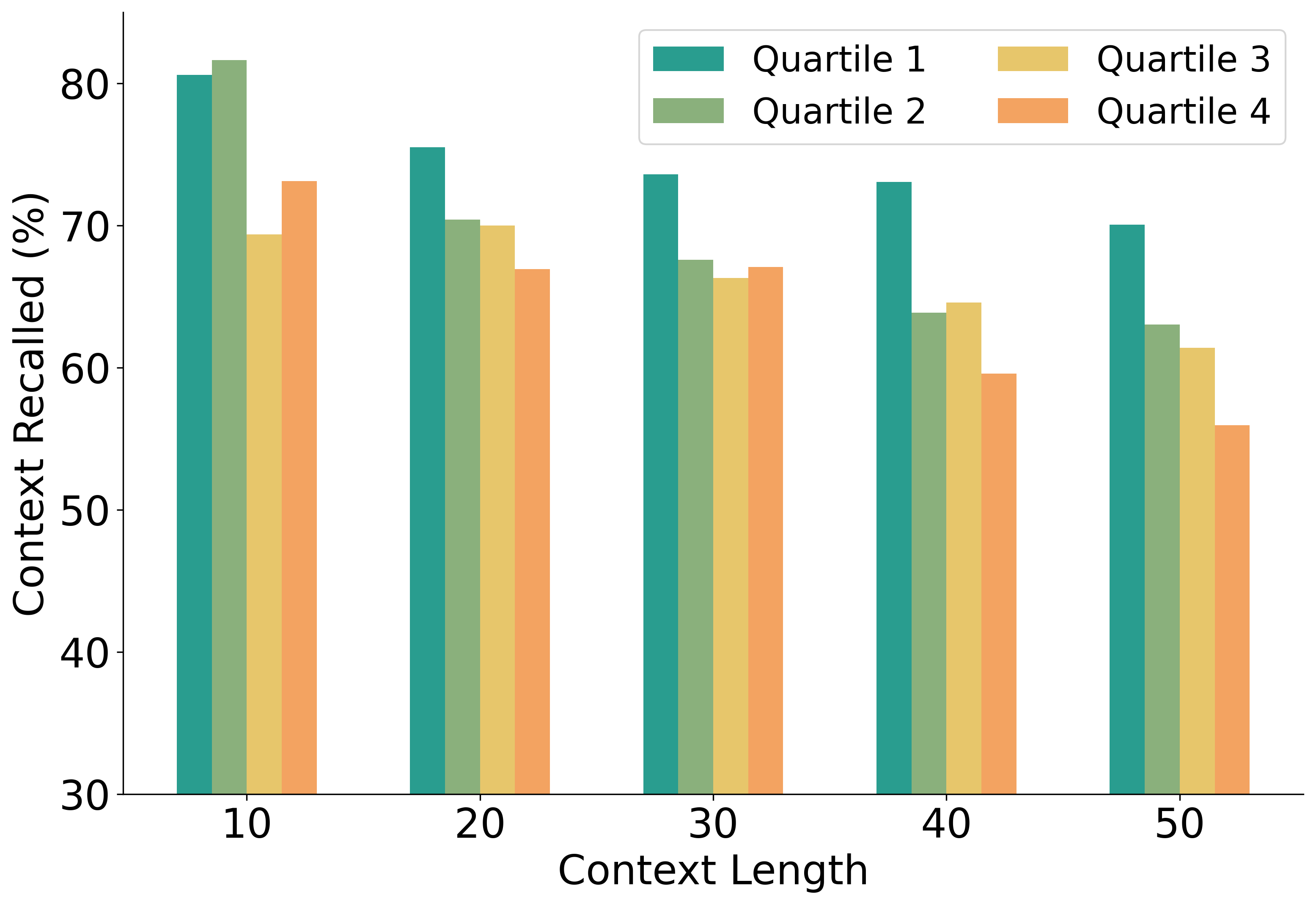}
    \caption{Qwen 3 235B}
    \end{subfigure}
    \caption{Context recall results for Spanish}
    \label{fig:context_recall_analysis_spanish}
\end{figure*}

\begin{figure*}[t!]
    \centering
    \begin{subfigure}[t]{0.32\textwidth}
    \includegraphics[width=\textwidth]{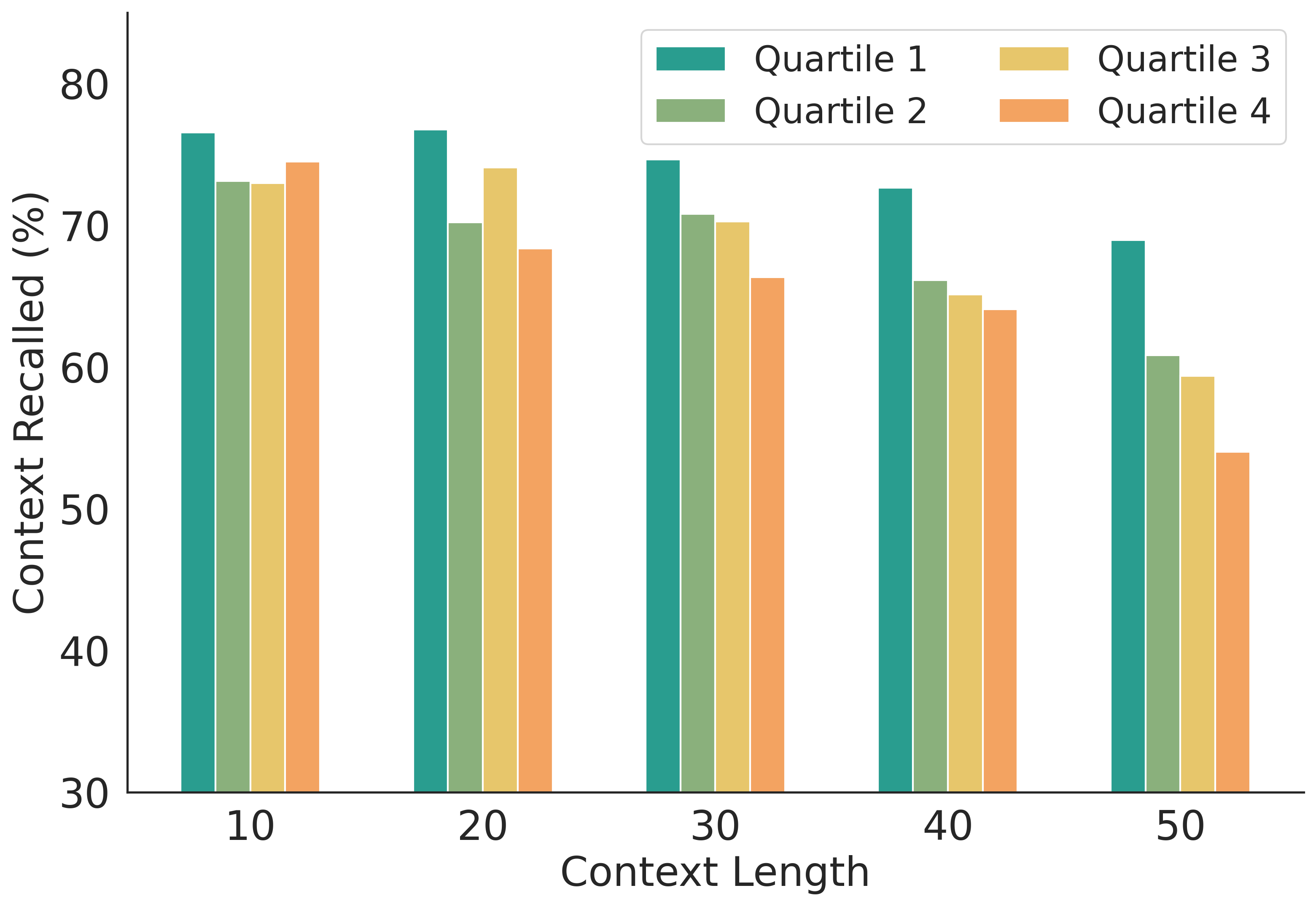}
    \caption{GPT-4o}
    \end{subfigure}
    \hfill
    \begin{subfigure}[t]{0.32\textwidth}
    \includegraphics[width=\textwidth]{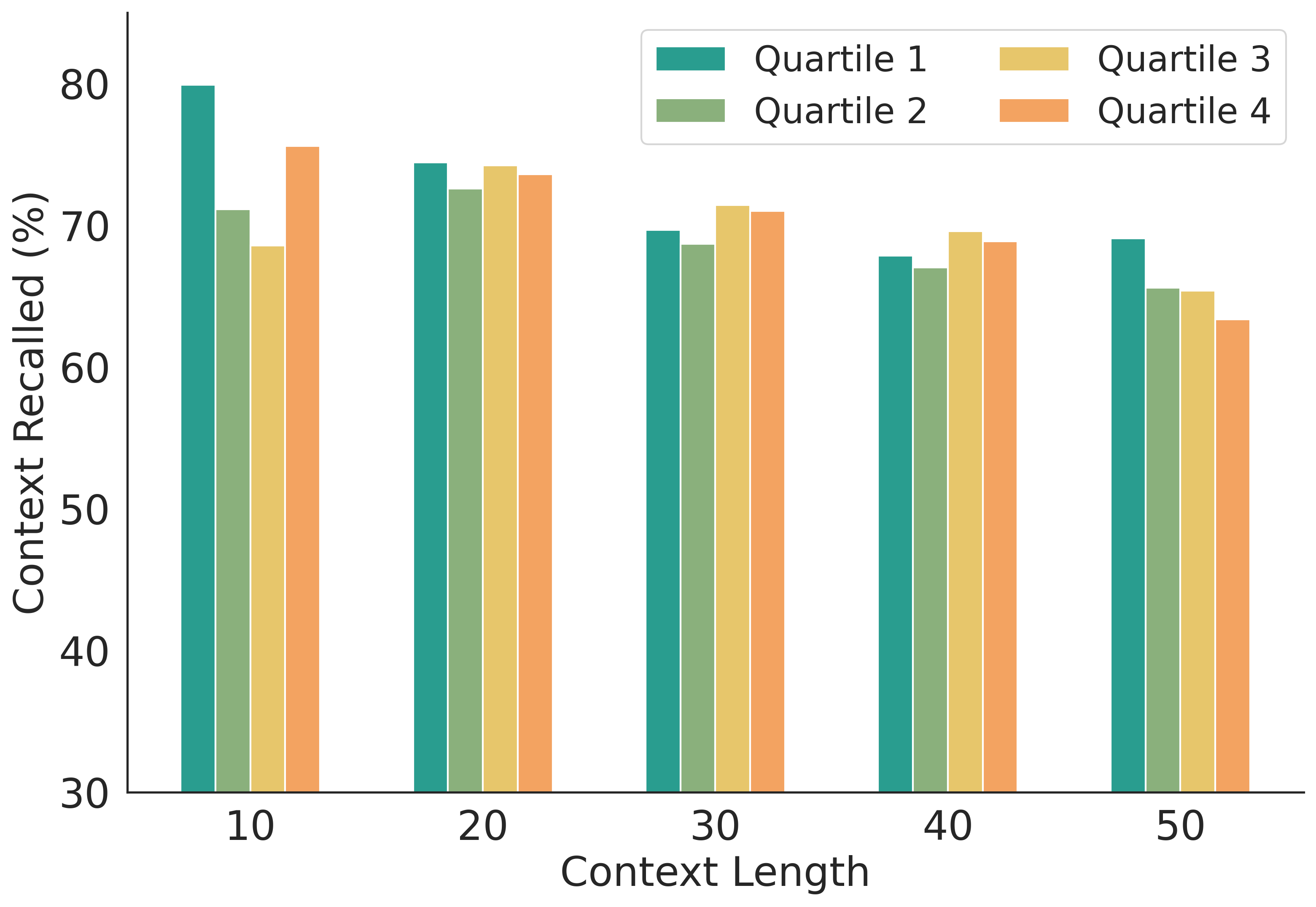}
    \caption{Gemini 1.5 Pro}
    \end{subfigure}
    \hfill
    \begin{subfigure}[t]{0.32\textwidth}
    \includegraphics[width=\textwidth]{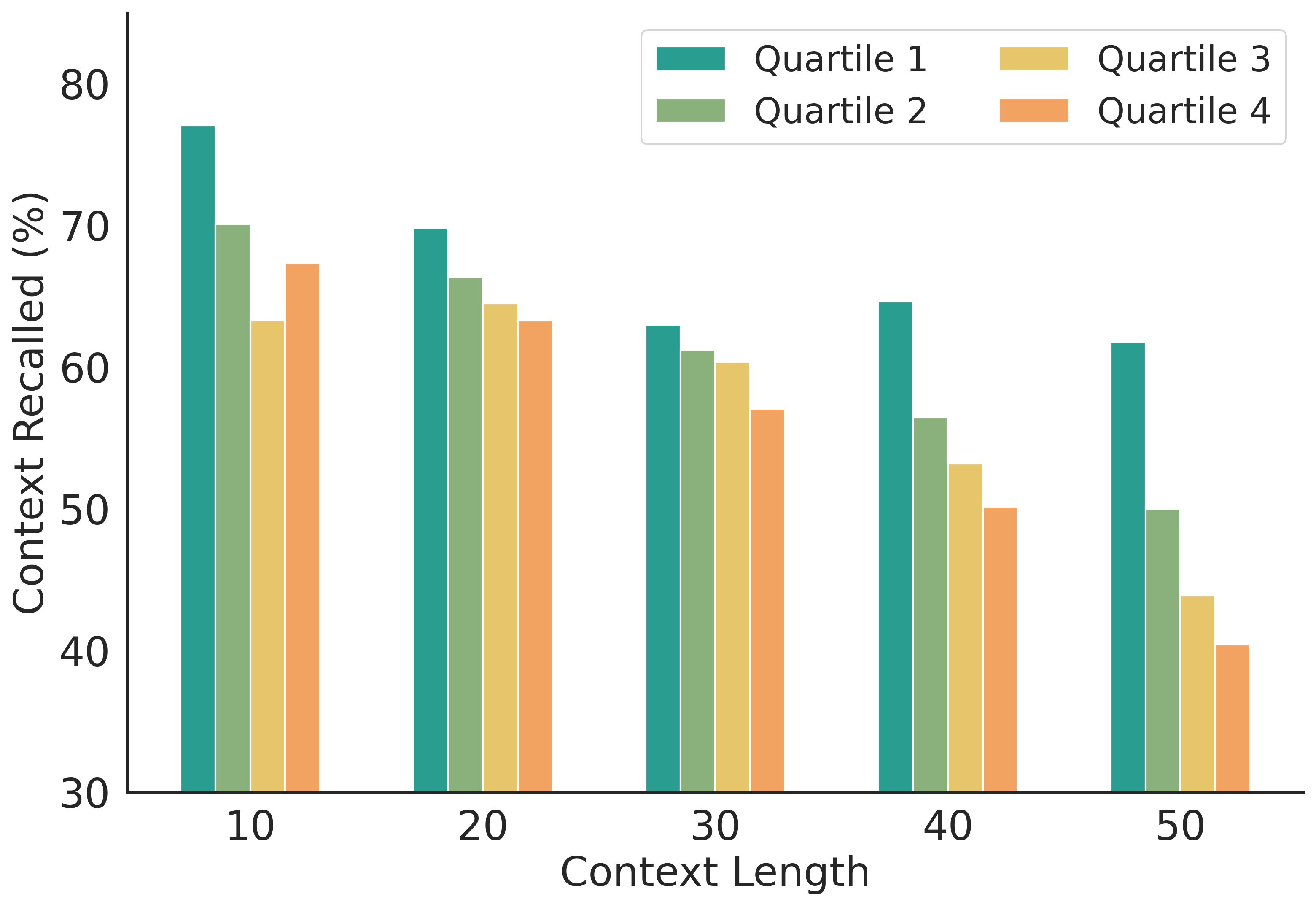}
    \caption{Llama 3.2 90B}
    \end{subfigure}
    \hfill
    \begin{subfigure}[t]{0.32\textwidth}
    \includegraphics[width=\textwidth]{figures/context_recall/fixed_y_range/da_llama323b.png}
    \caption{Llama 3.2 3B}
    \end{subfigure}
    \hfill
    \begin{subfigure}[t]{0.32\textwidth}
    \includegraphics[width=\textwidth]{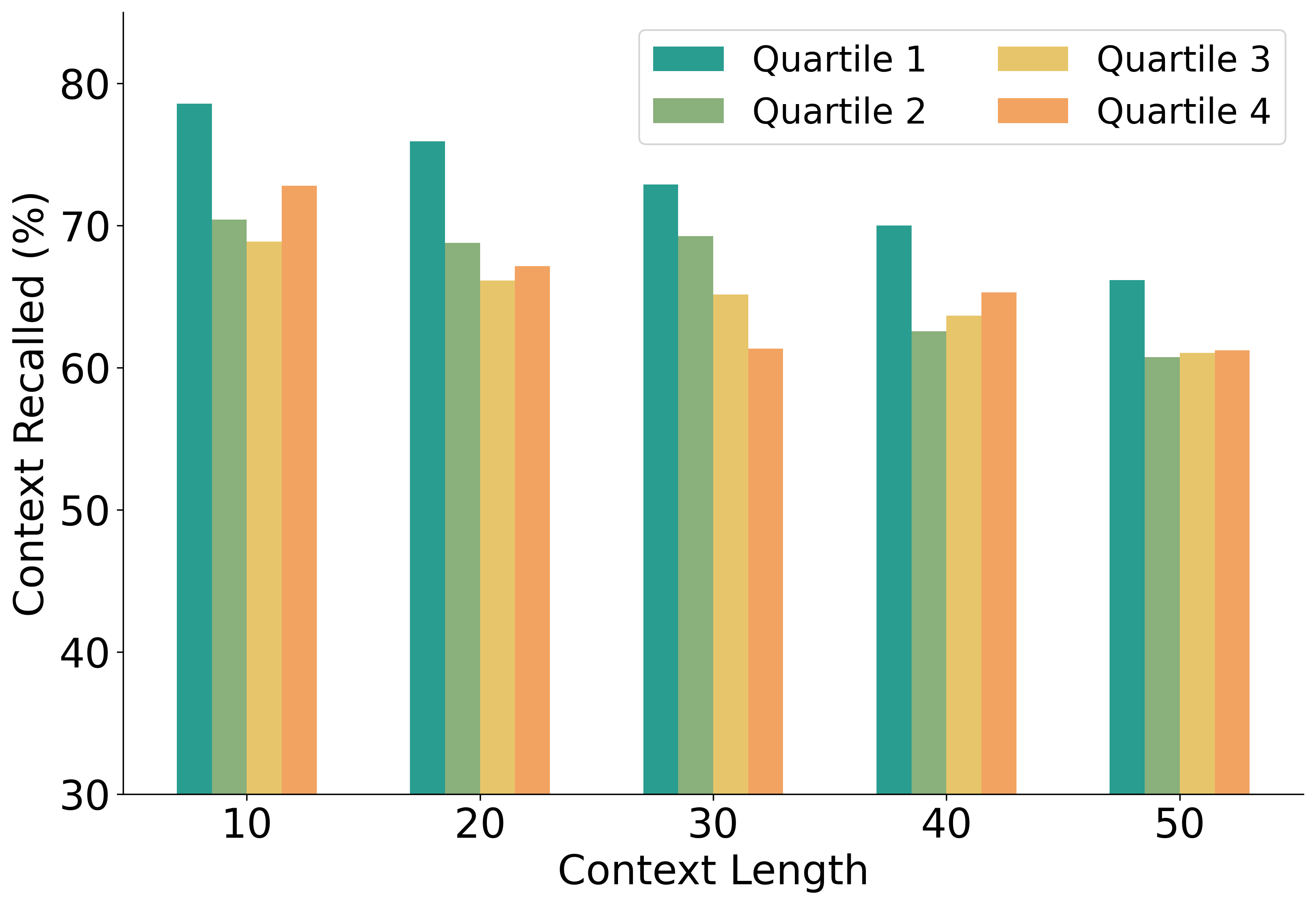}
    \caption{GPT-o3}
    \end{subfigure}
    \hfill
    \begin{subfigure}[t]{0.32\textwidth}
    \includegraphics[width=\textwidth]{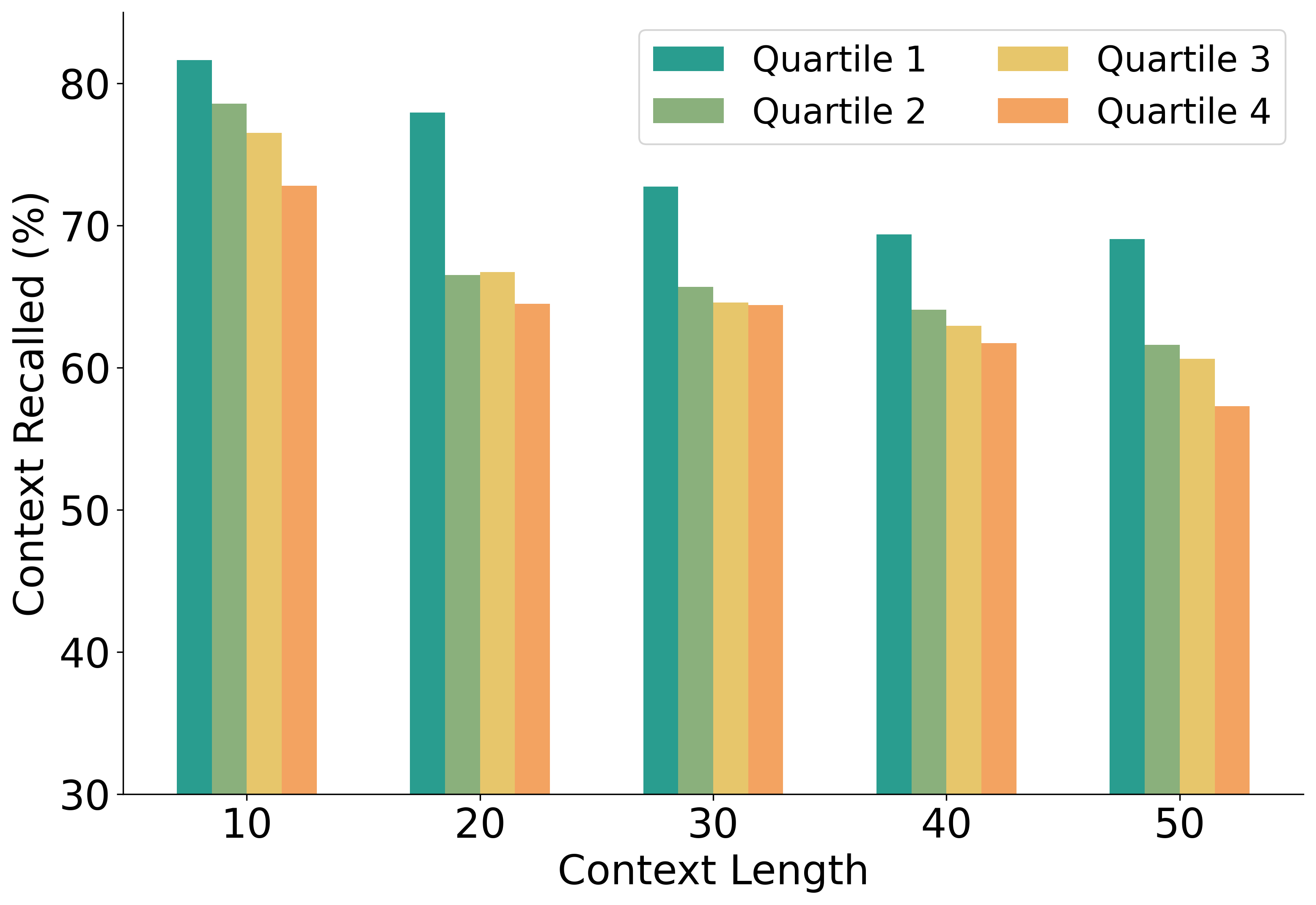}
    \caption{Qwen 3 235B}
    \end{subfigure}
    \caption{Context recall results for Danish}
    \label{fig:full_context_recall_analysis_danish}
\end{figure*}

\section{Full results of PK Distribution in Responses}
\label{app:full_results_position_of_0}
This section has full results of PK distribution in responses in English, Spanish, and Danish, respectively, for each model (Figure \ref{fig:position_of_0_analysis_english}, \ref{fig:position_of_0_analysis_spanish} and \ref{fig:full_position_of_0_analysis_danish}).

\begin{figure*}[t!]
    \centering
    \begin{subfigure}[t]{0.32\textwidth}
    \includegraphics[width=1\textwidth]{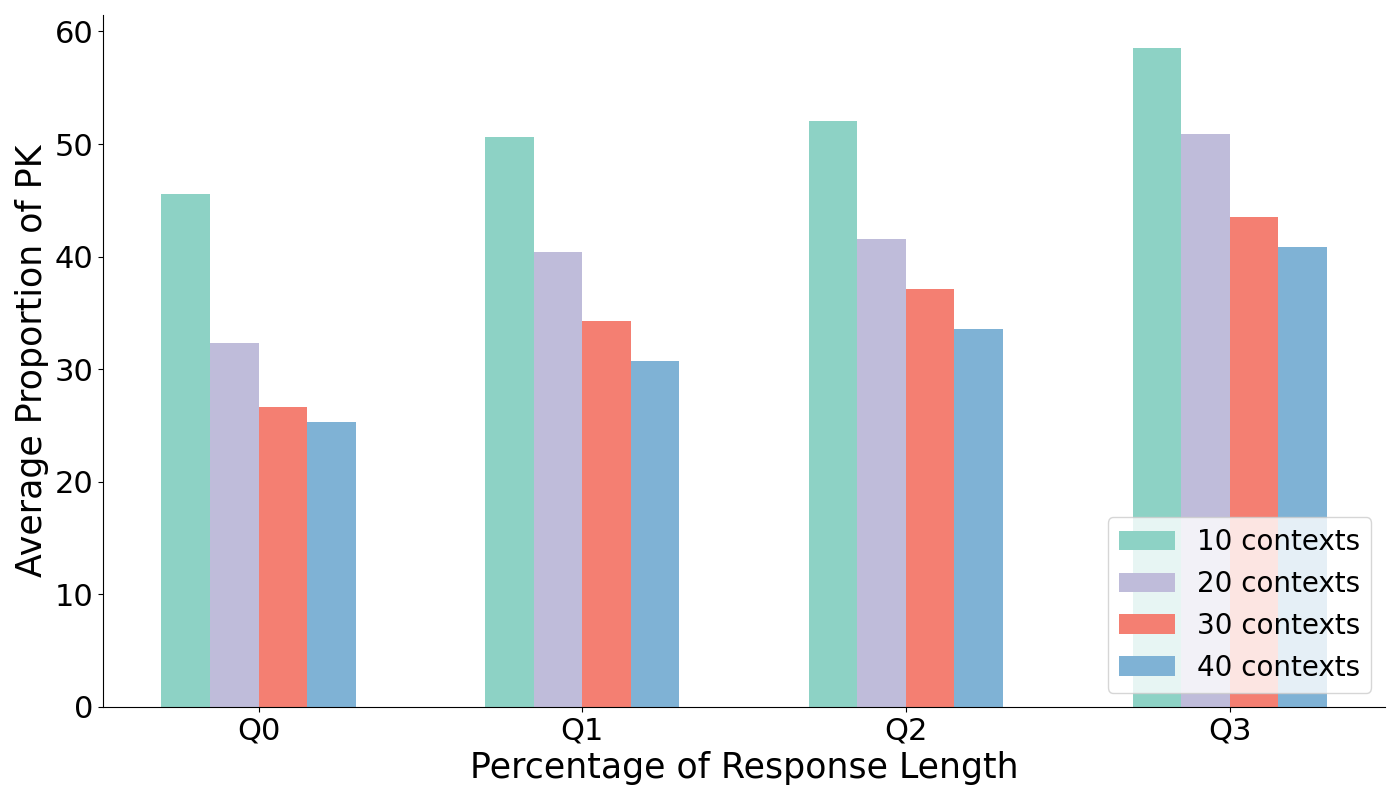}
    \caption{GPT-4o}
    \end{subfigure}
    \hfill
    \begin{subfigure}[t]{0.32\textwidth}
    \includegraphics[width=\textwidth]{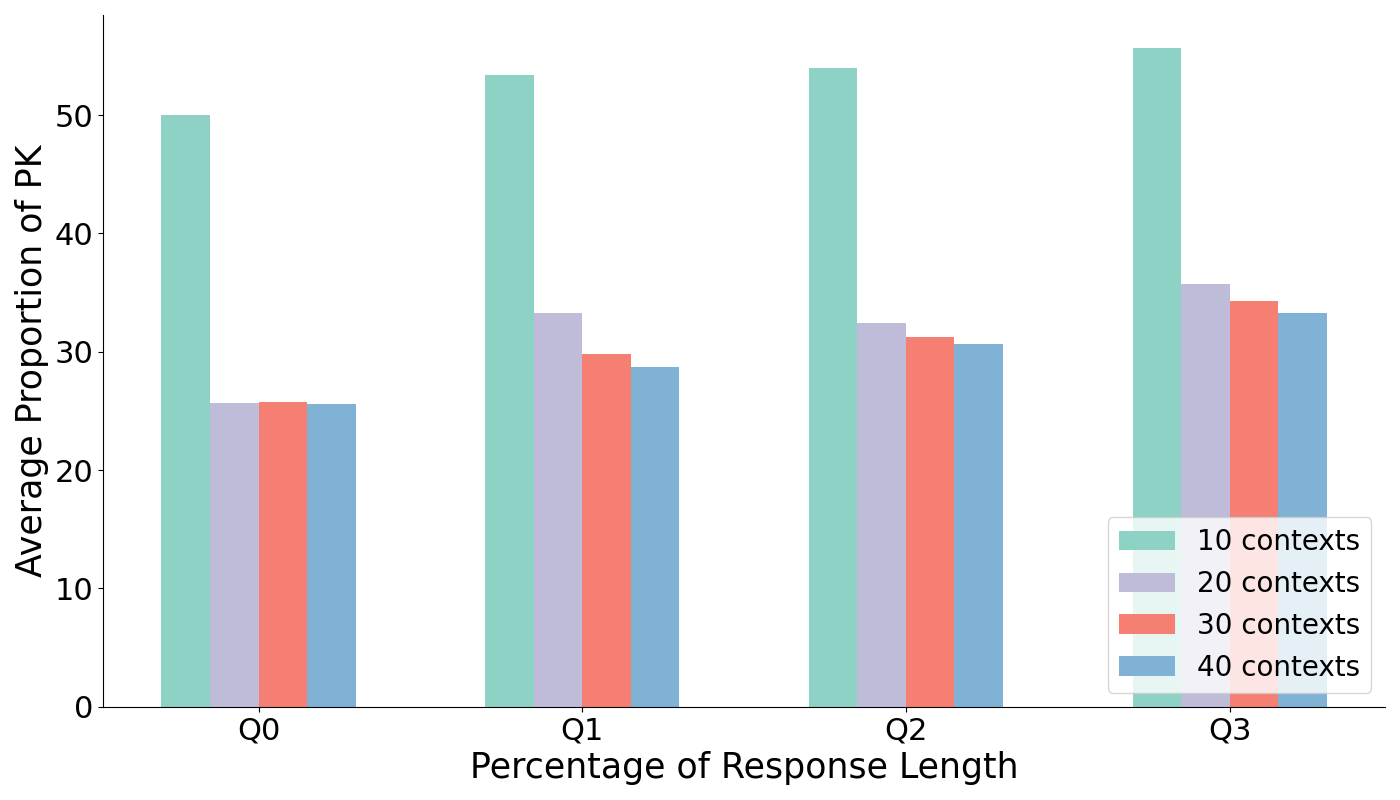}
    \caption{Gemini 1.5 Pro}
    \end{subfigure}
    \hfill
    \begin{subfigure}[t]{0.32\textwidth}
    \includegraphics[width=\textwidth]{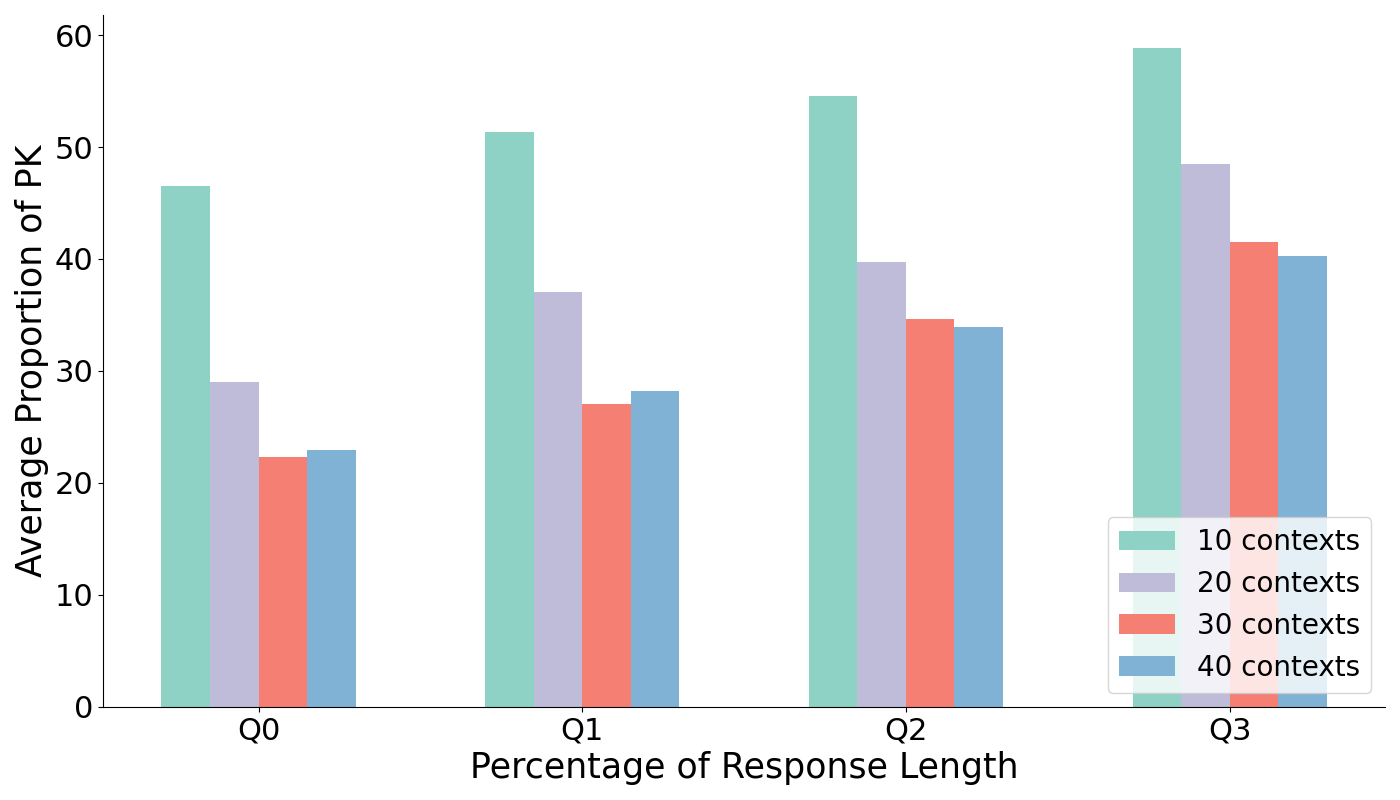}
    \caption{Llama 3.2 90B}
    \end{subfigure}
    \hfill
    \begin{subfigure}[t]{0.32\textwidth}
    \includegraphics[width=\textwidth]{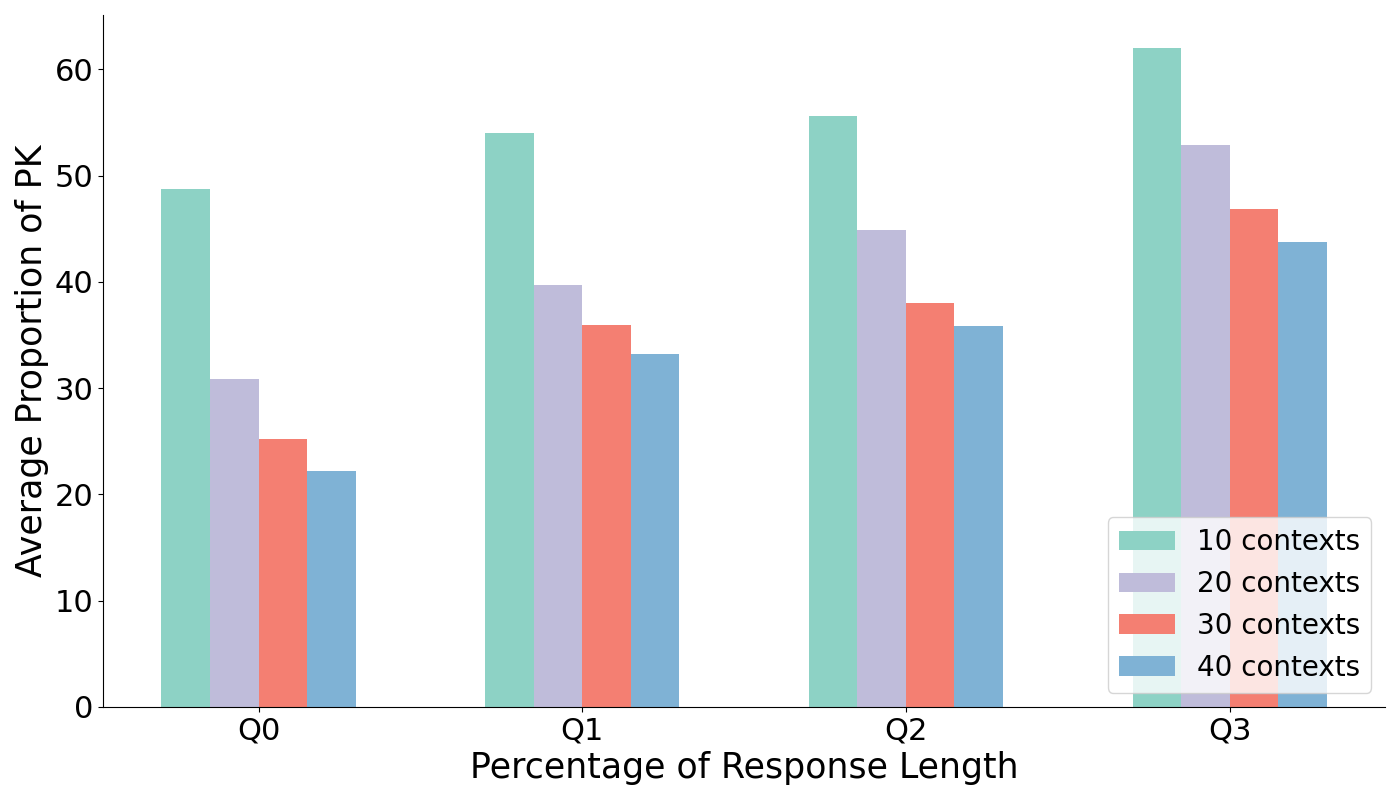}
    \caption{Llama 3.2 3B}
    \end{subfigure}
    \hfill
    \begin{subfigure}[t]{0.32\textwidth}
    \includegraphics[width=\textwidth]{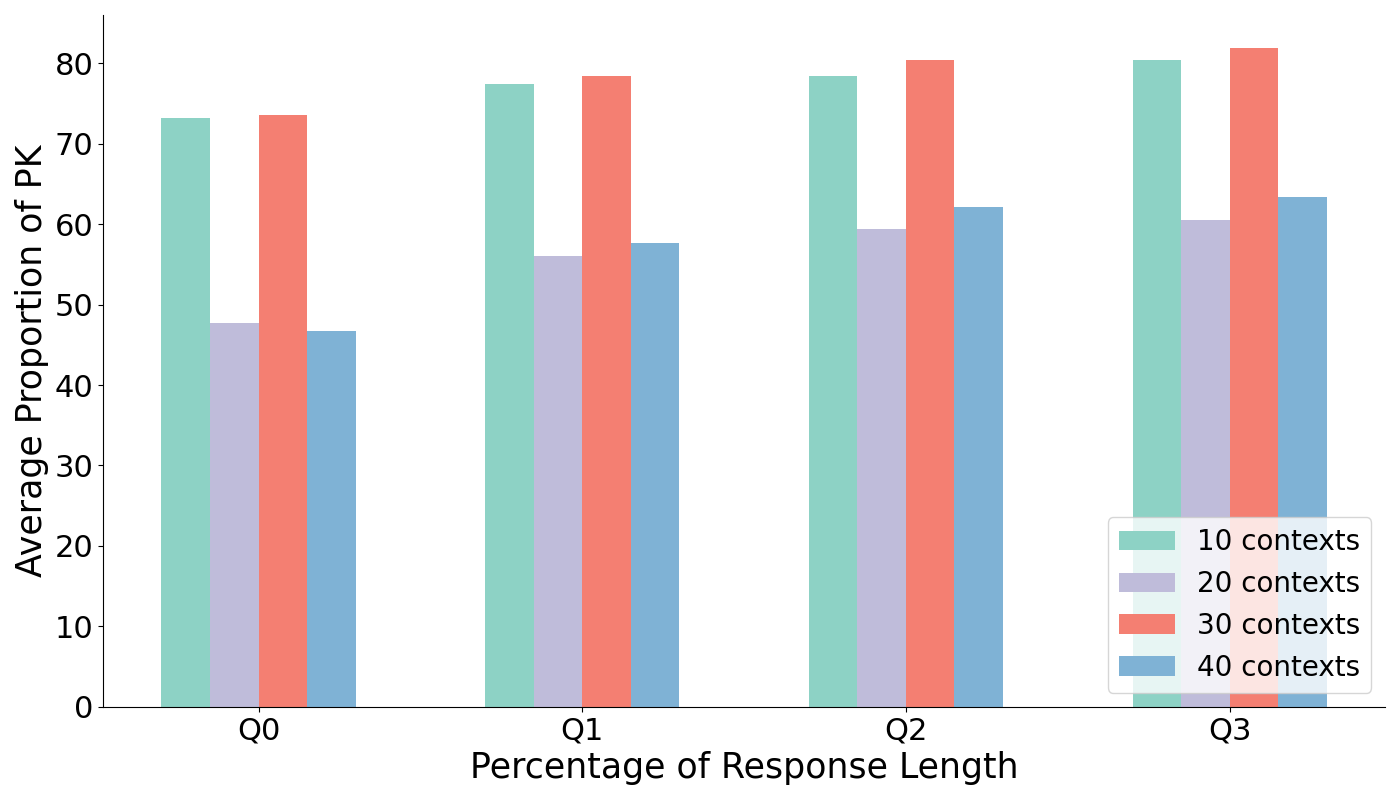}
    \caption{GPT-o3}
    \end{subfigure}
    \hfill
    \begin{subfigure}[t]{0.32\textwidth}
    \includegraphics[width=\textwidth]{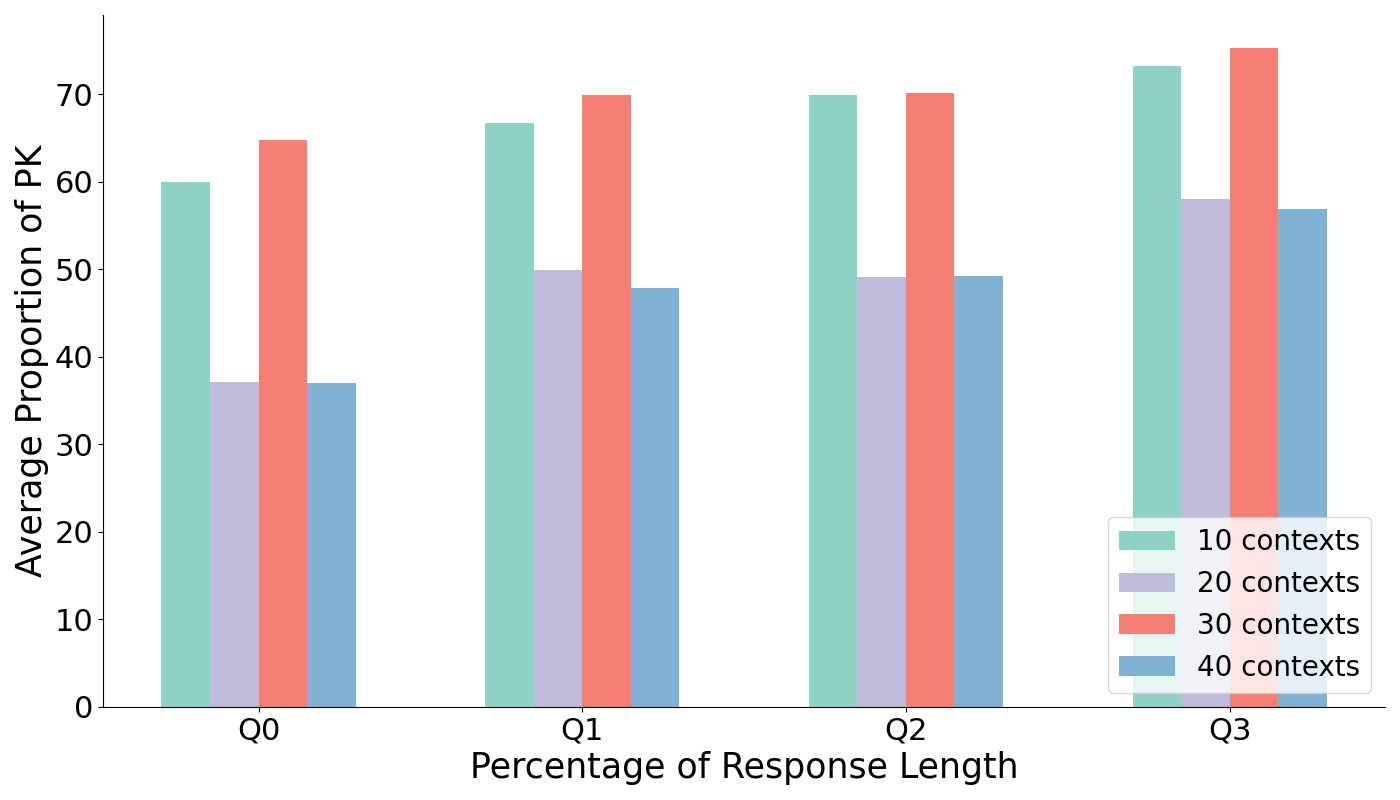}
    \caption{Qwen 3 235B}
    \end{subfigure}
    \caption{PK distribution in English responses}
    \label{fig:position_of_0_analysis_english}
\end{figure*}

\begin{figure*}[t!]
    \centering
    \begin{subfigure}[t]{0.32\textwidth}
    \includegraphics[width=\textwidth]{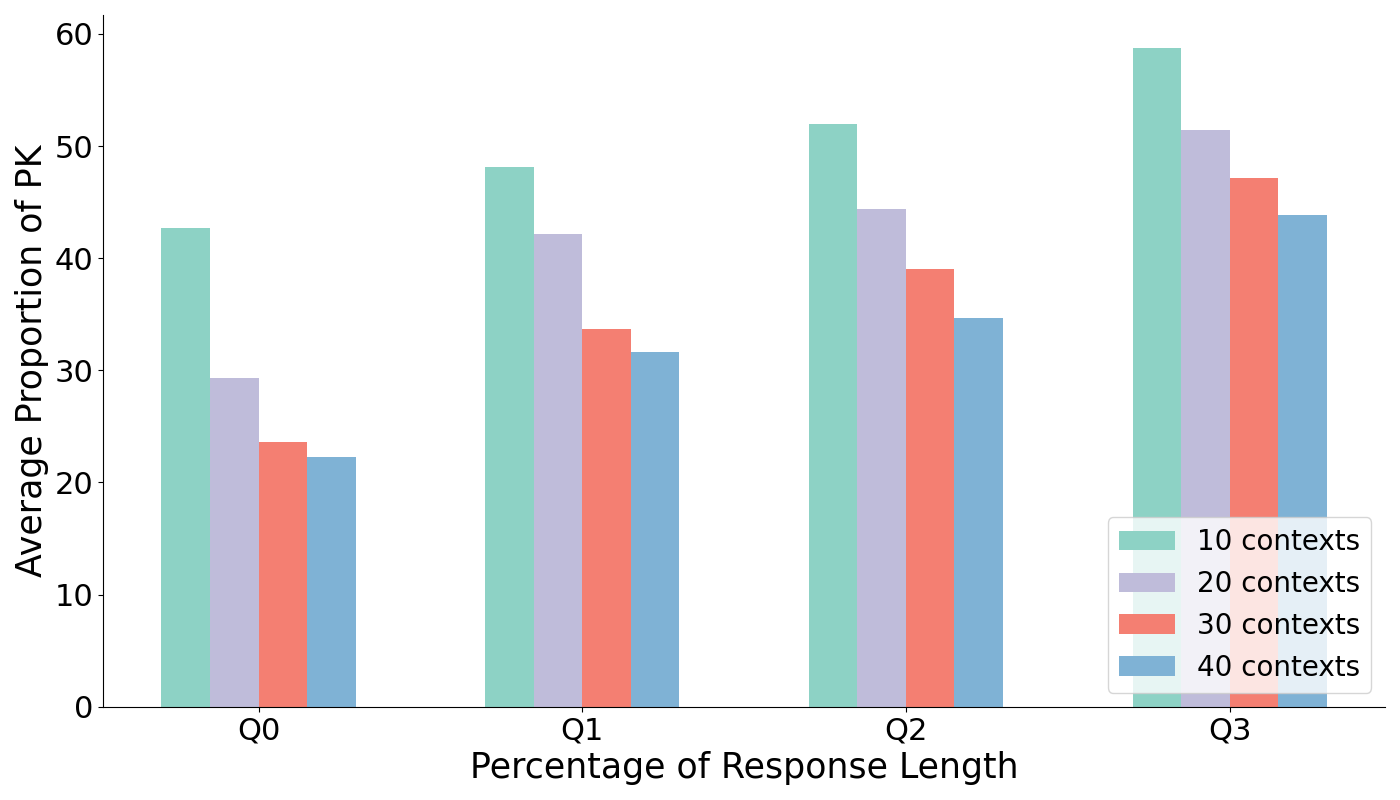}
    \caption{GPT-4o}
    \end{subfigure}
    \hfill
    \begin{subfigure}[t]{0.32\textwidth}
    \includegraphics[width=\textwidth]{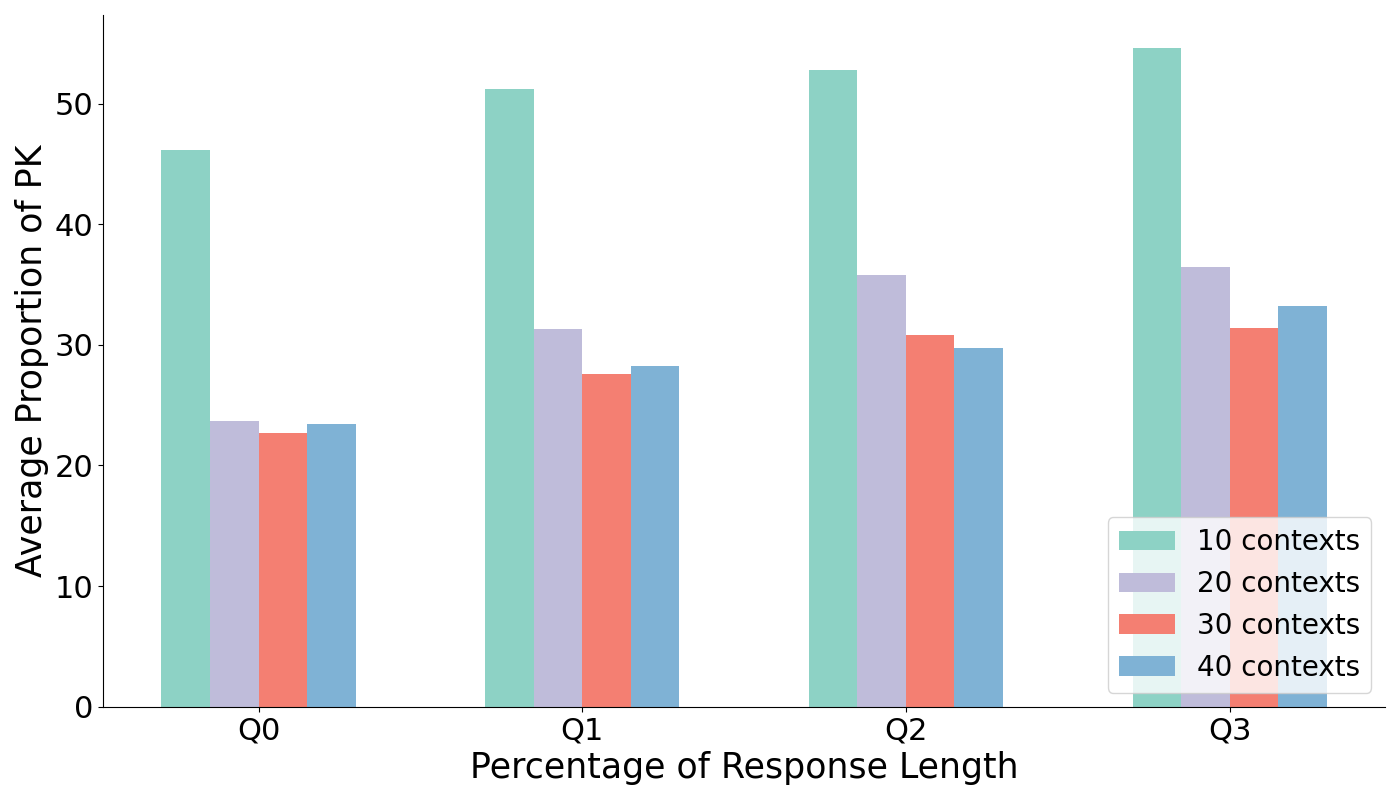}
    \caption{Gemini 1.5 Pro}
    \end{subfigure}
    \hfill
    \begin{subfigure}[t]{0.32\textwidth}
    \includegraphics[width=\textwidth]{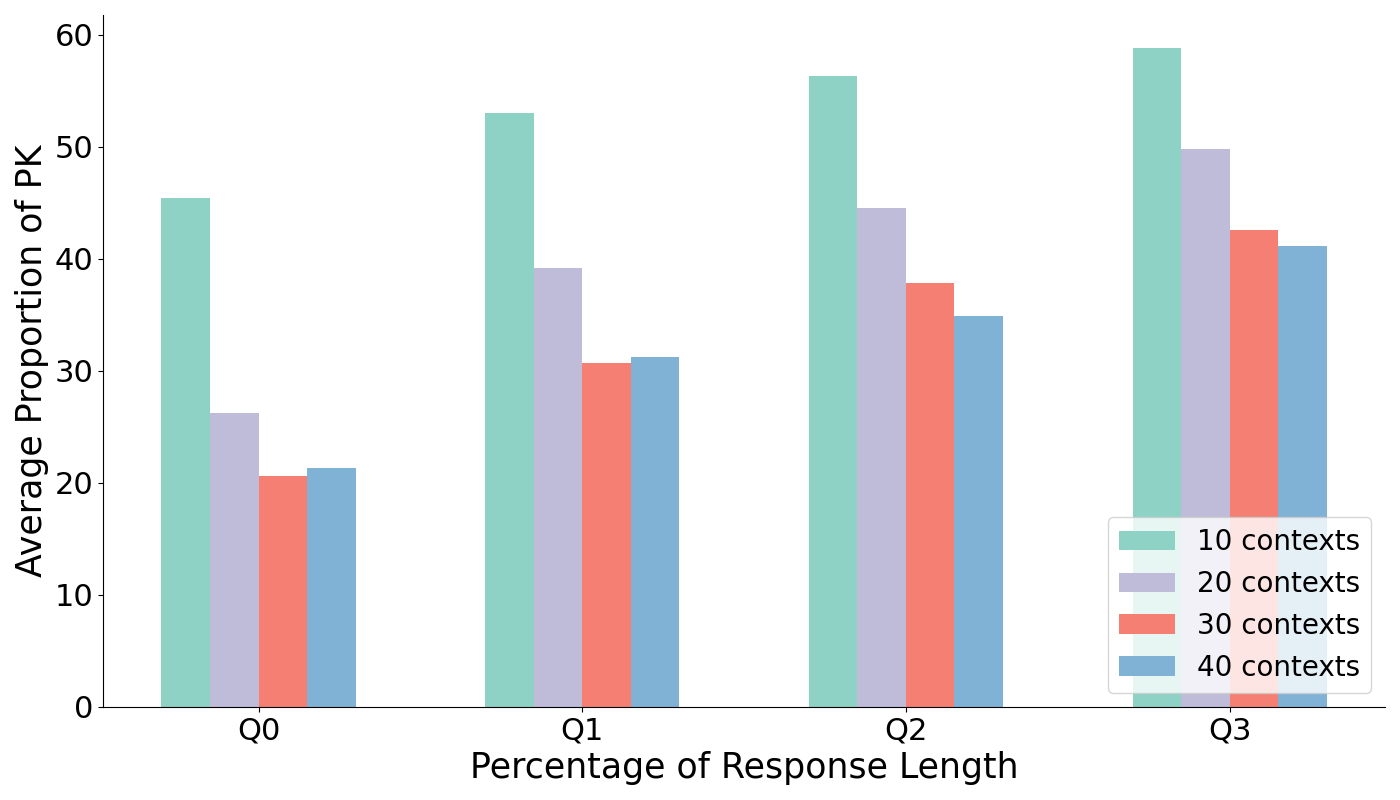}
    \caption{Llama 3.2 90B}
    \end{subfigure}
    \hfill
    \begin{subfigure}[t]{0.32\textwidth}
    \includegraphics[width=\textwidth]{figures/position_of_0/pk_pattern/es_llama323b.png}
    \caption{Llama 3.2 3B}
    \end{subfigure}
        \hfill
    \begin{subfigure}[t]{0.32\textwidth}
    \includegraphics[width=\textwidth]{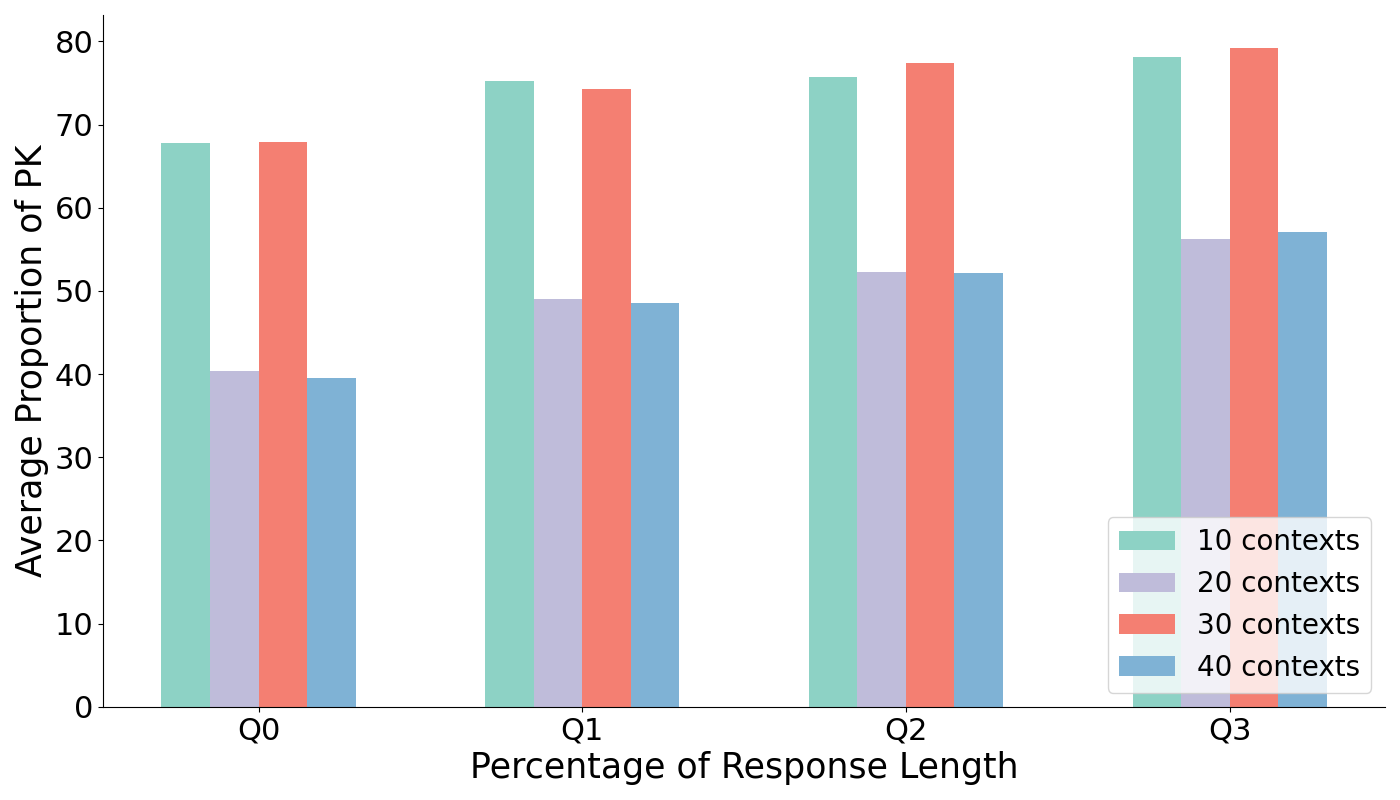}
    \caption{GPT-o3}
    \end{subfigure}
    \hfill
    \begin{subfigure}[t]{0.32\textwidth}
    \includegraphics[width=\textwidth]{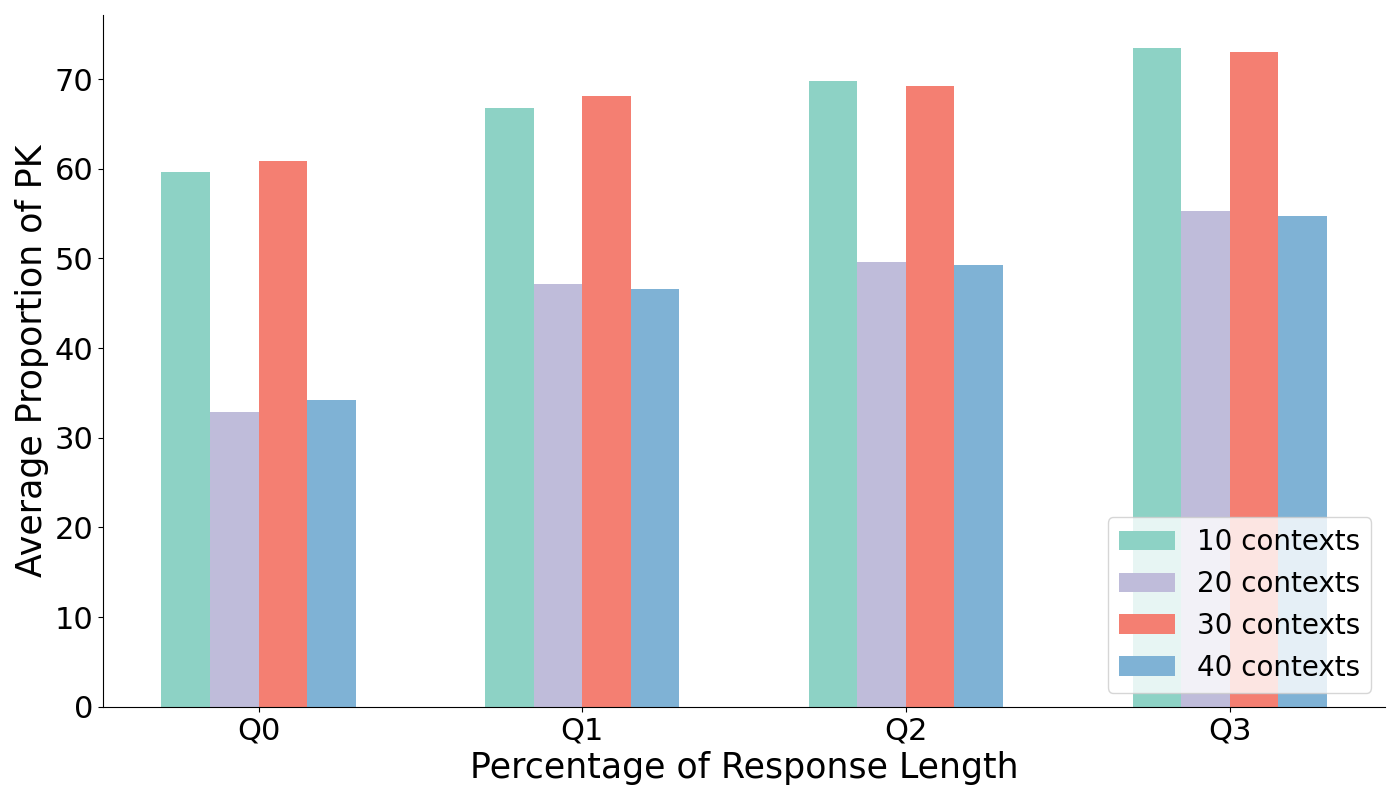}
    \caption{Qwen 3 235B}
    \end{subfigure}
    \caption{PK distribution in Spanish responses}
    \label{fig:position_of_0_analysis_spanish}
\end{figure*}

\begin{figure*}[t!]
    \centering
    \begin{subfigure}[t]{0.32\textwidth}
    \includegraphics[width=\textwidth]{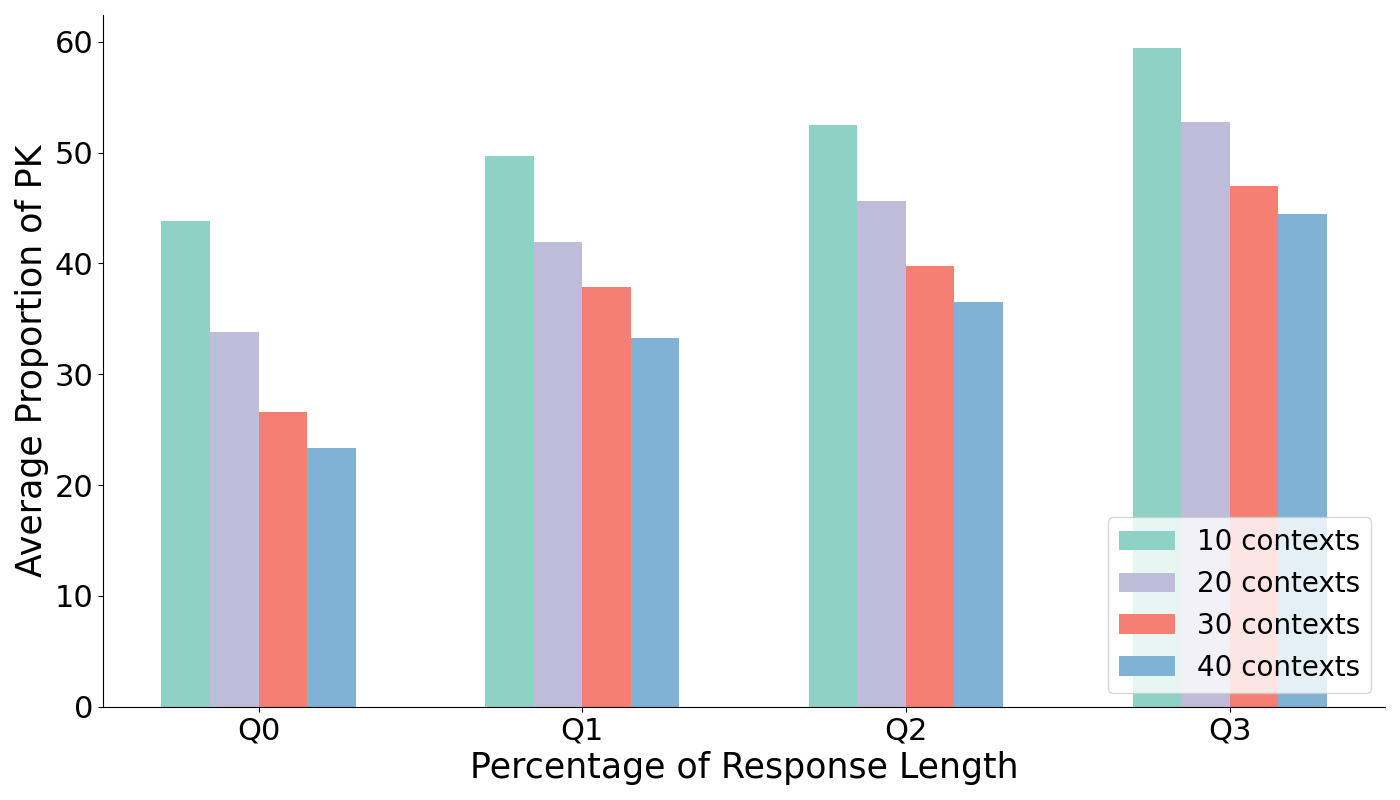}
    \caption{GPT-4o}
    \end{subfigure}
    \hfill
    \begin{subfigure}[t]{0.32\textwidth}
    \includegraphics[width=\textwidth]{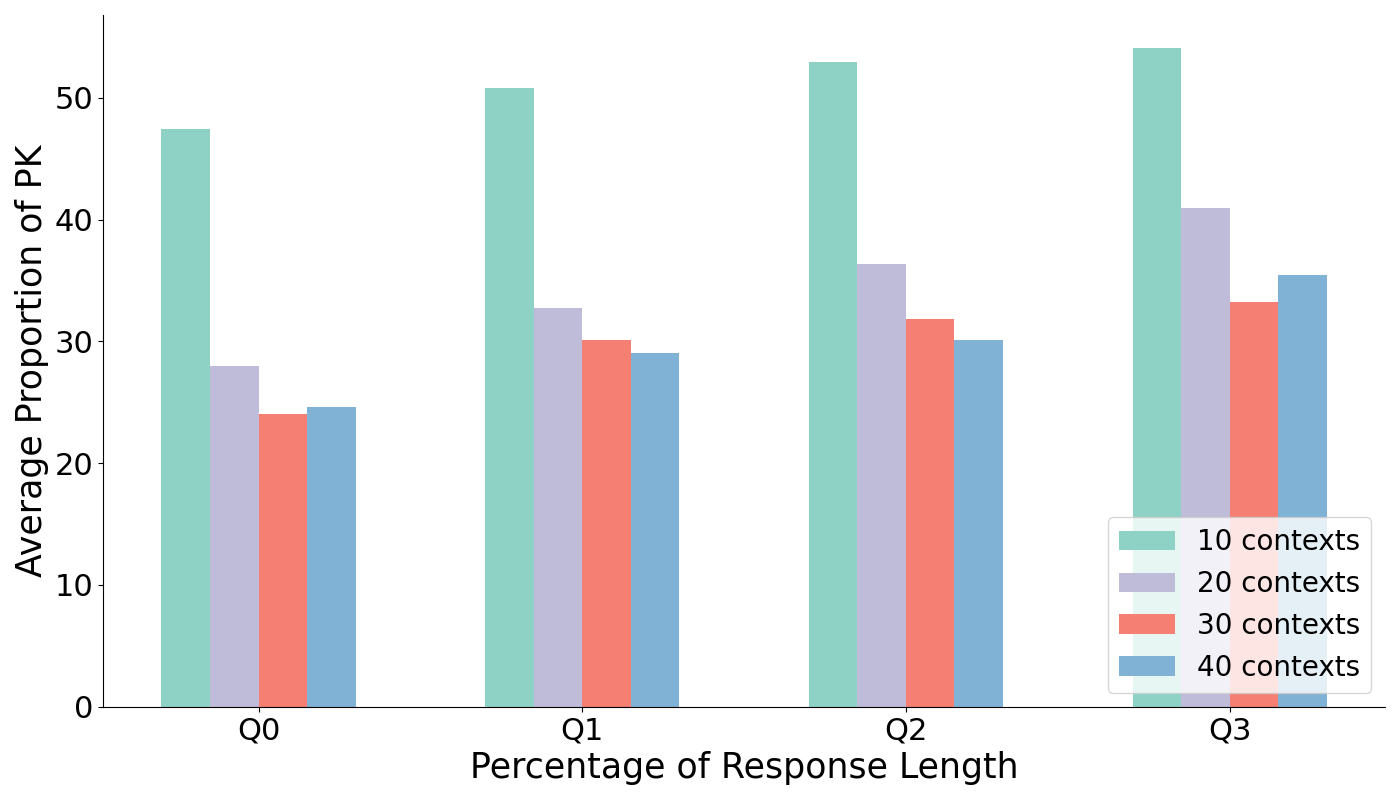}
    \caption{Gemini 1.5 Pro}
    \end{subfigure}
    \hfill
    \begin{subfigure}[t]{0.32\textwidth}
    \includegraphics[width=\textwidth]{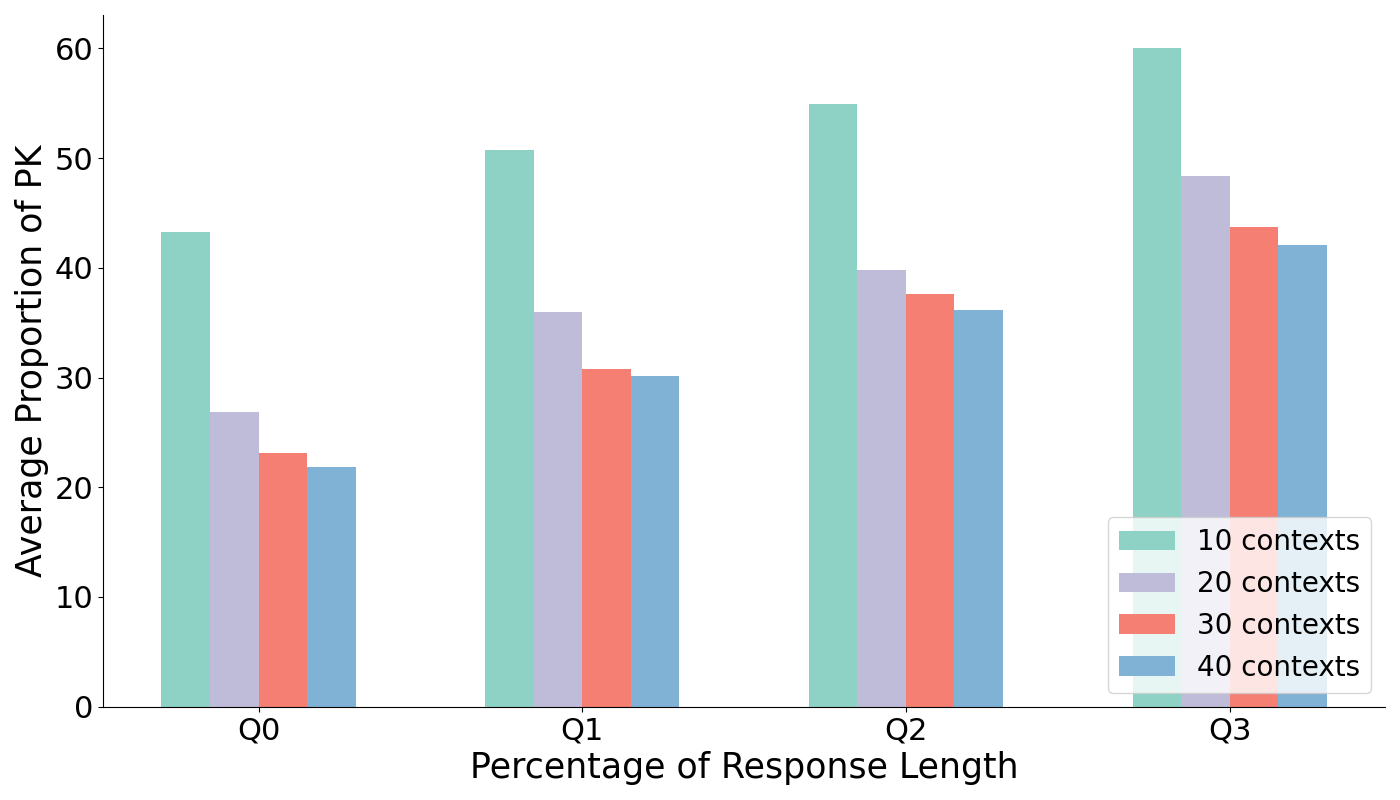}
    \caption{Llama 3.2 90B}
    \end{subfigure}
    \hfill
    \begin{subfigure}[t]{0.32\textwidth}
    \includegraphics[width=\textwidth]{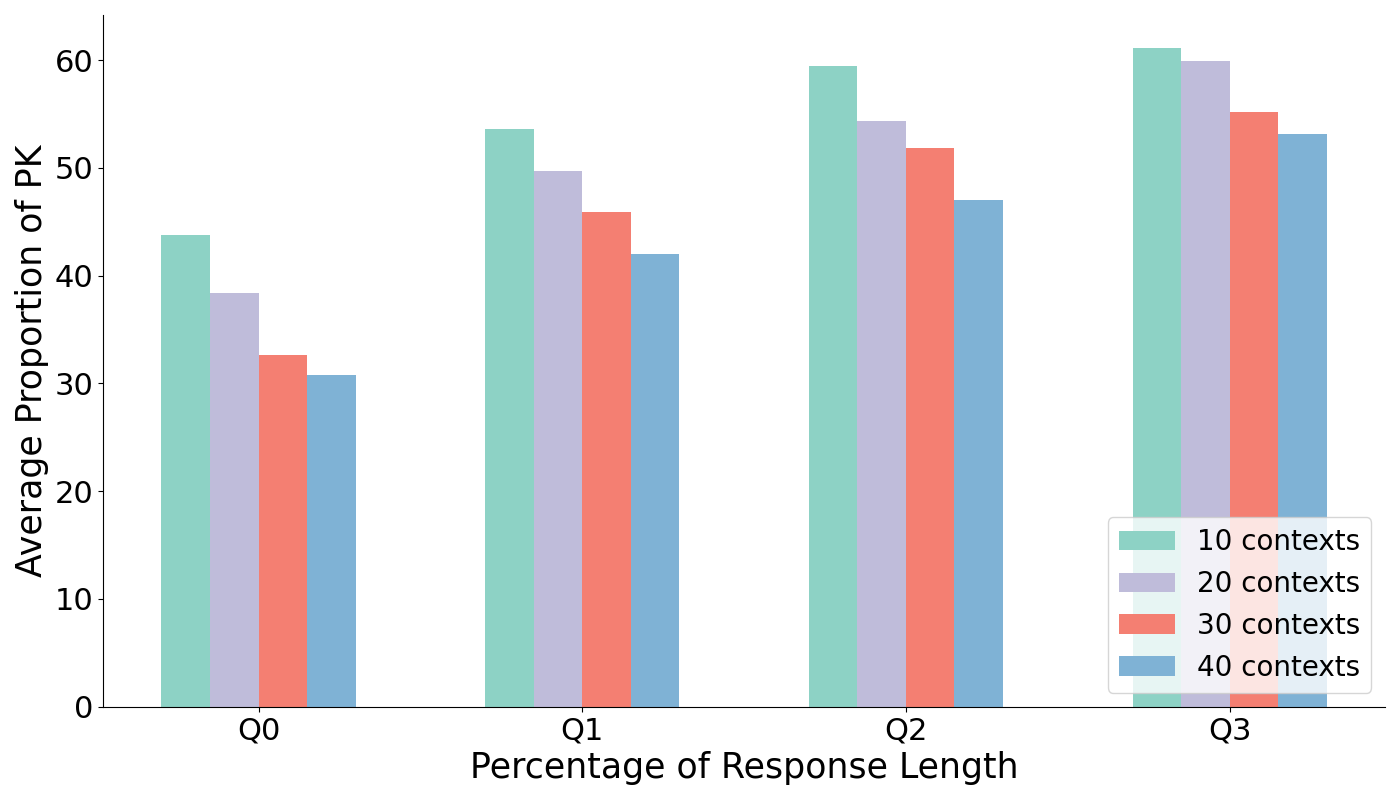}
    \caption{Llama 3.2 3B}
    \end{subfigure}
        \hfill
    \begin{subfigure}[t]{0.32\textwidth}
    \includegraphics[width=\textwidth]{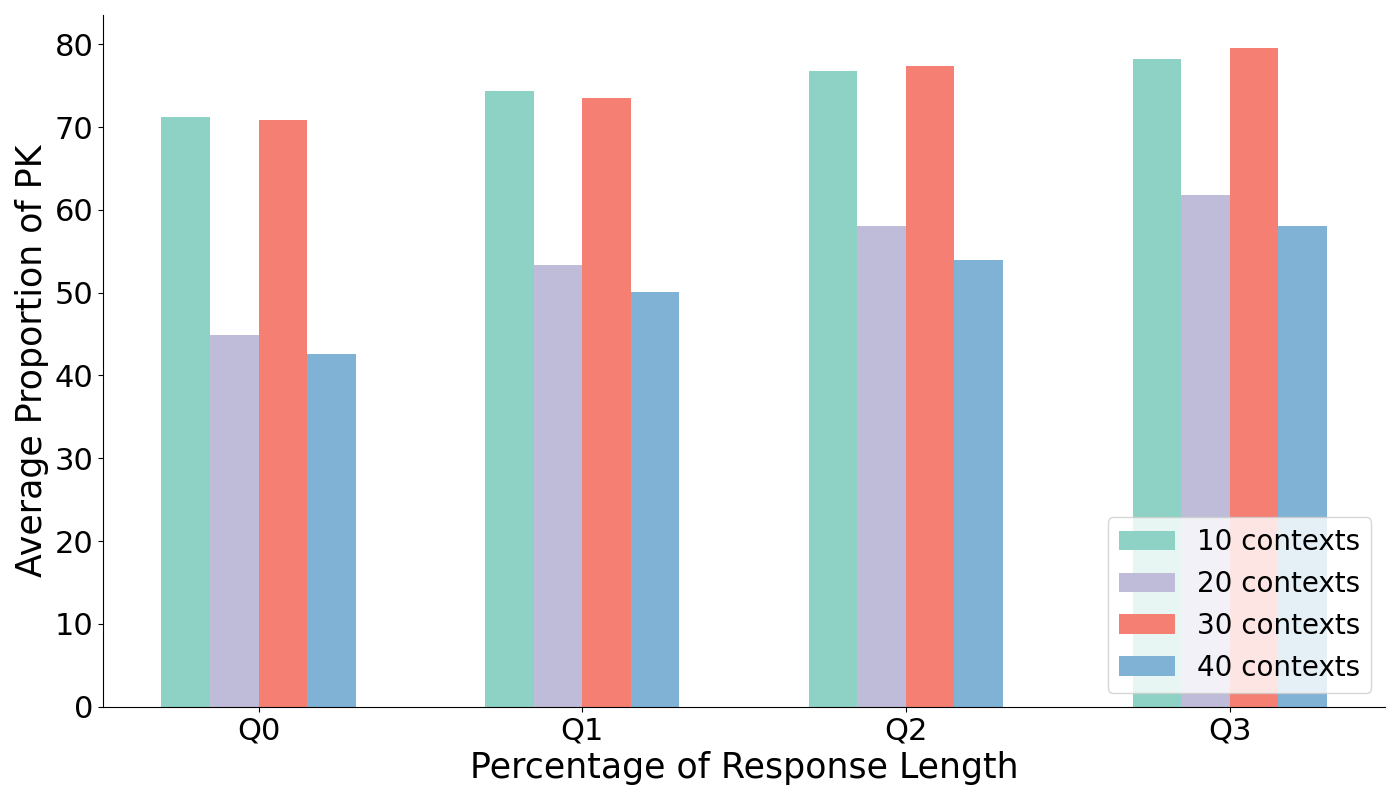}
    \caption{GPT-o3}
    \end{subfigure}
    \hfill
    \begin{subfigure}[t]{0.32\textwidth}
    \includegraphics[width=\textwidth]{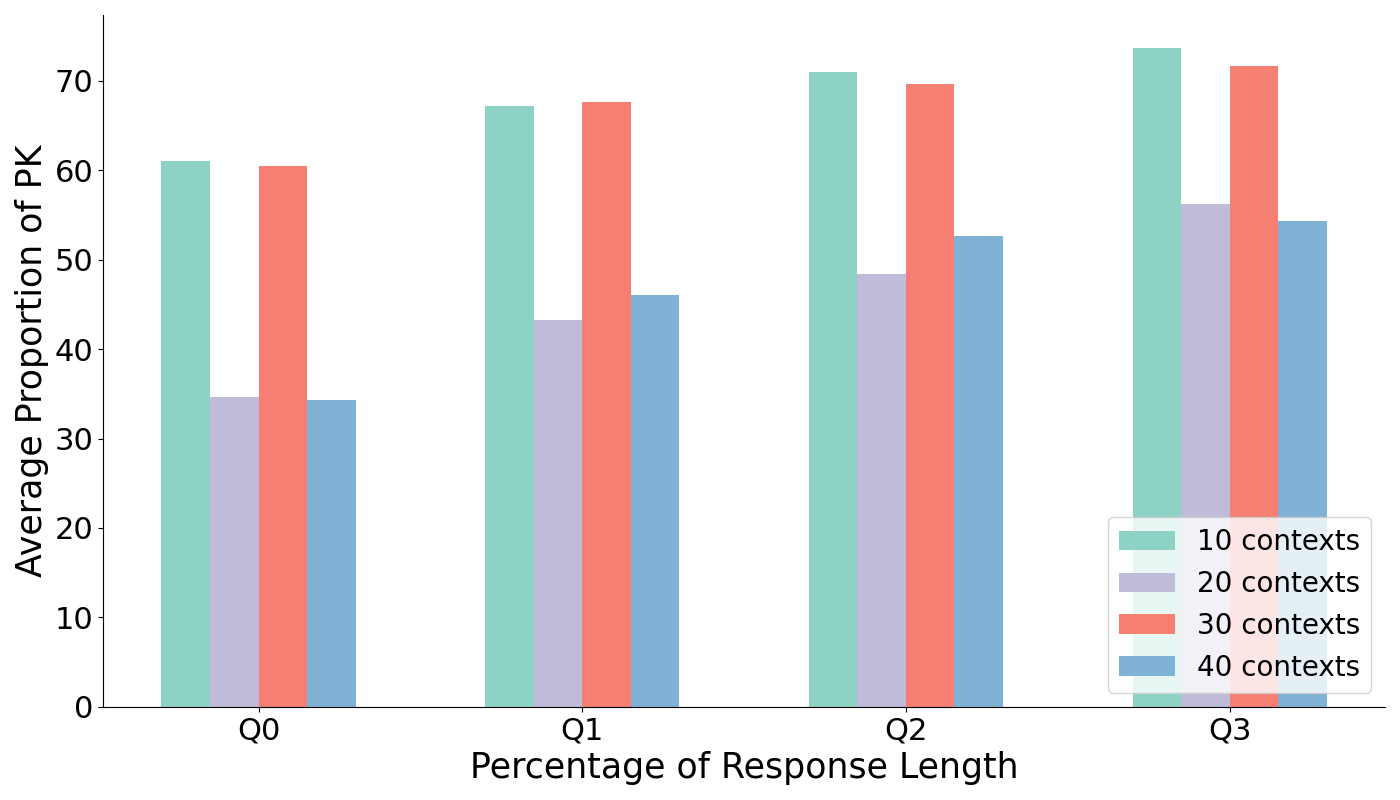}
    \caption{Qwen 3 235B}
    \end{subfigure}
    \caption{PK distribution in Danish responses}
    \label{fig:full_position_of_0_analysis_danish}
\end{figure*}

\newcommand{\elbow}{%
  \rule[2pt]{0.75pt}{1.2ex}%
  \makebox[6pt][l]{\hspace{0pt}\rule[2pt]{0.6em}{0.75pt}}%
}

\section{Contradiction Setting}
\label{app:contradiction_setting}
Here we present the results of our contradiction setting experiments. Table~\ref{tab:ck_score_contradiction} reports the contextual knowledge (CK) scores for each model across English, Danish, and Spanish. These scores are measured under three conditions: original, contradiction-rich, and split contexts. For the split context, we further distinguish between the cases where true information appears first and where false information appears first.

\begin{table*}[ht]
\centering
\small
\begin{tabular}{lcccccc}
\toprule
\textbf{Method} & \textbf{GPT-4o} & \textbf{Gemini 1.5 Pro} & \textbf{LLaMA 3.2 3B} & \textbf{LLaMA 3.2 90B} & \textbf{GPT-o3} & \textbf{Qwen 3 235B} \\
\midrule
\multicolumn{7}{c}{\textbf{\textit{English}}} \\
\textbf{All Factual}                 & 72.11 & 76.29 & 73.13 & 76.32 & 49.58 & 55.30 \\
\textbf{All Counterfactual}          & 69.40 & 64.97 & 68.27 & 66.91 & 46.34 & 52.85 \\
\textbf{Split}                       & 69.86 & 69.63 & 70.73 & 69.42 & 55.06 & 56.82 \\
\textbf{\quad\elbow\ True First}     & 71.22 & 71.97 & 72.97 & 74.44 & 55.75 & 58.02 \\
\textbf{\quad\elbow\ Fasle First}    & 68.51 & 67.28 & 68.48 & 64.40 & 54.37 & 55.62 \\

\midrule
\multicolumn{7}{c}{\textbf{\textit{Danish}}} \\
\textbf{All Factual}                & 68.17 & 72.71 & 62.28 & 70.62 & 48.76 & 55.61 \\
\textbf{All Counterfactual}         & 53.26 & 48.75 & 52.86 & 61.90 & 42.39 & 50.22 \\
\textbf{Split}                      & 60.62 & 64.90 & 56.93 & 68.09 & 51.06 & 55.10 \\
\textbf{\quad\elbow\ True First}    & 62.61 & 66.08 & 59.40 & 67.80 & 52.9 & 56.08 \\
\textbf{\quad\elbow\ False First}   & 58.64 & 63.72 & 54.45 & 68.96 & 49.21 & 54.12 \\

\midrule
\multicolumn{7}{c}{\textbf{\textit{Spanish}}} \\
\textbf{All Factual}                & 49.47 & 71.56 & 66.73 & 68.91 & 52.16 & 52.93 \\
\textbf{All Counterfactual}         & 59.17 & 51.65 & 63.75 & 54.7 & 48.13 & 50.35 \\
\textbf{Split}                      & 62.77 & 67.14 & 65.1 & 69.43 & 51.64 & 50.11 \\
\textbf{\quad\elbow\ True First}    & 64.46 & 70.91 & 66.09 & 73.21 & 52.9 & 50.93 \\
\textbf{\quad\elbow\ False First}   & 61.08 & 64.09 & 64.11 & 65.65 & 50.39 & 49.3 \\

\bottomrule
\end{tabular}
\caption{CK Score for each model across three languages (English, Danish, Spanish) under original, contradiction-rich, and split context settings.}
\label{tab:ck_score_contradiction}
\end{table*}

\section{CK Prompts and CoT Prompts}
\label{app:ck_prompts_with_cot}

\begin{itemize}
\item The \textbf{original} prompt: "\textit{With this information, Tell me about [Topic]:}"
\item A \textbf{strict} prompt that constrains responses strictly to the provided context: "\textit{Using the provided contexts only, tell me about \texttt{[Topic]}. Use only the provided contexts to answer the question. Avoid introducing any additional information not found in the contexts.} "
\item A \textbf{balanced} prompt that explicitly instructs models to consider all quartiles equally: "\textit{With this information, tell me about \textit{[Topic]}. Focus on creating a response that is balanced and draws fairly from relevant contexts. Your response must reflect balanced usage. Avoid omitting details from any context, even if they seem less relevant. Ensure that no single context dominates your answer.}" 
\item A \textbf{CK Prompt} prompt:   "\textit{With this information, tell me about \texttt{[Topic]}. Use only the provided contexts to answer the question. Avoid introducing any additional information not found in the contexts. At the same time, focus on creating a response that is balanced and draws fairly from relevant contexts. Your response must reflect balanced usage. Avoid omitting details from any context, even if they seem less relevant. Ensure that no single context dominates your answer.}"
\item \textbf{CoT}: \textit{With this information, tell me about \texttt{[Topic]}. Think through the provided contexts step by step before answering. Identify relevant information from each part of the context, and explain how it helps answer the question. Then, provide a final response. Return your output in JSON format: \{ "reasoning": "your reasoning here", "answer": "your final response here" \}. Only include the JSON object in your response.}
\item A \textbf{CoT + CK Prompt} prompt: \textit{With this information, tell me about \texttt{[Topic]}. Use only the provided contexts to answer the question. Avoid introducing any additional information not found in the contexts. At the same time, focus on creating a response that is balanced and draws fairly from relevant contexts. Your response must reflect balanced usage. Avoid omitting details from any context, even if they seem less relevant. Ensure that no single context dominates your answer. Think through the provided contexts step by step before answering. Identify relevant information from each part of the context, and explain how it helps answer the question. Then, provide a final response. Return your output in JSON format: \{ "reasoning": "your reasoning here", "answer": "your final response here" \}. Only include the JSON object in your response.}
\end{itemize}

\section{FActScore}
\label{app:rest_fact_scores}
The original FActScore evaluation relies on English-only Wikipedia as its primary knowledge source. To extend its applicability to Spanish and Danish, we created a custom multilingual knowledge base by using a Wikipedia dataset\footnote{\url{https://huggingface.co/datasets/wikimedia/wikipedia}} that includes Spanish and Danish articles. To better accommodate the multilingual setting, we also replaced the original RoBERTa tokenizer \cite{liu2019robertarobustlyoptimizedbert} with the XLM-RoBERTa tokenizer \cite{conneau2020unsupervisedcrosslingualrepresentationlearning}.

\section{Results of Different CK Prompts}
\label{app:prompt_solutions}
Here we have results of three versions of CK prompts we used to solve contexts ratio and recall imbalance in English, Spanish and Danish with GPT-4o and LLama 3.2 90B. For GPT-4o's results see Figure \ref{fig:different_instruction_gpt4o_english}, \ref{fig:different_instruction_gpt4o_spanish} and \ref{fig:different_instruction_gpt4o_danish} . For Llama 3.2 90B see Figure \ref{fig:different_instruction_llama3290b_english}, \ref{fig:different_instruction_llama3290b_spanish} and \ref{fig:different_instruction_llama3290b_danish}. Qwen 3 235B see figure \ref{fig:different_instruction_qwen3235b_english}, \ref{fig:different_instruction_qwen3235b_spanish} and \ref{fig:different_instruction_qwen3235b_danish}.

\begin{figure*}[t!]
    \centering
    \begin{subfigure}[t]{0.32\textwidth}
    \includegraphics[width=1\textwidth]{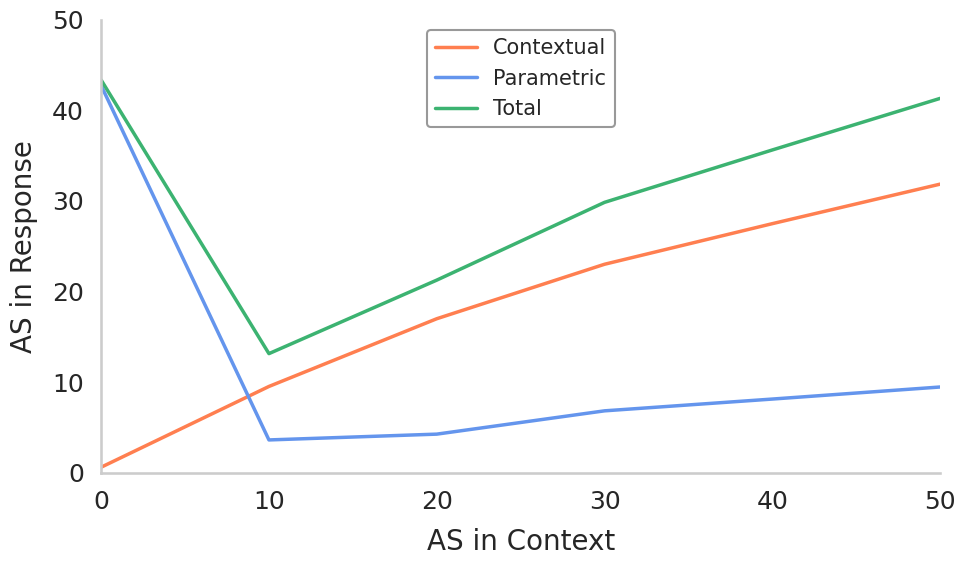}
    \caption{Strict Context Only Instruction}
    \end{subfigure}
    \hfill
    \begin{subfigure}[t]{0.32\textwidth}
    \includegraphics[width=\textwidth]{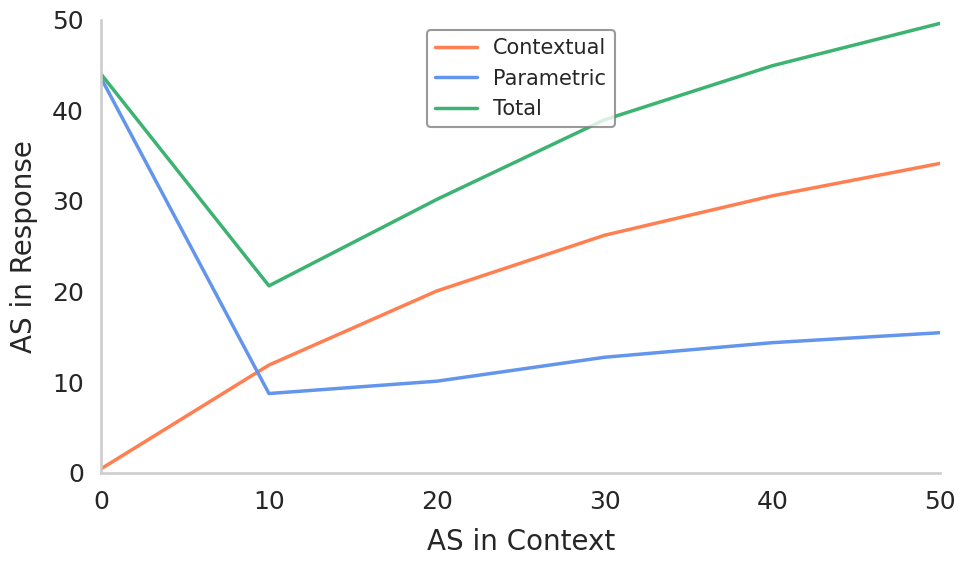}
    \caption{Balanced Use of Context Instruction}
    \end{subfigure}
    \hfill
    \begin{subfigure}[t]{0.32\textwidth}
    \includegraphics[width=\textwidth]{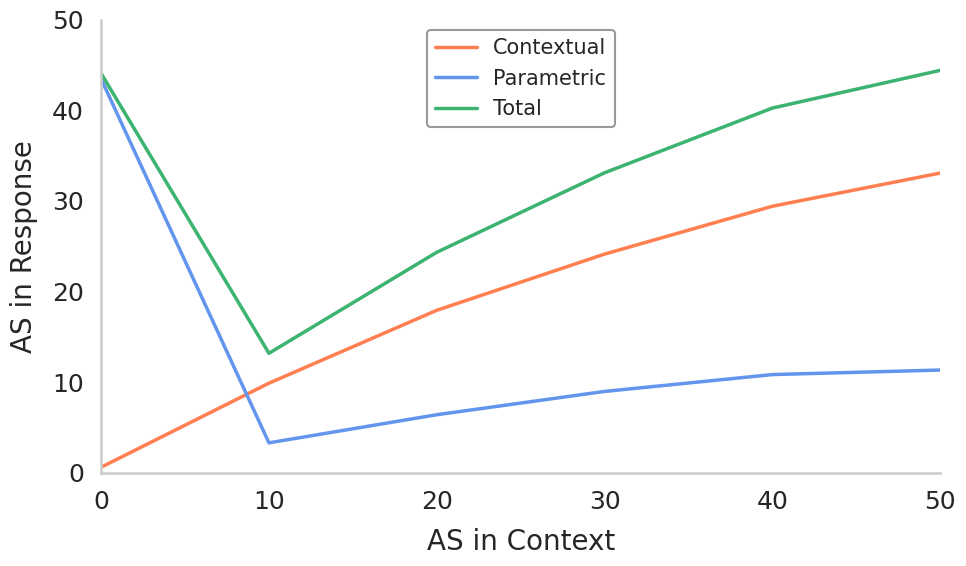}
    \caption{CK Prompt Instruction}
    \end{subfigure}
    \hfill
    \begin{subfigure}[t]{0.32\textwidth}
    \includegraphics[width=\textwidth]{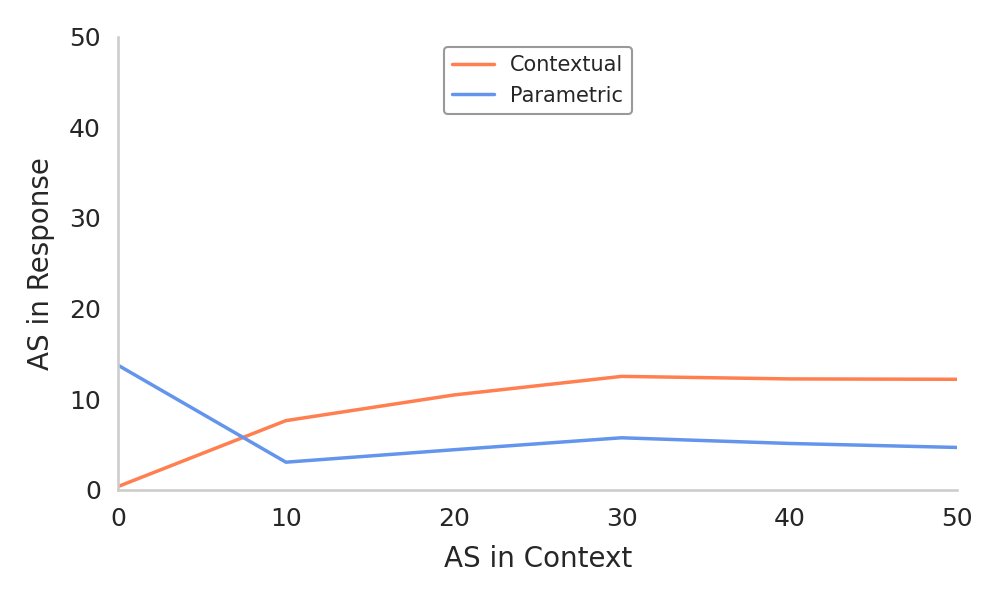}
    \caption{CoT}
    \end{subfigure}
    \begin{subfigure}[t]{0.32\textwidth}
    \includegraphics[width=\textwidth]{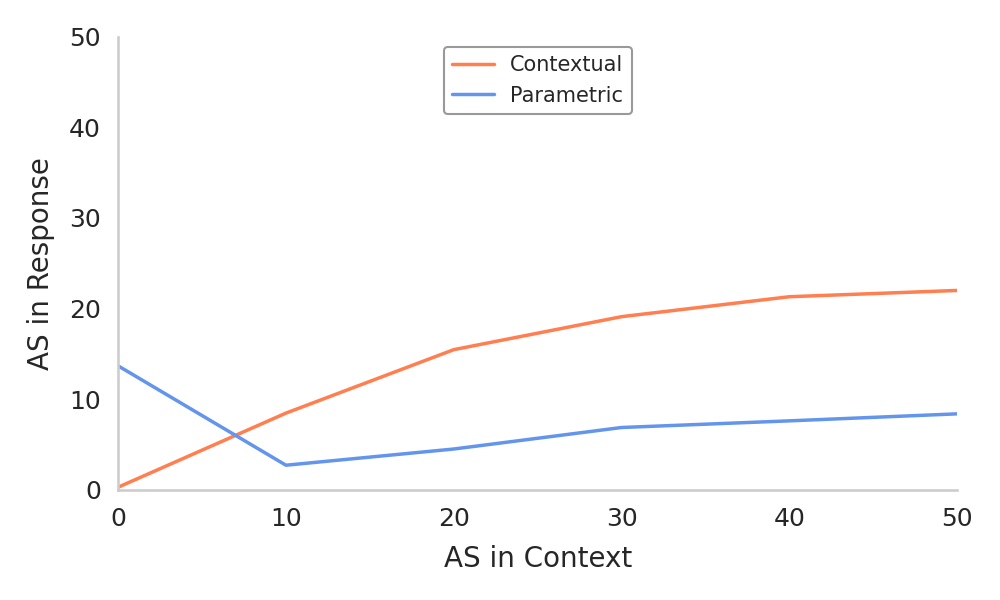}
    \caption{CoT + CK Prompt Instruction}
    \end{subfigure}

    \hfill
    \begin{subfigure}[t]{0.32\textwidth}
    \includegraphics[width=\textwidth]{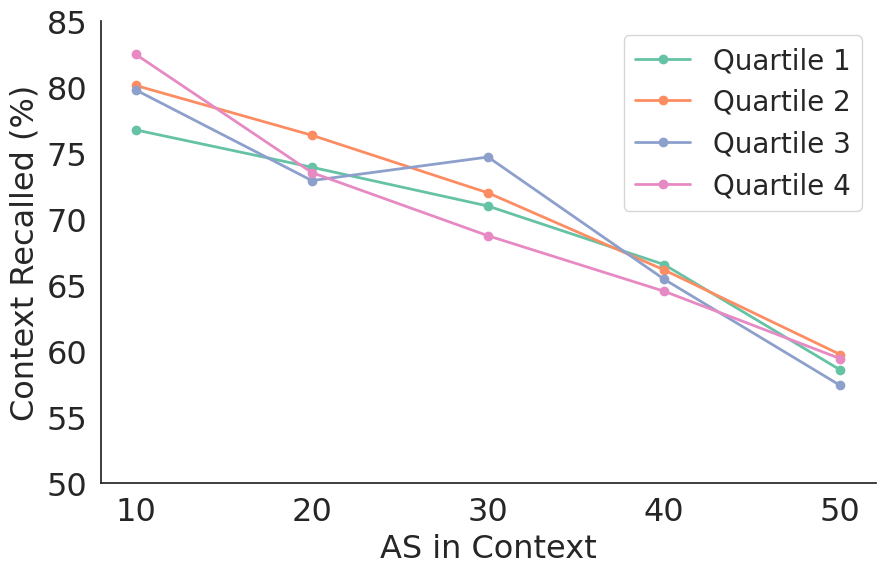}
    \caption{Strict Context Only Instruction}
    \end{subfigure}
    \hfill
    \begin{subfigure}[t]{0.32\textwidth}
    \includegraphics[width=\textwidth]{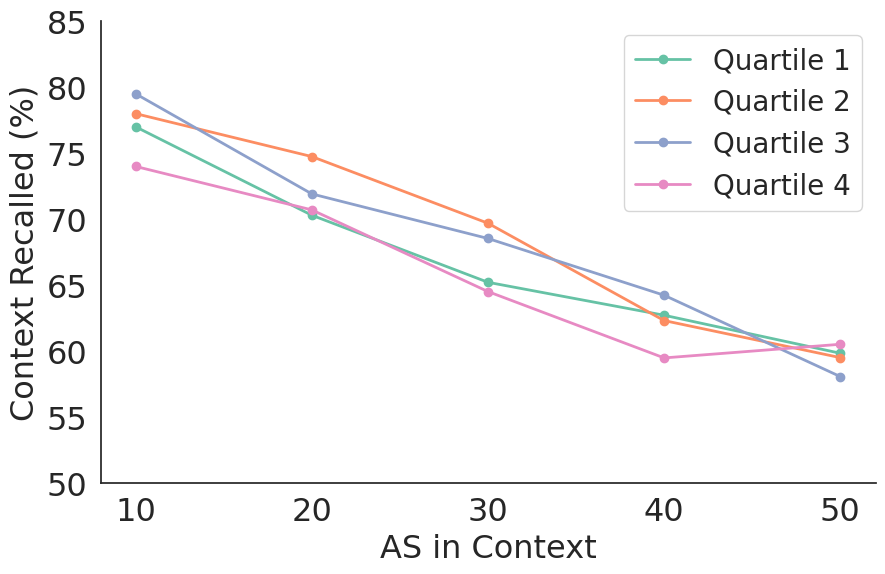}
    \caption{Balanced Use of Context Instruction}
    \end{subfigure}
    \hfill
    \begin{subfigure}[t]{0.32\textwidth}
    \includegraphics[width=\textwidth]{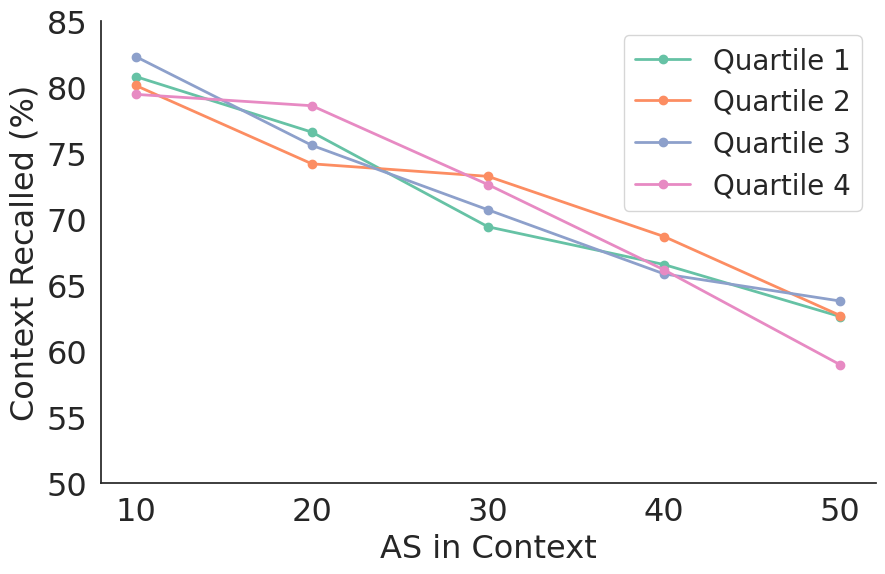}
    \caption{CK Prompt Instruction}
    \end{subfigure}
    \begin{subfigure}[t]{0.32\textwidth}
    \includegraphics[width=\textwidth]{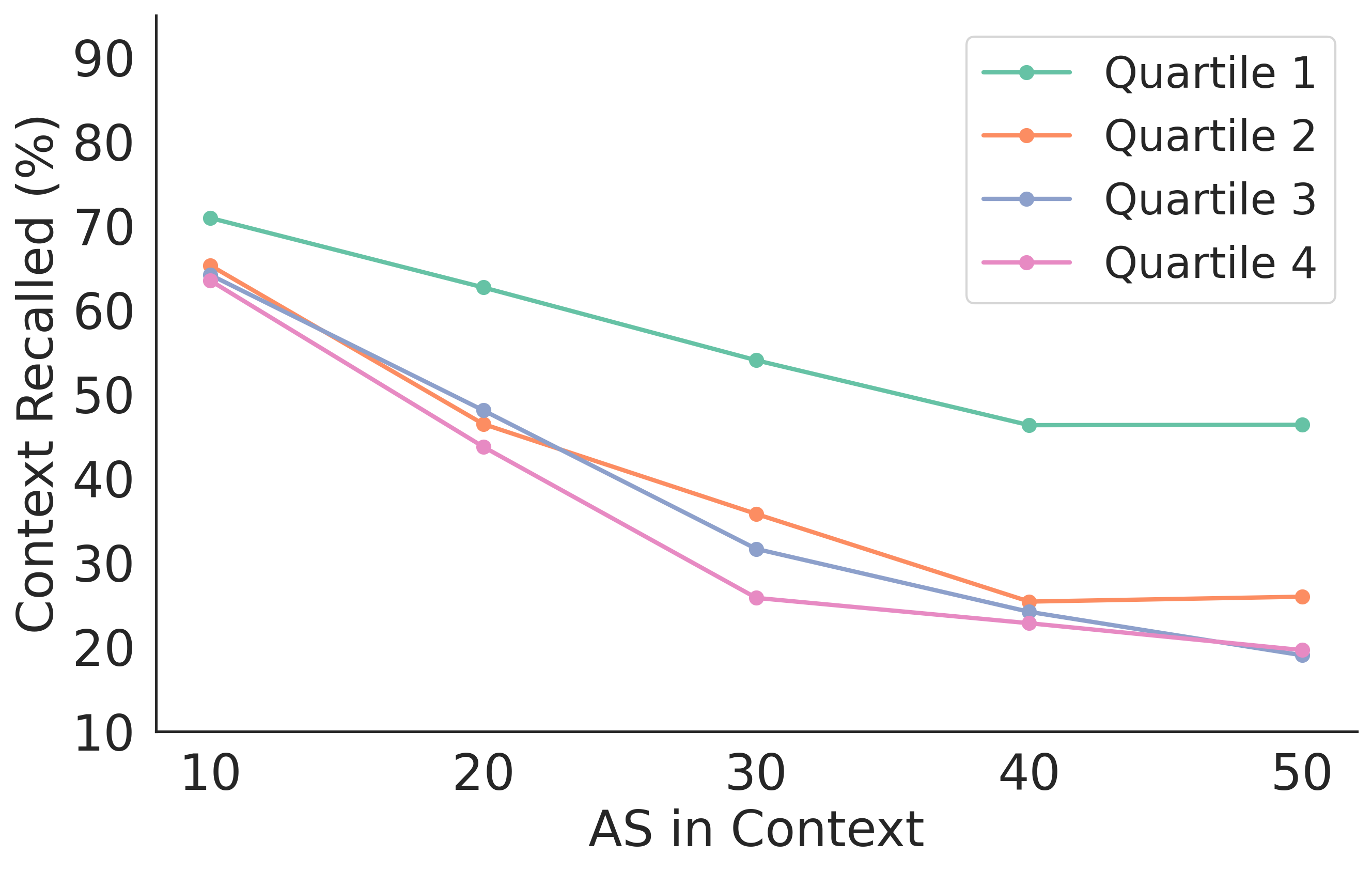}
    \caption{CoT}
    \end{subfigure}
    \begin{subfigure}[t]{0.32\textwidth}
    \includegraphics[width=\textwidth]{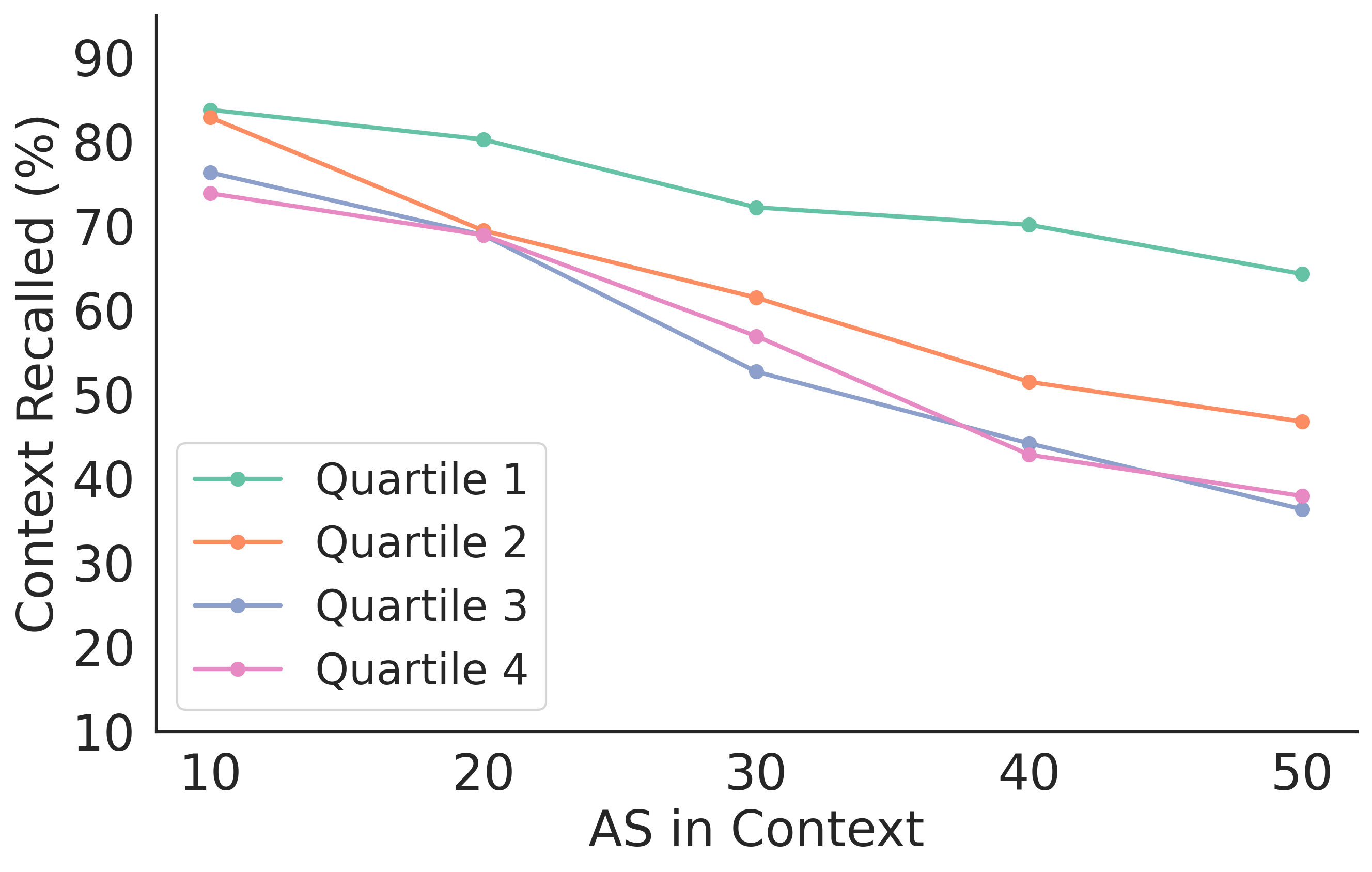}
    \caption{CoT + CK Prompt Instruction}
    \end{subfigure}
    \caption{GPT-4o average number of CK/PK in responses (Top) and context recall (Bottom) across Different Instructions in English}
    \label{fig:different_instruction_gpt4o_english}
\end{figure*}

\begin{figure*}[t!]
    \centering
    \begin{subfigure}[t]{0.32\textwidth}
    \includegraphics[width=1\textwidth]{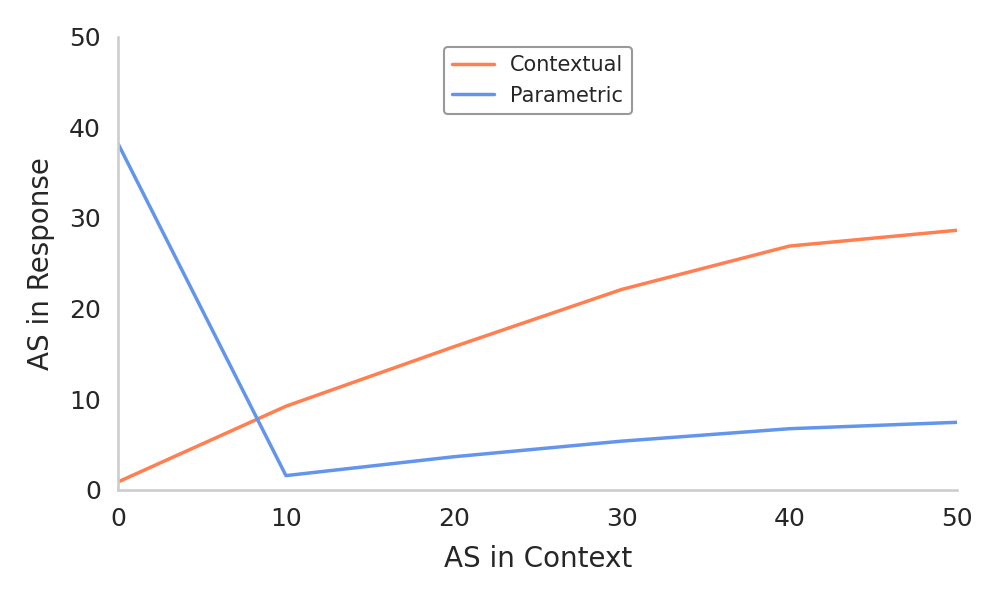}
    \caption{Strict Context Only Instruction}
    \end{subfigure}
    \hfill
    \begin{subfigure}[t]{0.32\textwidth}
    \includegraphics[width=\textwidth]{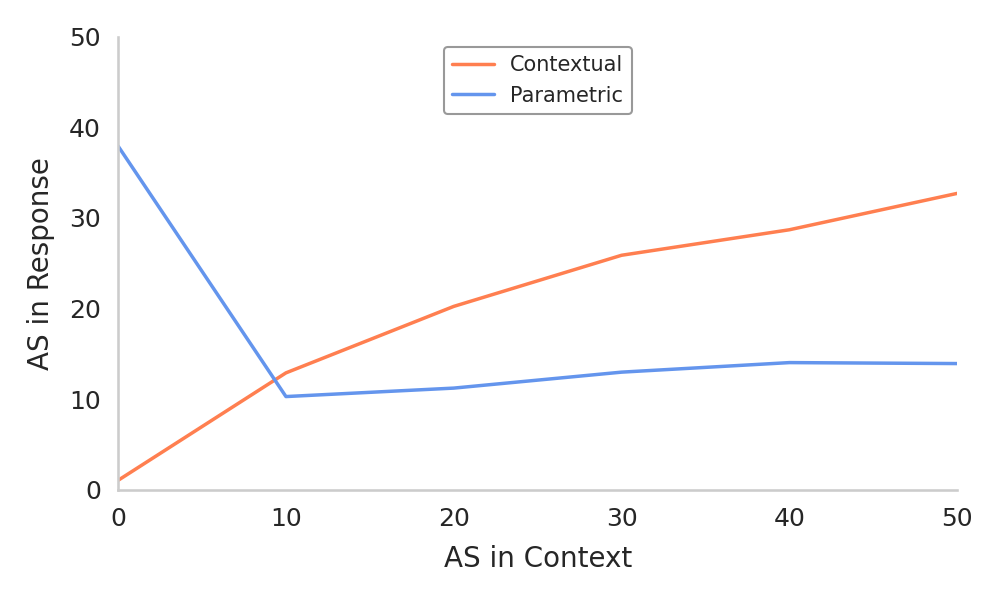}
    \caption{Balanced Use of Context Instruction}
    \end{subfigure}
    \hfill
    \begin{subfigure}[t]{0.32\textwidth}
    \includegraphics[width=\textwidth]{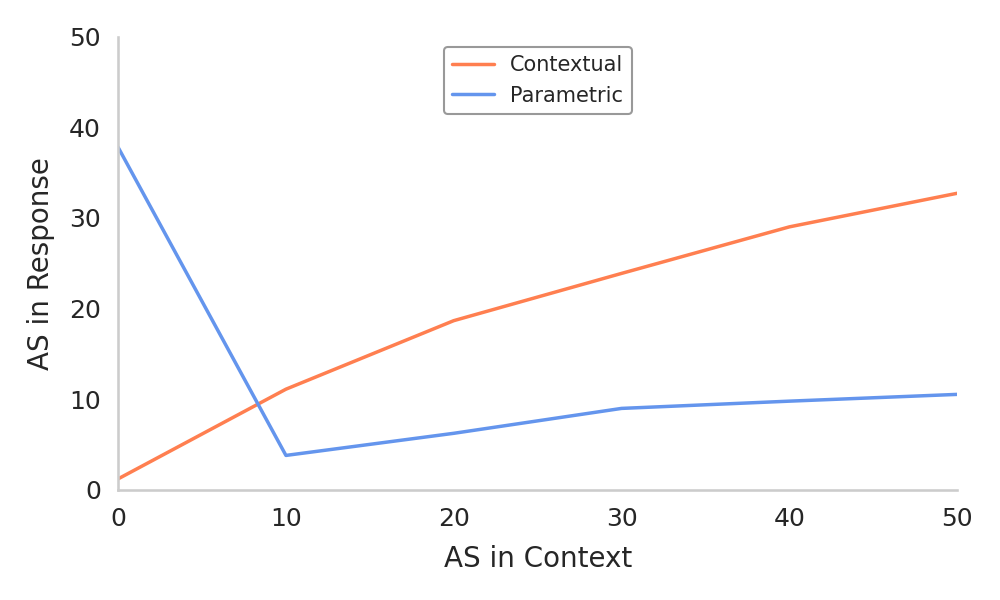}
    \caption{CK Prompt Instruction}
    \end{subfigure}
    \begin{subfigure}[t]{0.32\textwidth}
    \includegraphics[width=\textwidth]{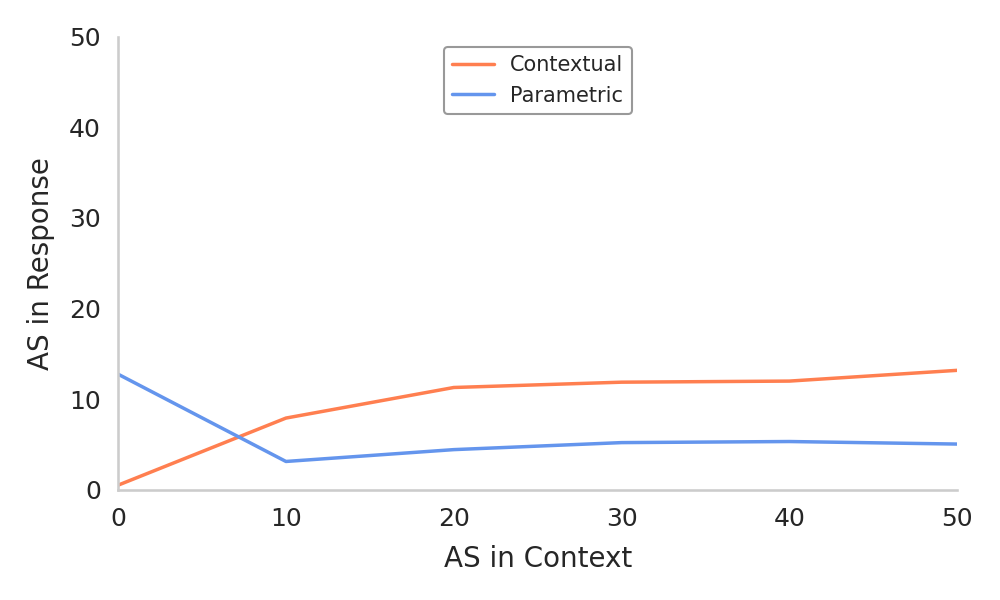}
    \caption{CoT}
    \end{subfigure}
    \begin{subfigure}[t]{0.32\textwidth}
    \includegraphics[width=\textwidth]{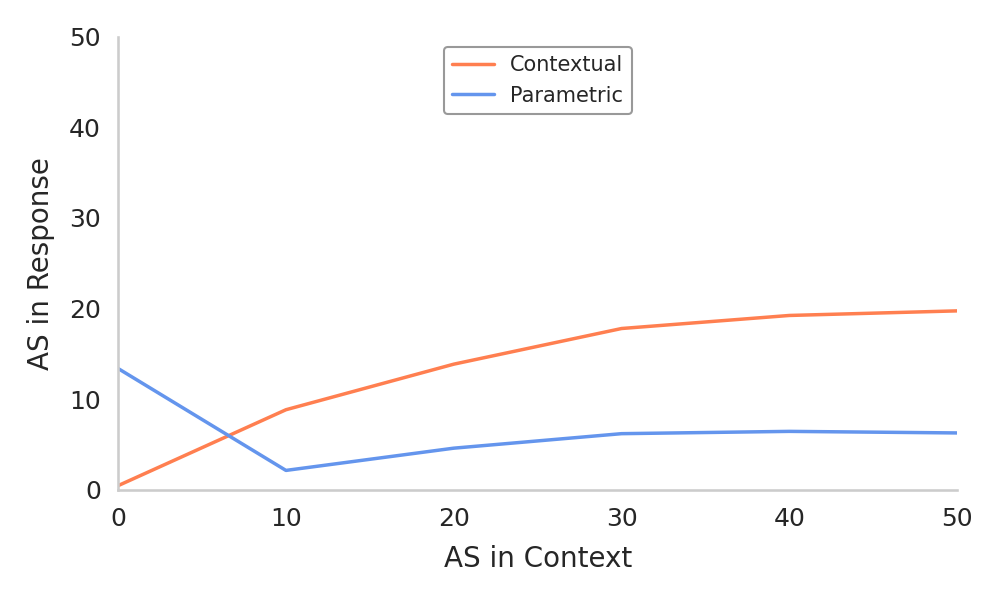}
    \caption{CoT + CK Prompt Instruction}
    \end{subfigure}
    
    \hfill
    \begin{subfigure}[t]{0.32\textwidth}
    \includegraphics[width=\textwidth]{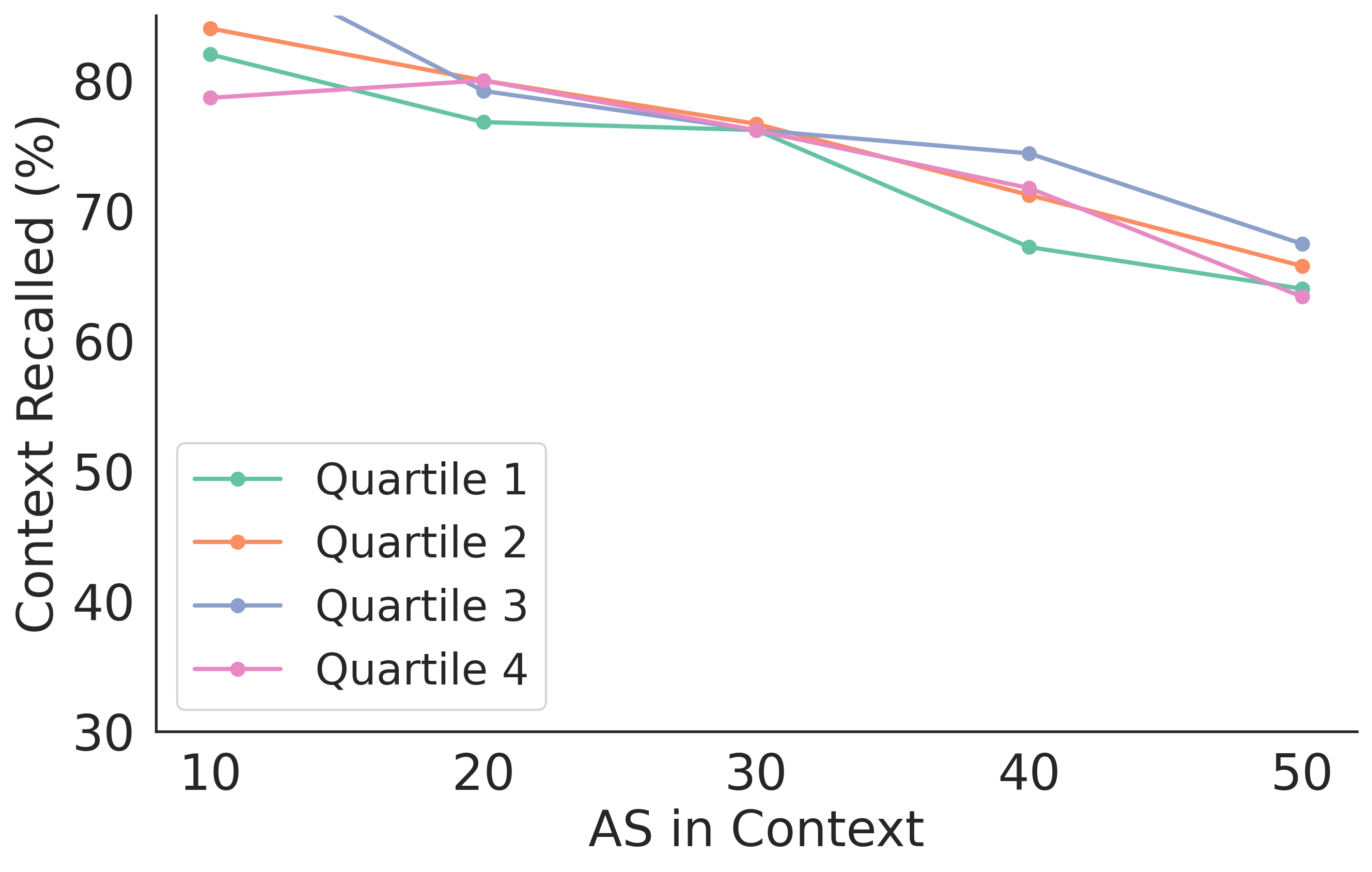}
    \caption{Strict Context Only Instruction}
    \end{subfigure}
    \hfill
    \begin{subfigure}[t]{0.32\textwidth}
    \includegraphics[width=\textwidth]{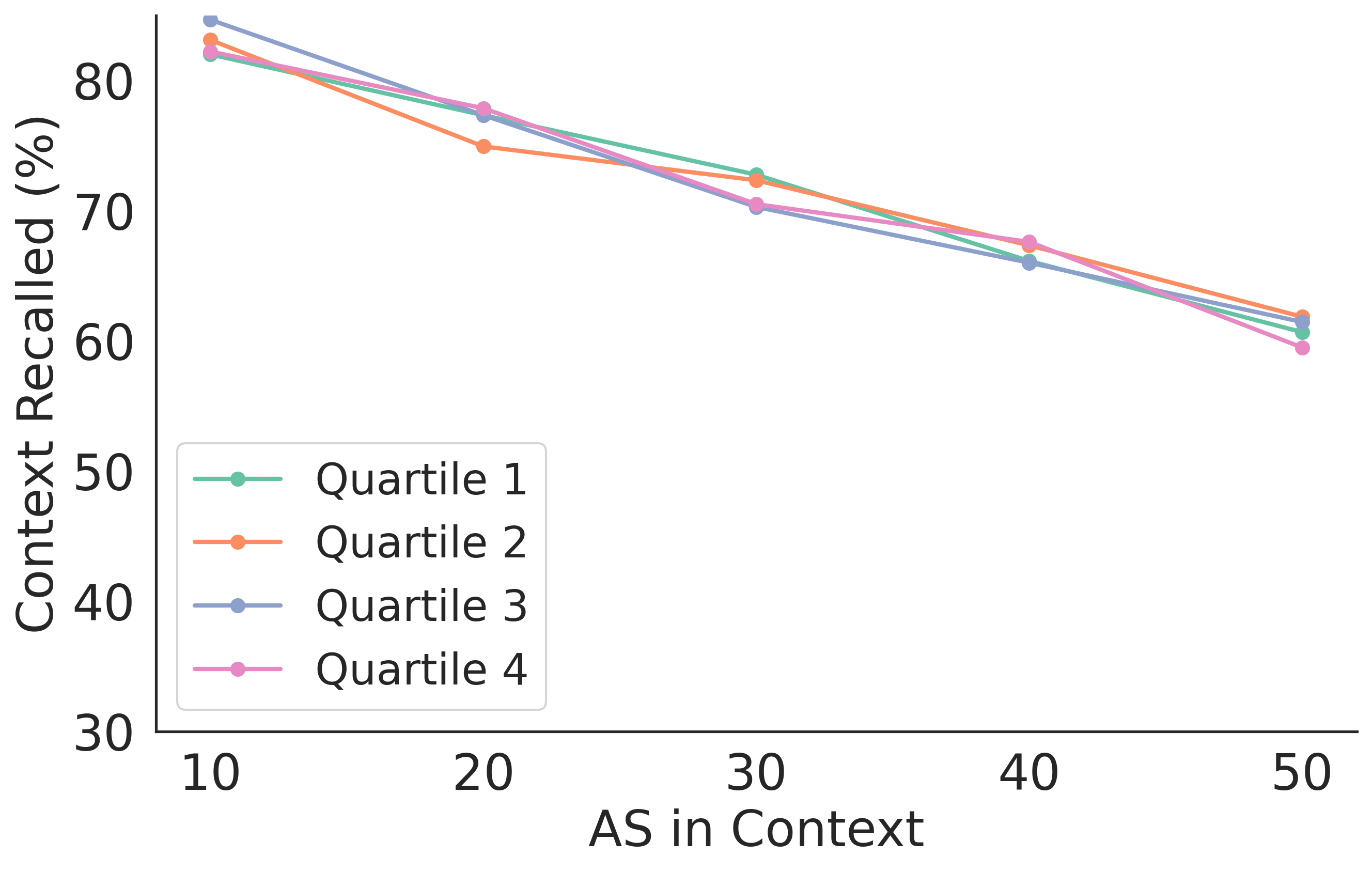}
    \caption{Balanced Use of Context Instruction}
    \end{subfigure}
    \hfill
    \begin{subfigure}[t]{0.32\textwidth}
    \includegraphics[width=\textwidth]{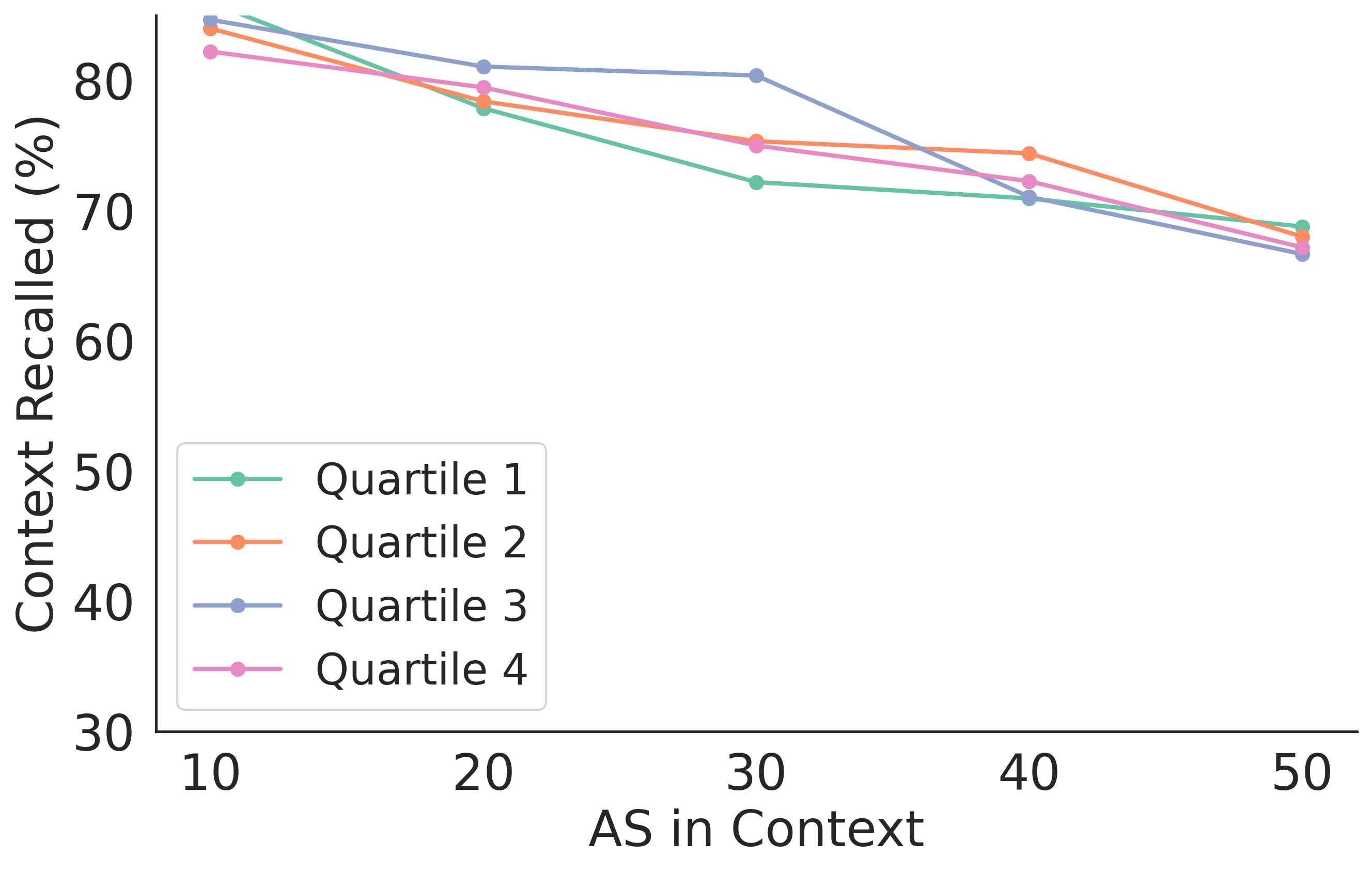}
    \caption{CK Prompt Instruction}
    \end{subfigure}
    \begin{subfigure}[t]{0.32\textwidth}
    \includegraphics[width=\textwidth]{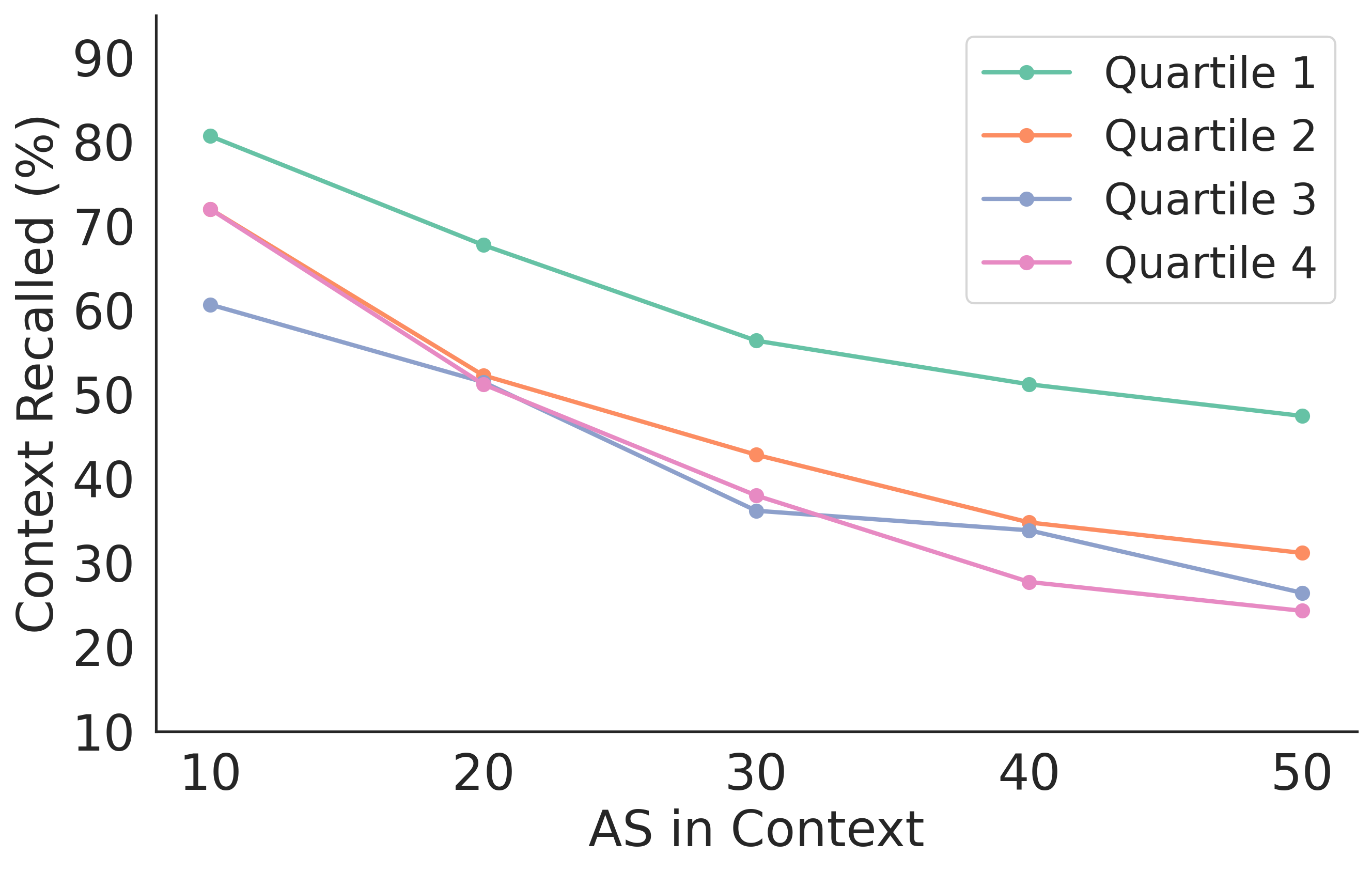}
    \caption{CoT}
    \end{subfigure}
    \begin{subfigure}[t]{0.32\textwidth}
    \includegraphics[width=\textwidth]{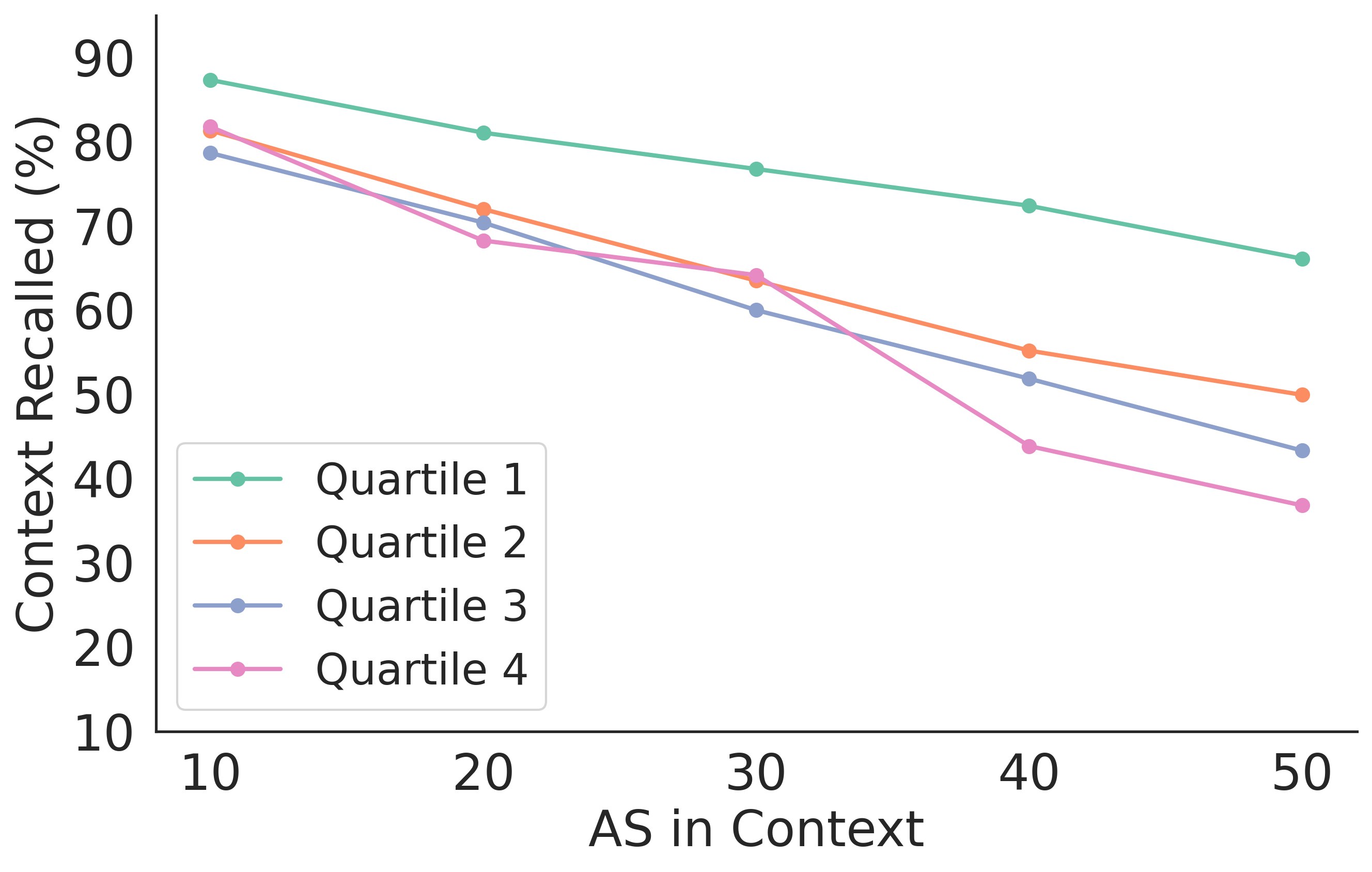}
    \caption{CoT + CK Prompt Instruction}
    \end{subfigure}
    \caption{GPT-4o average number of CK/PK in responses (Top) and context recall (Bottom) across Different Instructions in Spanish}
    \label{fig:different_instruction_gpt4o_spanish}
\end{figure*}

\begin{figure*}[t!]
    \centering
    \begin{subfigure}[t]{0.32\textwidth}
    \includegraphics[width=1\textwidth]{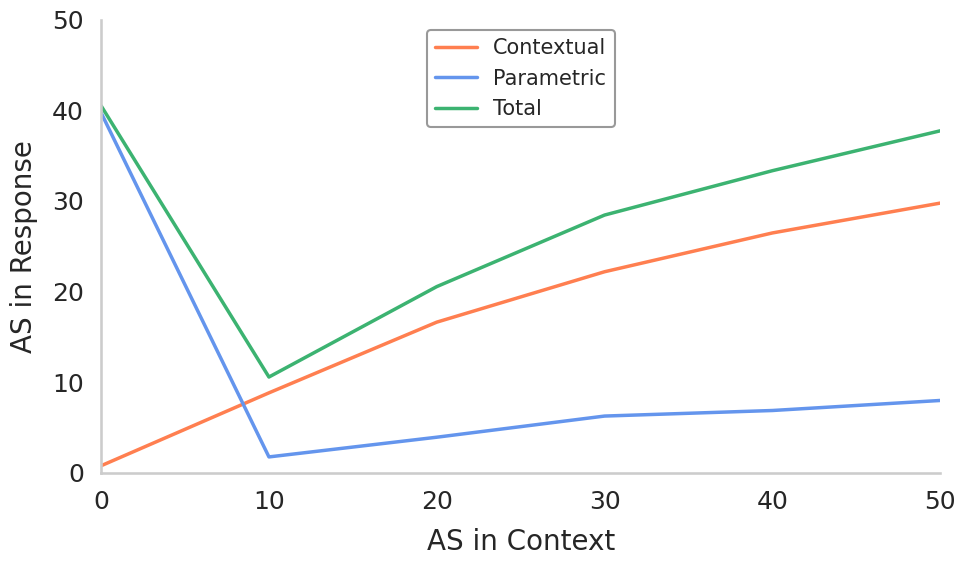}
    \caption{Strict Context Only Instruction}
    \end{subfigure}
    \hfill
    \begin{subfigure}[t]{0.32\textwidth}
    \includegraphics[width=\textwidth]{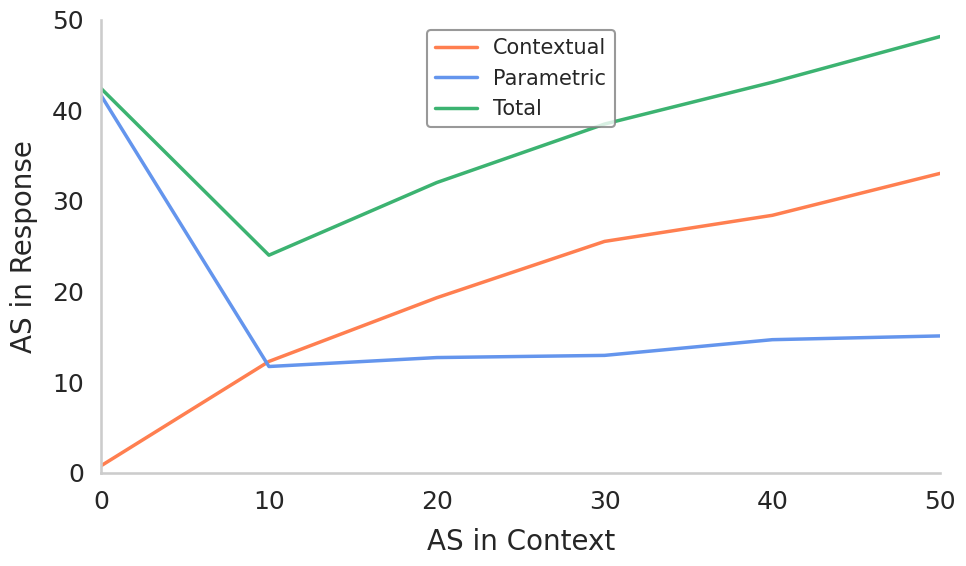}
    \caption{Balanced Use of Context Instruction}
    \end{subfigure}
    \hfill
    \begin{subfigure}[t]{0.32\textwidth}
    \includegraphics[width=\textwidth]{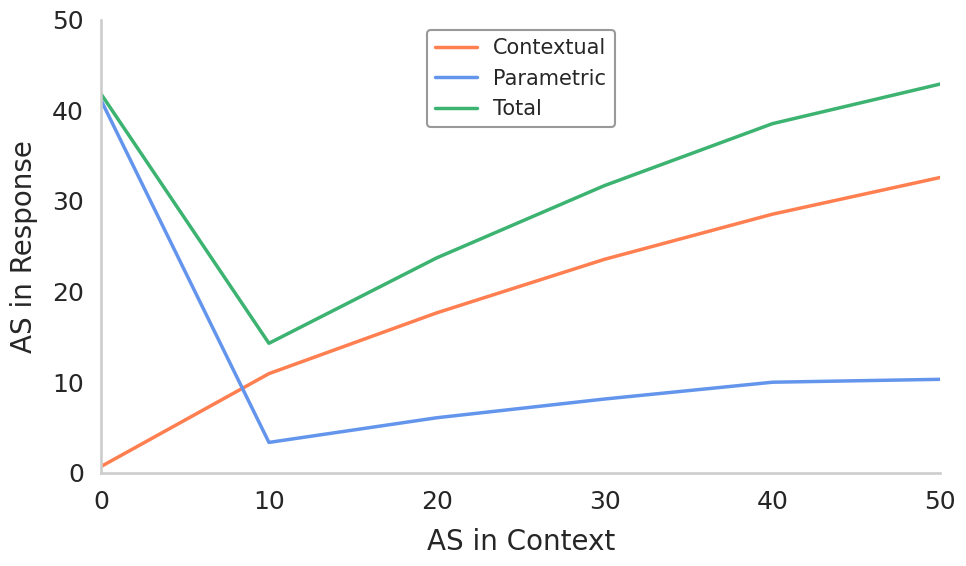}
    \caption{CK Prompt Instruction}
    \end{subfigure}
    \begin{subfigure}[t]{0.32\textwidth}
    \includegraphics[width=\textwidth]{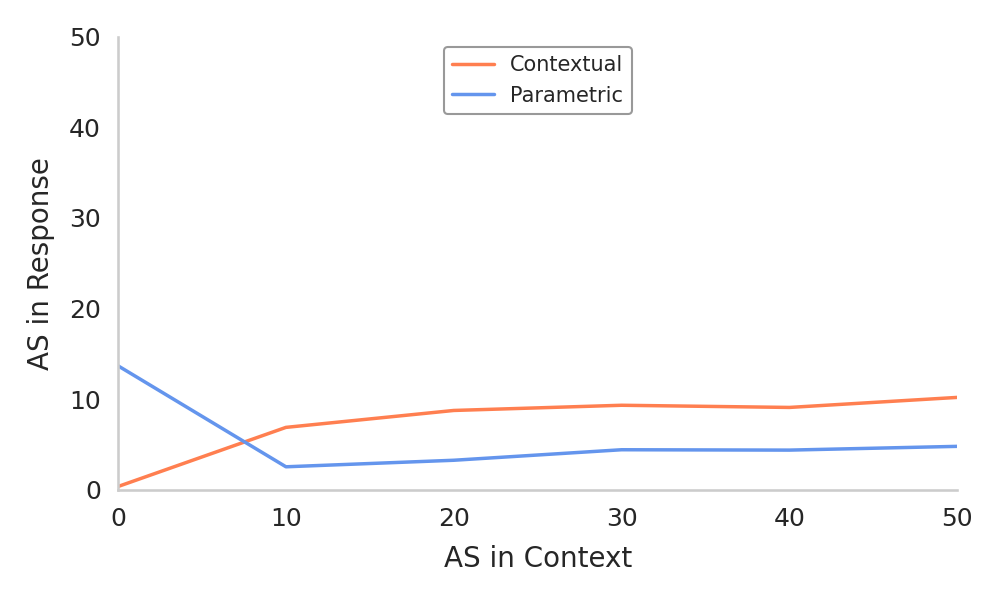}
    \caption{CoT}
    \end{subfigure}
    \begin{subfigure}[t]{0.32\textwidth}
    \includegraphics[width=\textwidth]{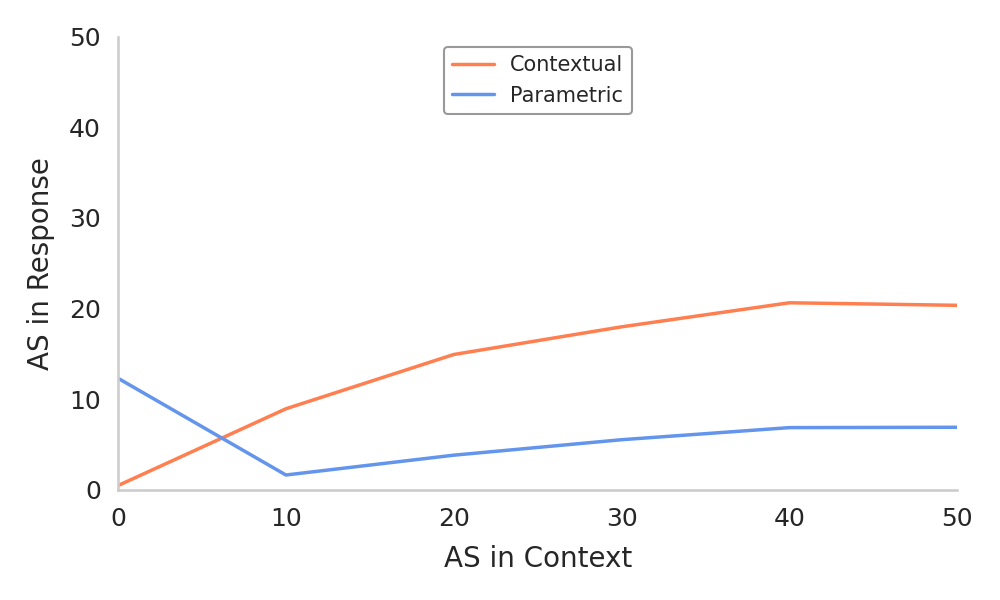}
    \caption{CoT + CK Prompt Instruction}
    \end{subfigure}
    
    \hfill
    \begin{subfigure}[t]{0.32\textwidth}
    \includegraphics[width=\textwidth]{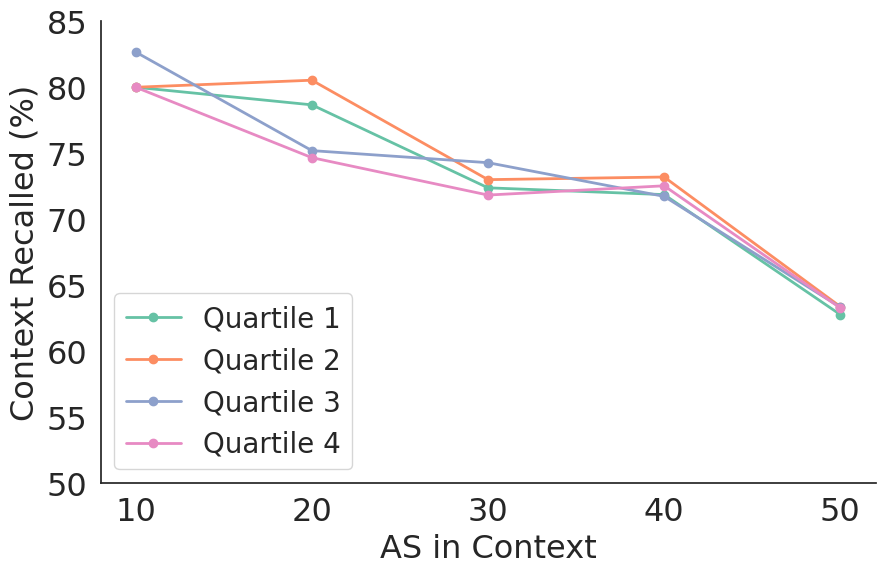}
    \caption{Strict Context Only Instruction}
    \end{subfigure}
    \hfill
    \begin{subfigure}[t]{0.32\textwidth}
    \includegraphics[width=\textwidth]{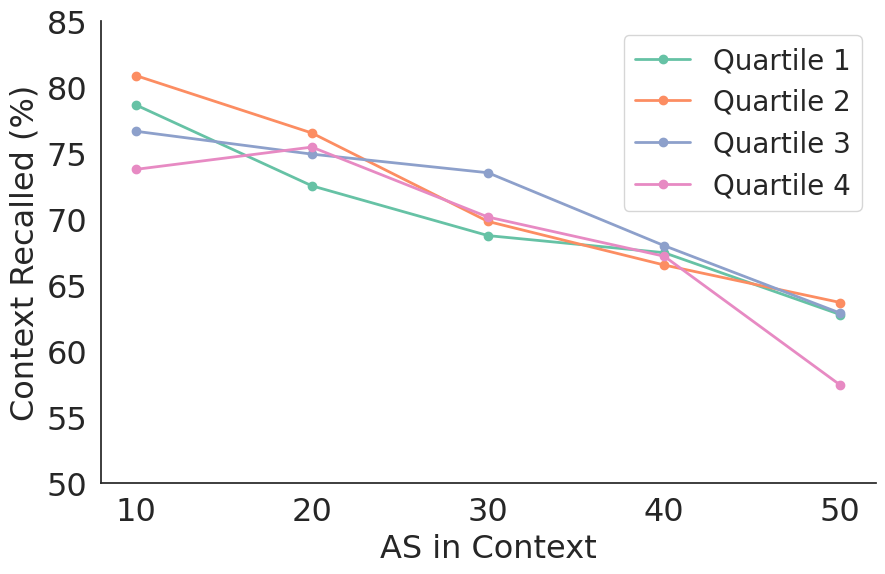}
    \caption{Balanced Use of Context Instruction}
    \end{subfigure}
    \hfill
    \begin{subfigure}[t]{0.32\textwidth}
    \includegraphics[width=\textwidth]{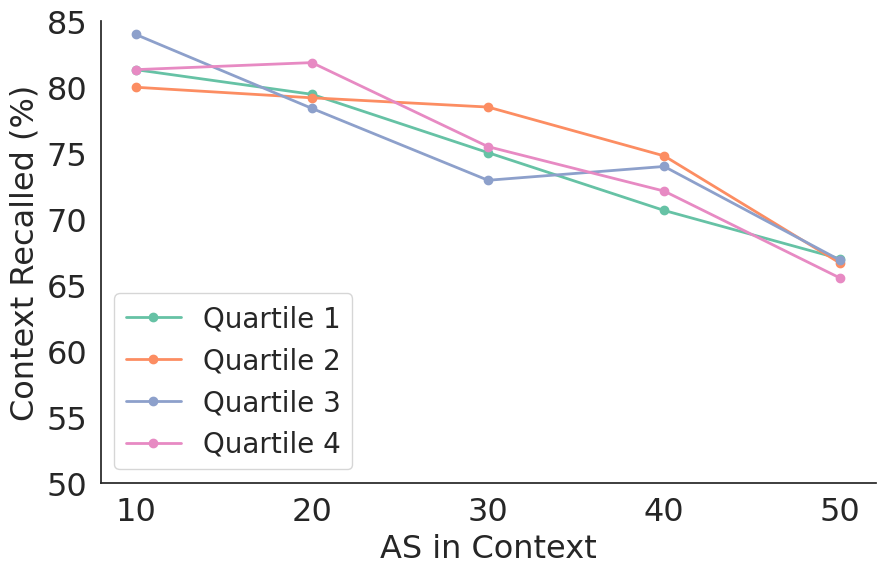}
    \caption{CK Prompt Instruction}
    \end{subfigure}
    \begin{subfigure}[t]{0.32\textwidth}
    \includegraphics[width=\textwidth]{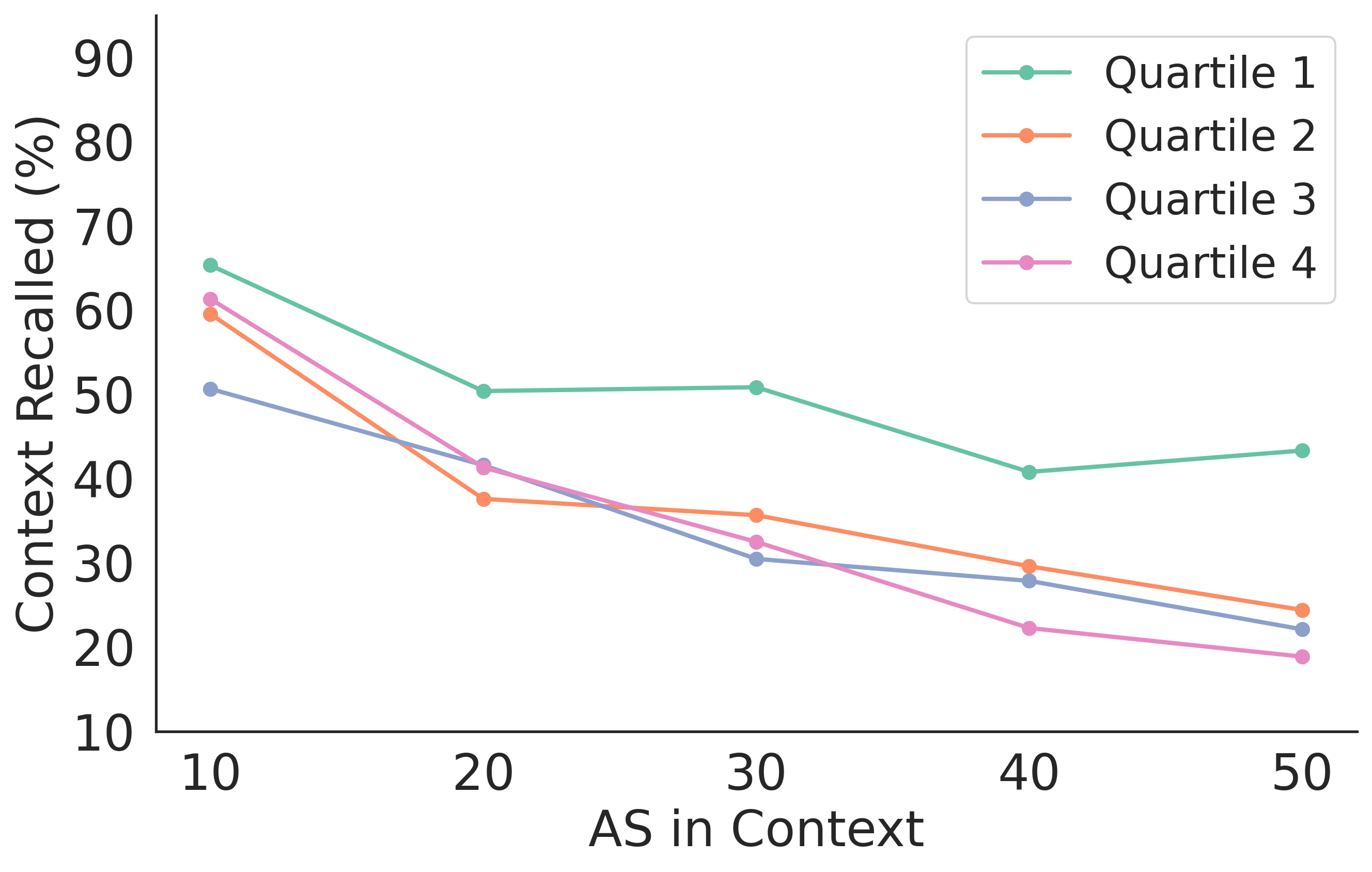}
    \caption{CoT}
    \end{subfigure}
    \begin{subfigure}[t]{0.32\textwidth}
    \includegraphics[width=\textwidth]{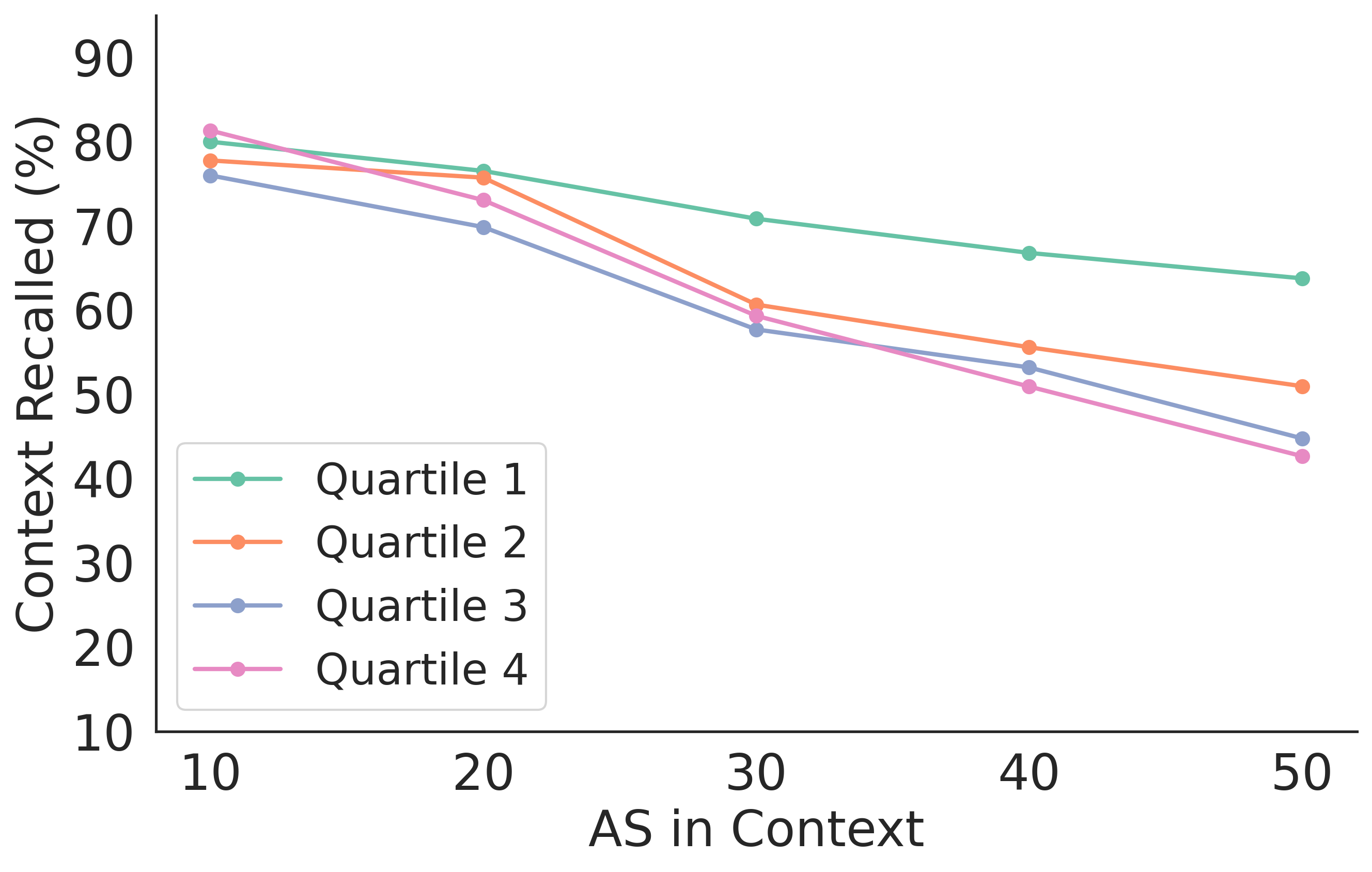}
    \caption{CoT + CK Prompt Instruction}
    \end{subfigure}
    \caption{GPT-4o average number of CK/PK in responses (Top) and context recall (Bottom) across Different Instructions in  Danish}
    \label{fig:different_instruction_gpt4o_danish}
\end{figure*}

\begin{figure*}[t!]
    \centering
    \begin{subfigure}[t]{0.32\textwidth}
    \includegraphics[width=1\textwidth]{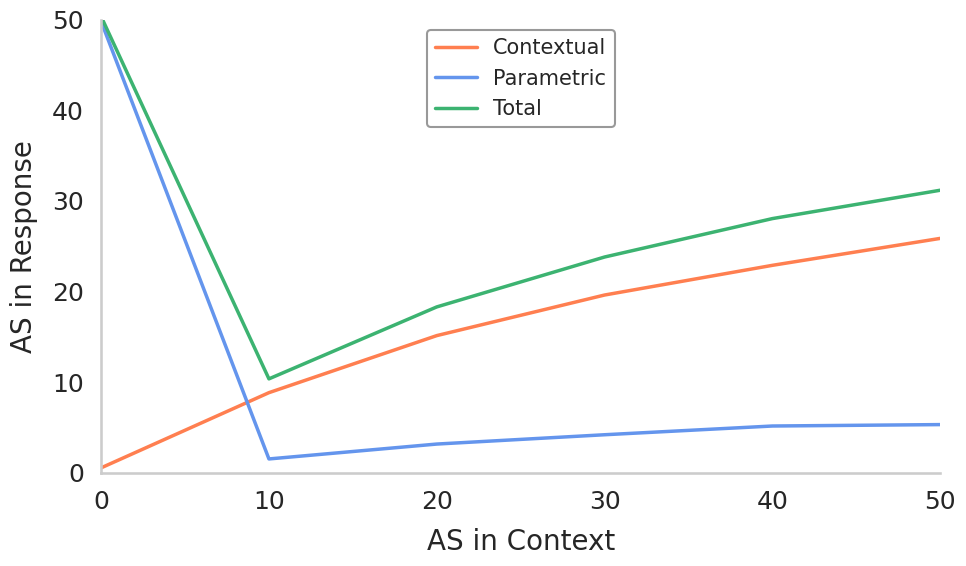}
    \caption{Strict Context Only Instruction}
    \end{subfigure}
    \hfill
    \begin{subfigure}[t]{0.32\textwidth}
    \includegraphics[width=\textwidth]{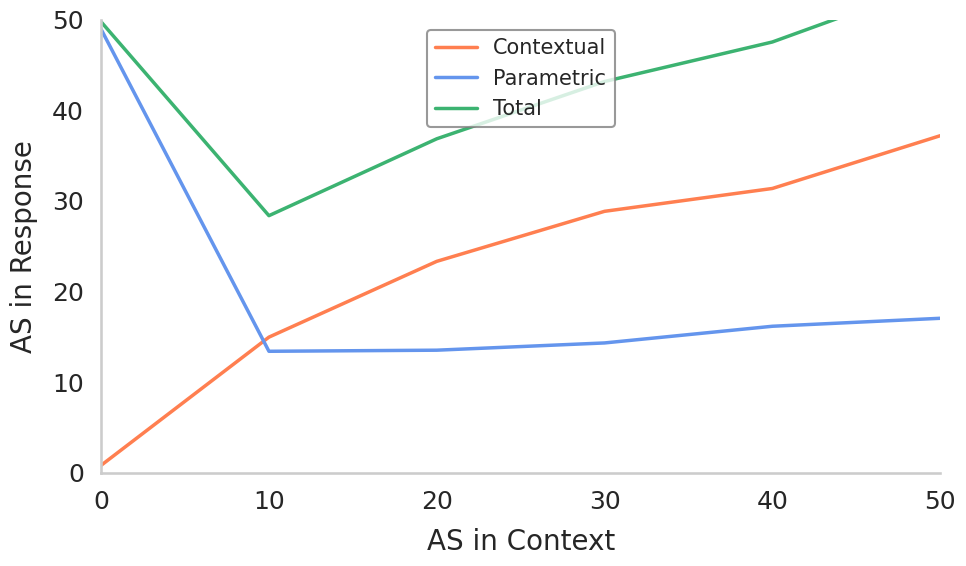}
    \caption{Balanced Use of Context Instruction}
    \end{subfigure}
    \hfill
    \begin{subfigure}[t]{0.32\textwidth}
    \includegraphics[width=\textwidth]{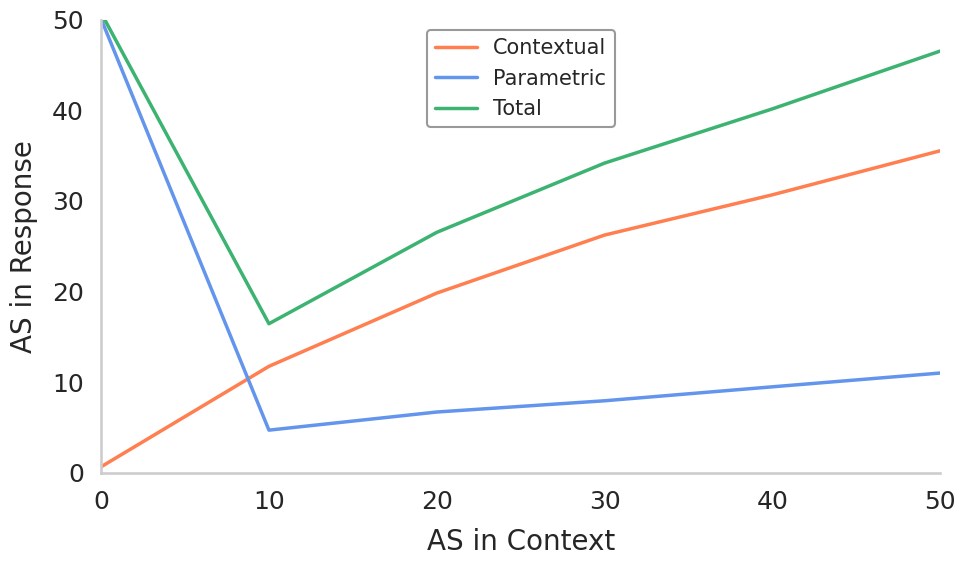}
    \caption{CK Prompt Instruction}
    \end{subfigure}
    \hfill

    \begin{subfigure}[t]{0.32\textwidth}
    \includegraphics[width=\textwidth]{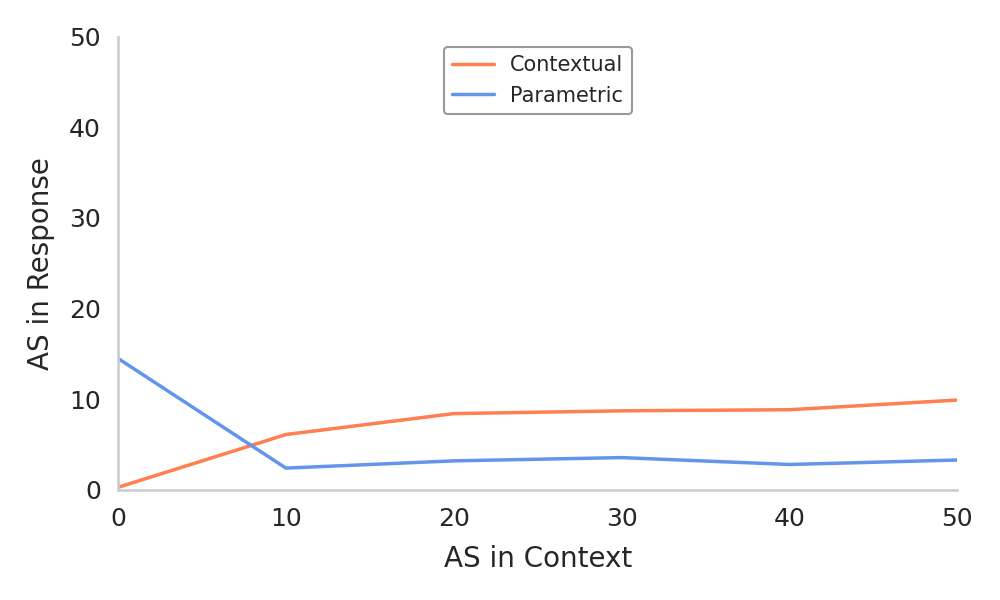}
    \caption{CoT}
    \end{subfigure}
    \begin{subfigure}[t]{0.32\textwidth}
    \includegraphics[width=\textwidth]{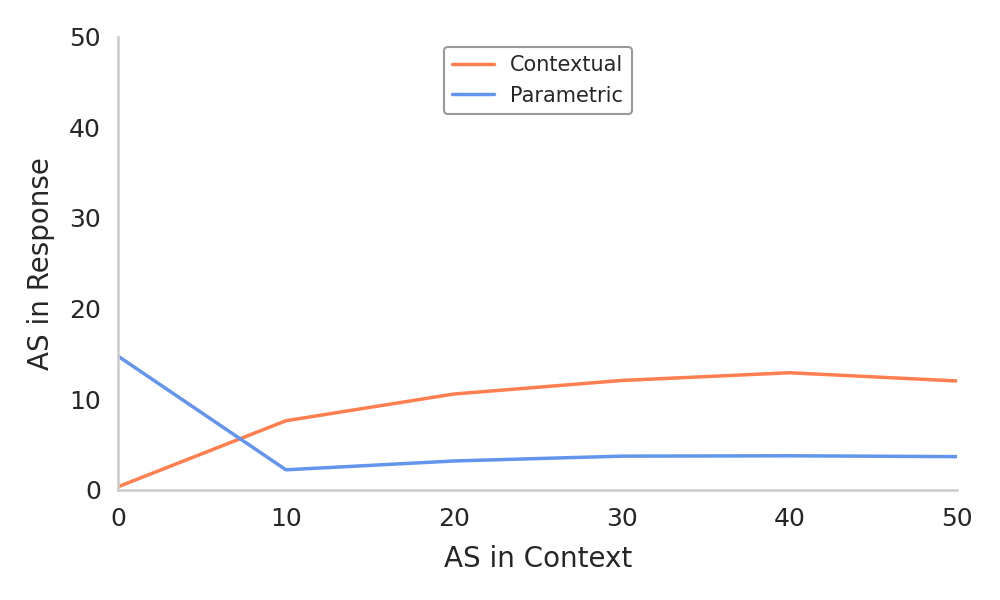}
    \caption{CoT + CK Prompt Instruction}
    \end{subfigure}
    
    \begin{subfigure}[t]{0.32\textwidth}
    \includegraphics[width=\textwidth]{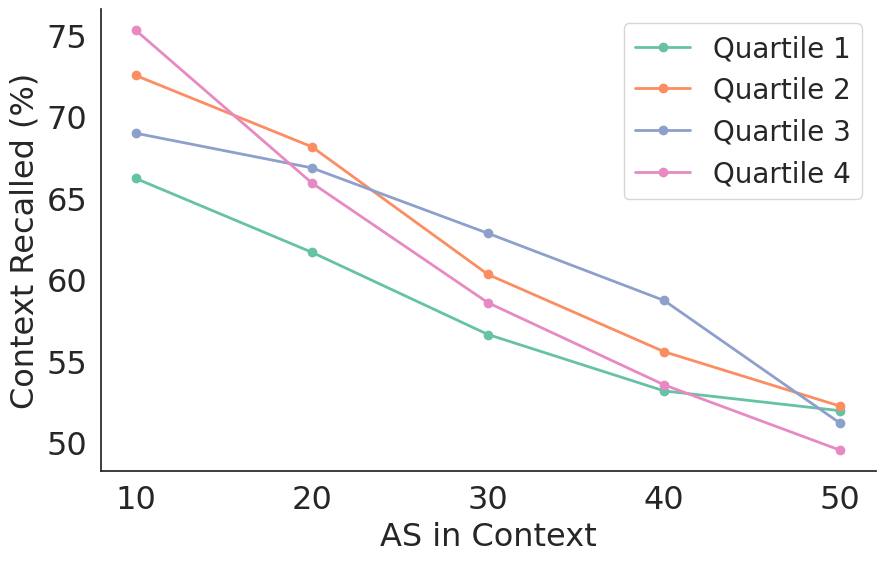}
    \caption{Strict Context Only Instruction}
    \end{subfigure}
    \hfill
    \begin{subfigure}[t]{0.32\textwidth}
    \includegraphics[width=\textwidth]{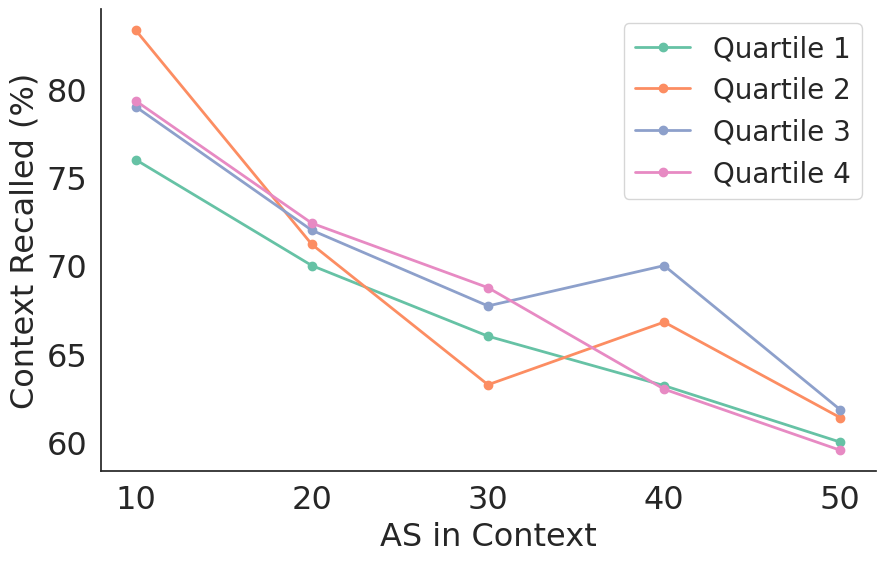}
    \caption{Balanced Use of Context Instruction}
    \end{subfigure}
    \hfill
    \begin{subfigure}[t]{0.32\textwidth}
    \includegraphics[width=\textwidth]{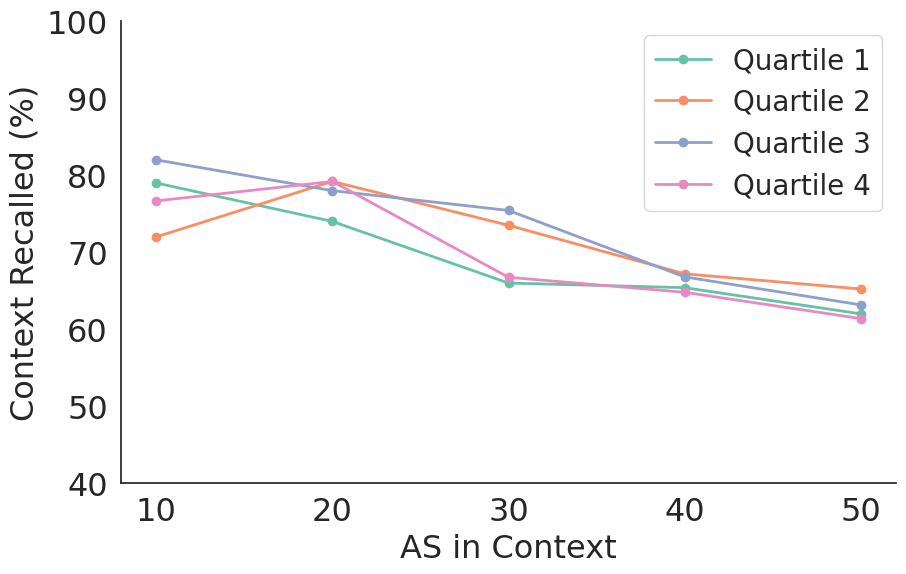}
    \caption{CK Prompt Instruction}
    \end{subfigure}

    \begin{subfigure}[t]{0.32\textwidth}
    \includegraphics[width=\textwidth]{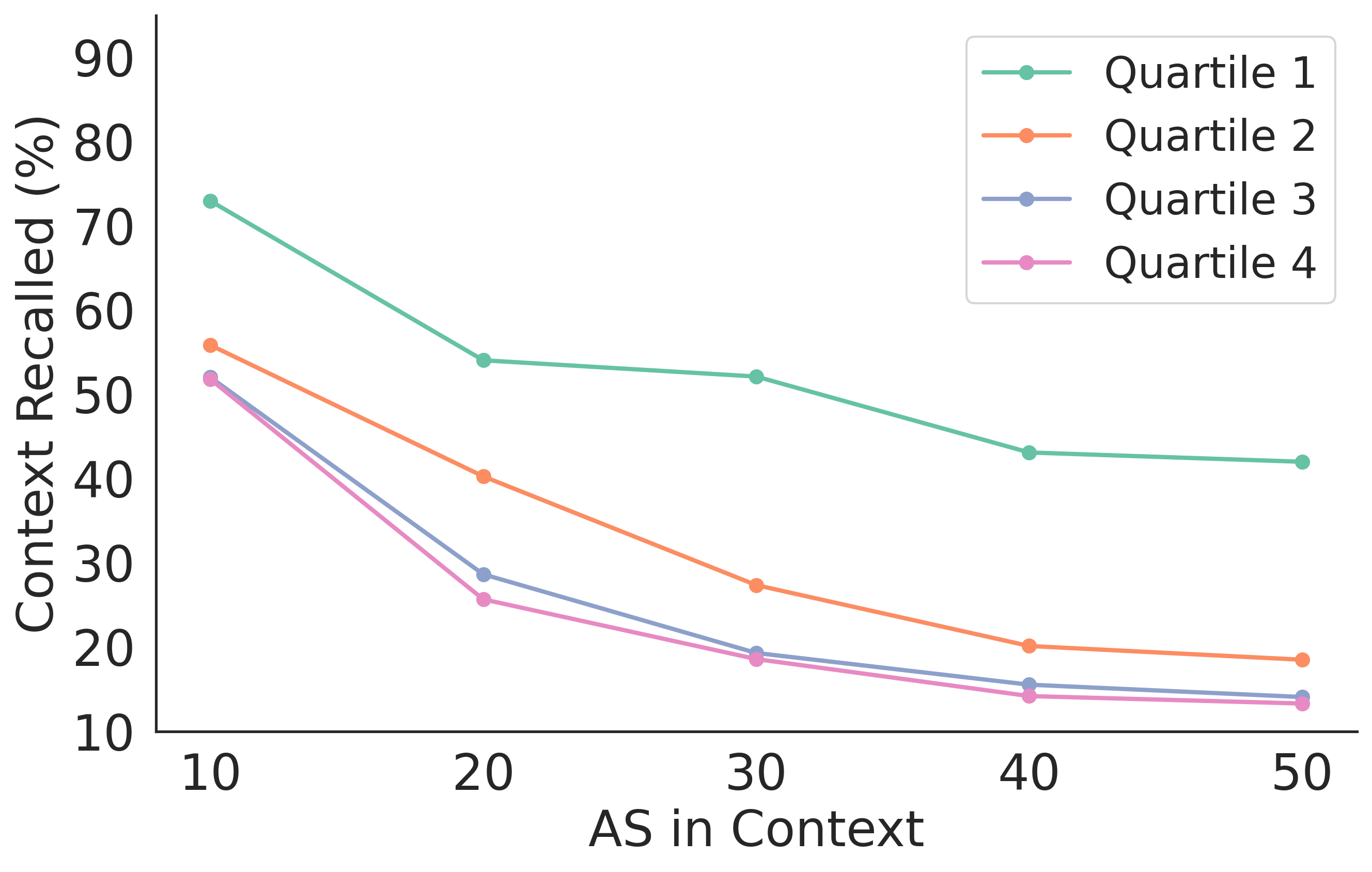}
    \caption{CoT}
    \end{subfigure}
    \begin{subfigure}[t]{0.32\textwidth}
    \includegraphics[width=\textwidth]{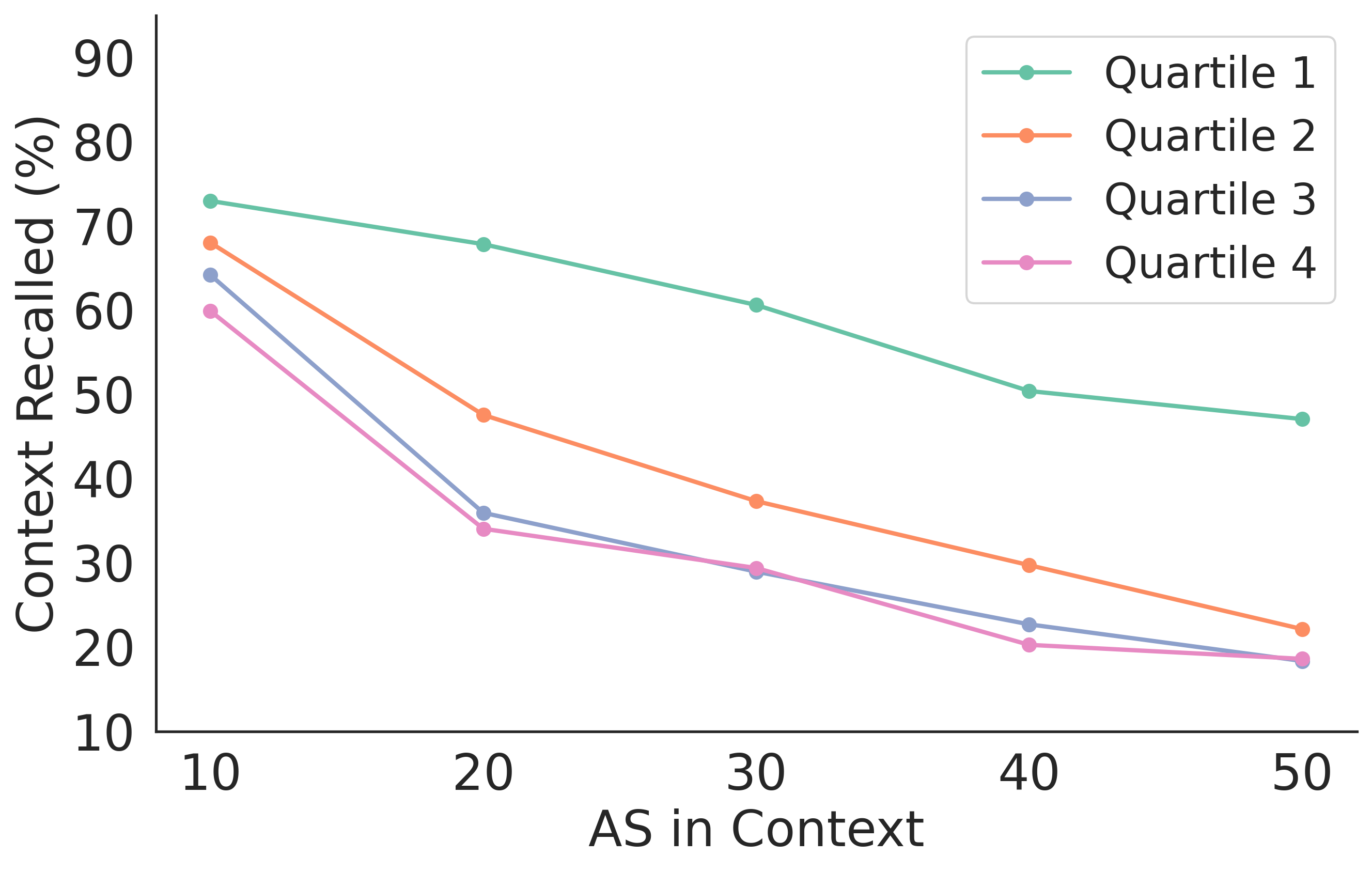}
    \caption{CoT + CK Prompt Instruction}
    \end{subfigure}
    
    \caption{Llama 3.2 90B average number of CK/PK in responses (Top) and context recall (Bottom) across Different Instructions in English}
    \label{fig:different_instruction_llama3290b_english}
\end{figure*}

\begin{figure*}[t!]
    \centering
    \begin{subfigure}[t]{0.32\textwidth}
    \includegraphics[width=1\textwidth]{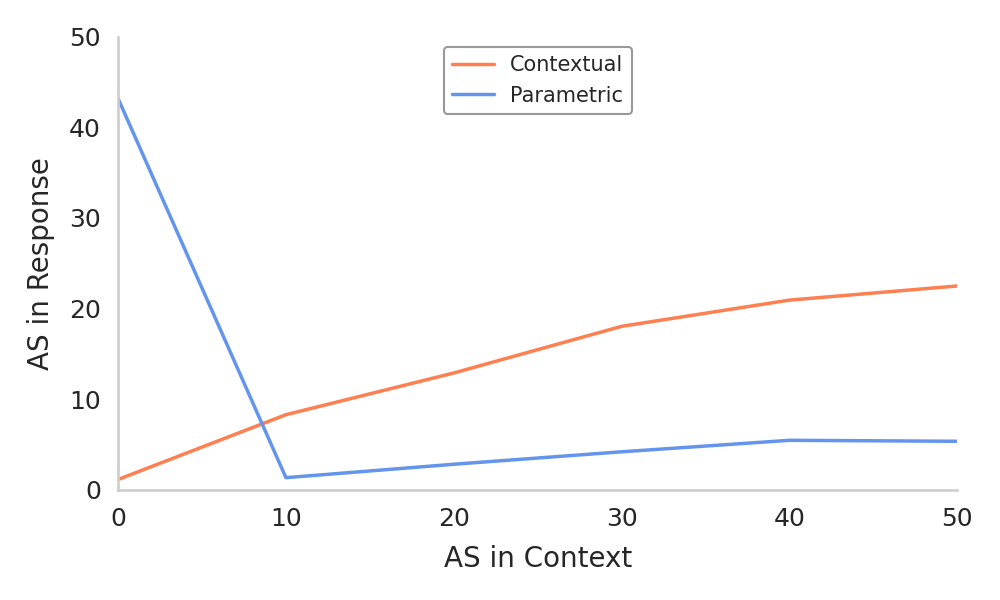}
    \caption{Strict Context Only Instruction}
    \end{subfigure}
    \hfill
    \begin{subfigure}[t]{0.32\textwidth}
    \includegraphics[width=\textwidth]{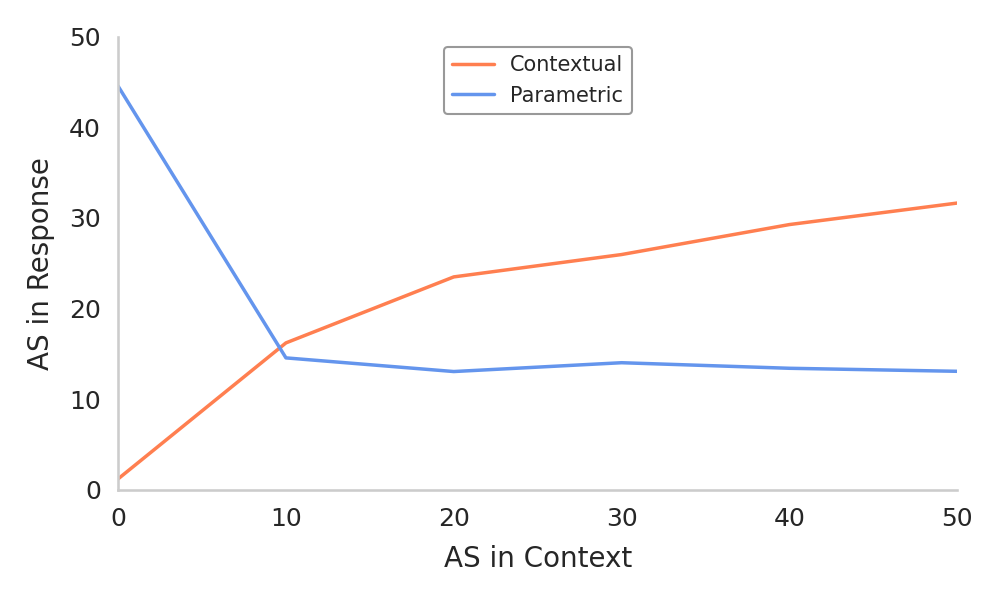}
    \caption{Balanced Use of Context Instruction}
    \end{subfigure}
    \hfill
    \begin{subfigure}[t]{0.32\textwidth}
    \includegraphics[width=\textwidth]{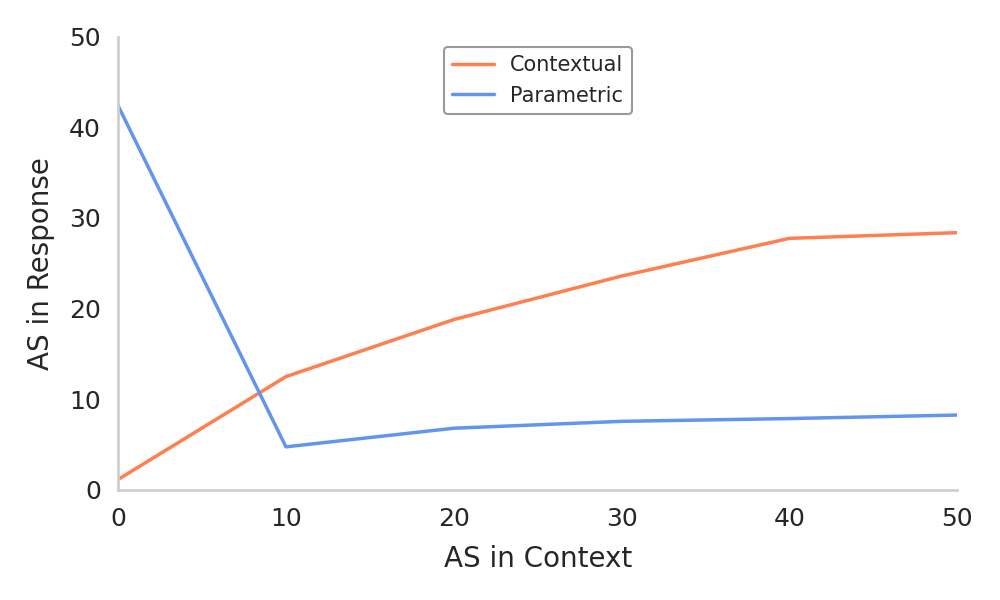}
    \caption{CK Prompt Instruction}
    \end{subfigure}
    \hfill

    \begin{subfigure}[t]{0.32\textwidth}
    \includegraphics[width=\textwidth]{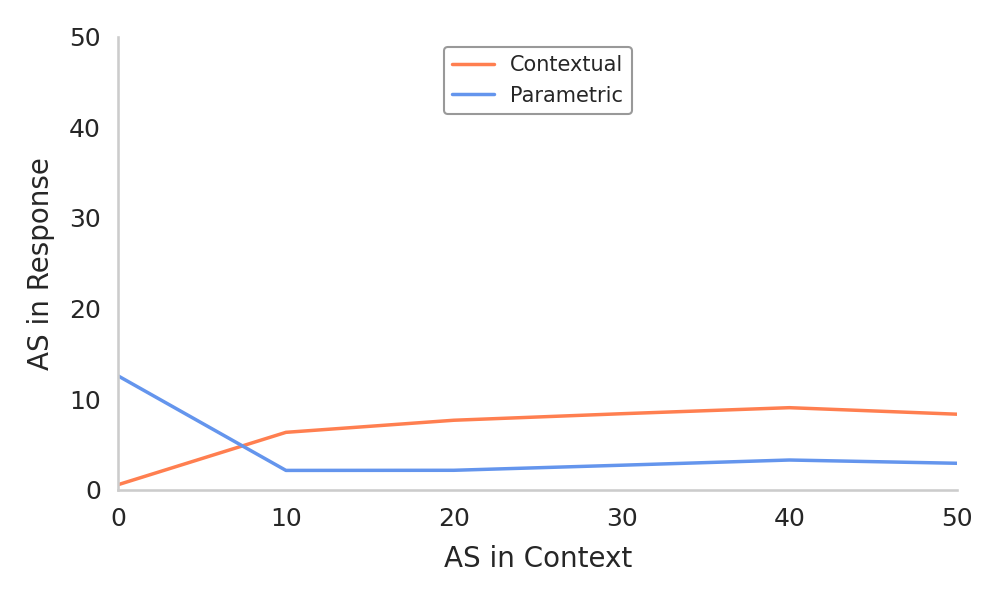}
    \caption{CoT}
    \end{subfigure}
    \begin{subfigure}[t]{0.32\textwidth}
    \includegraphics[width=\textwidth]{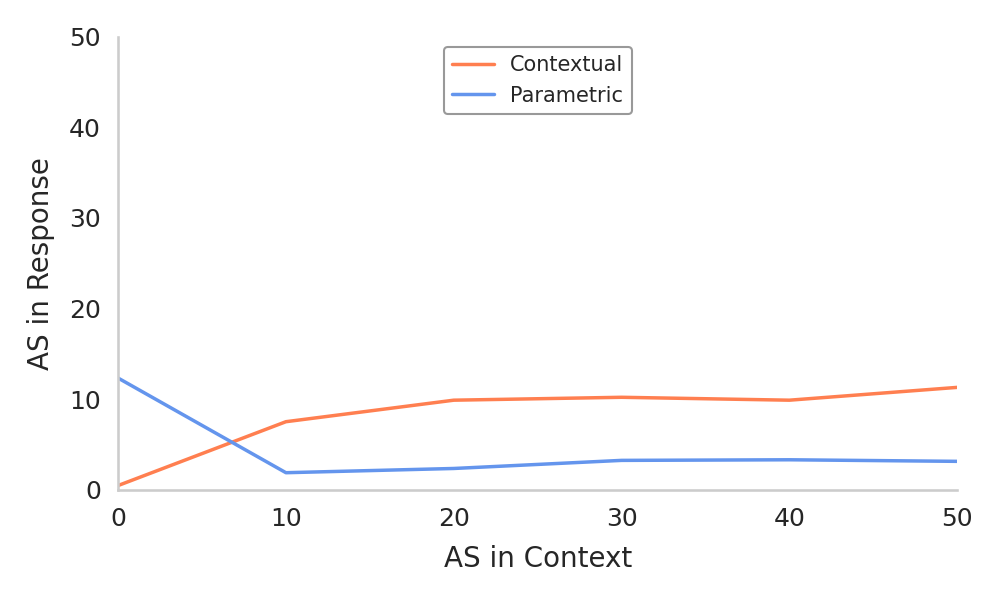}
    \caption{CoT + CK Prompt Instruction}
    \end{subfigure}
    
    \begin{subfigure}[t]{0.32\textwidth}
    \includegraphics[width=\textwidth]{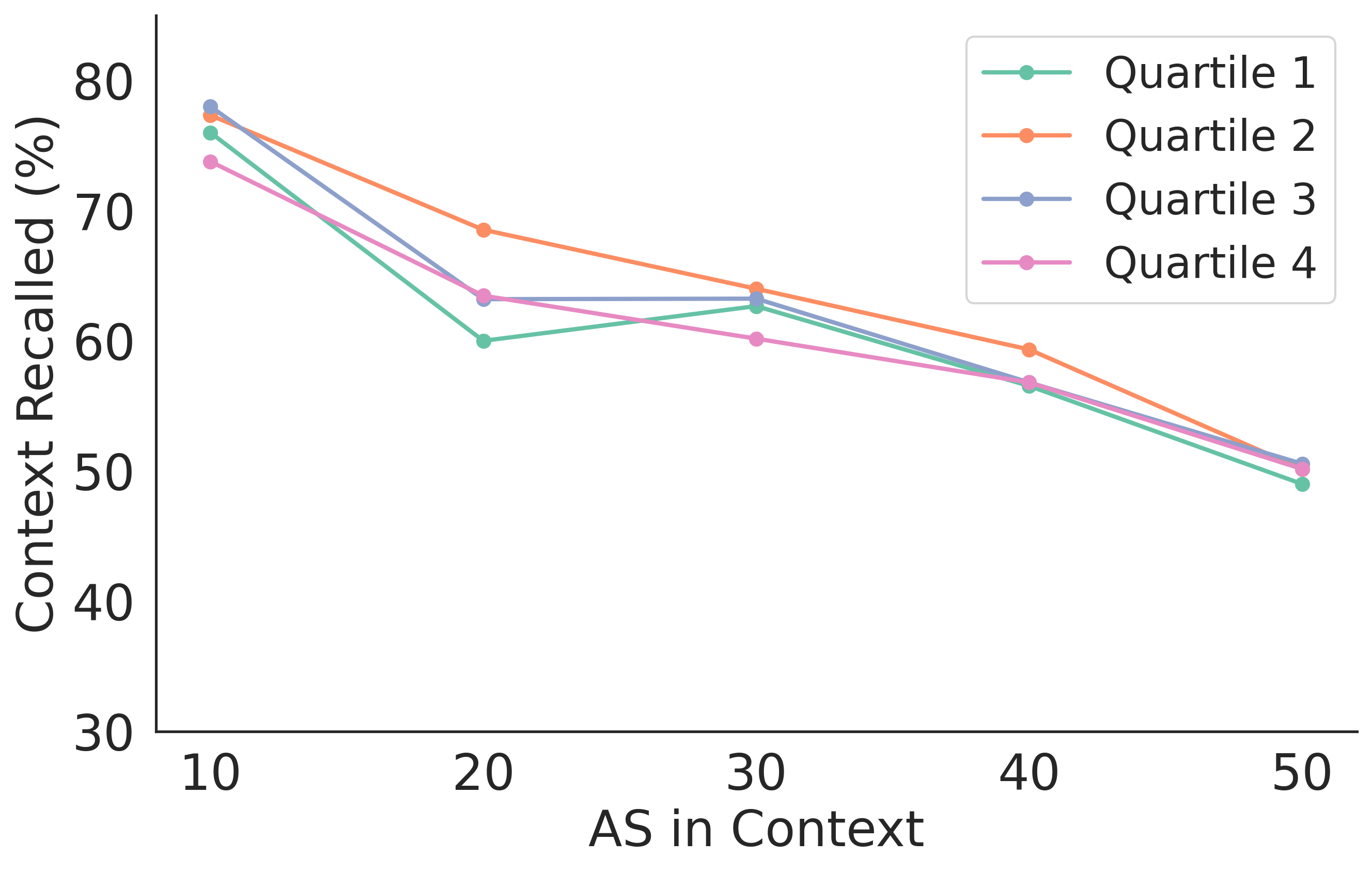}
    \caption{Strict Context Only Instruction}
    \end{subfigure}
    \hfill
    \begin{subfigure}[t]{0.32\textwidth}
    \includegraphics[width=\textwidth]{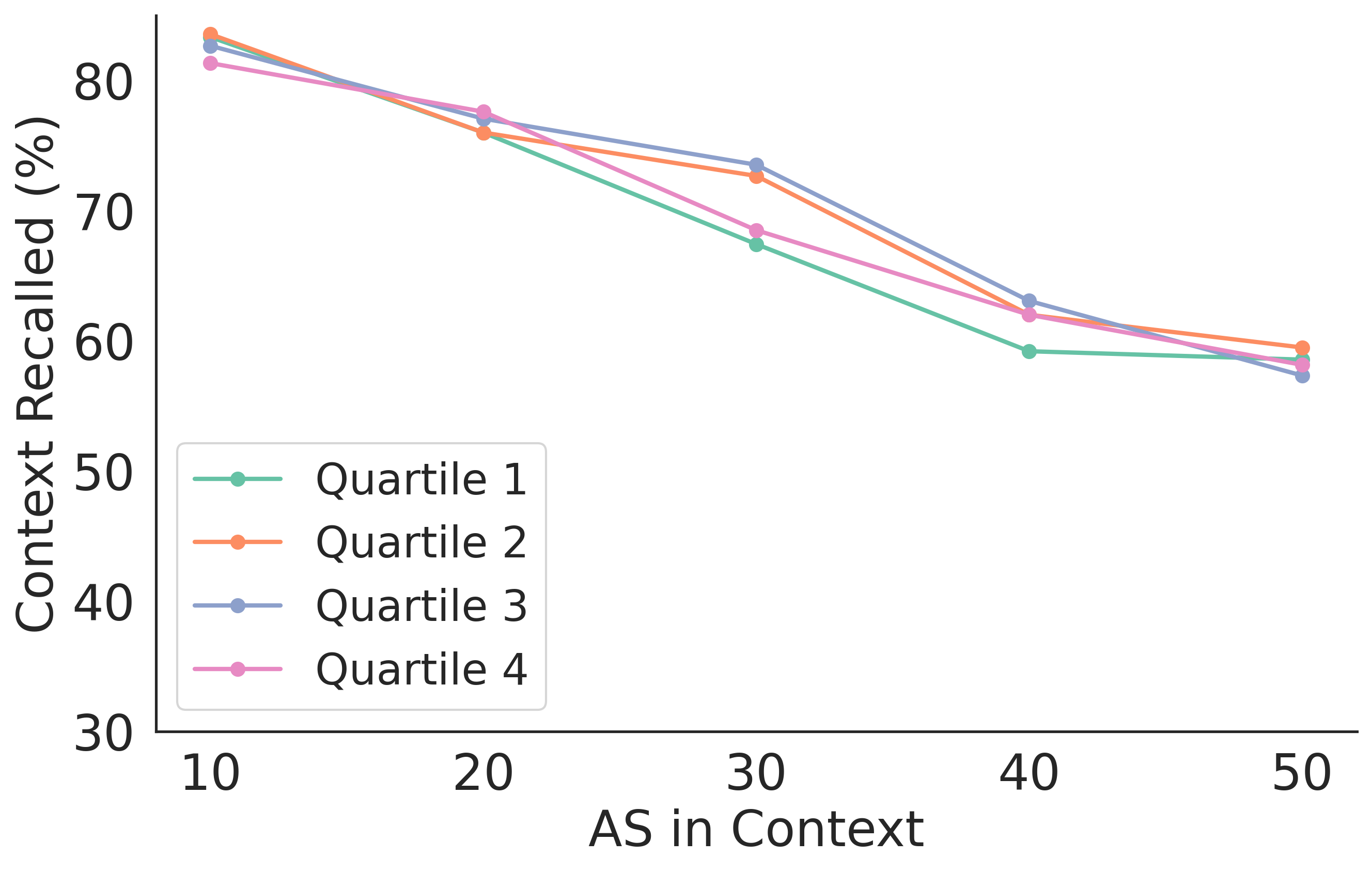}
    \caption{Balanced Use of Context Instruction}
    \end{subfigure}
    \hfill
    \begin{subfigure}[t]{0.32\textwidth}
    \includegraphics[width=\textwidth]{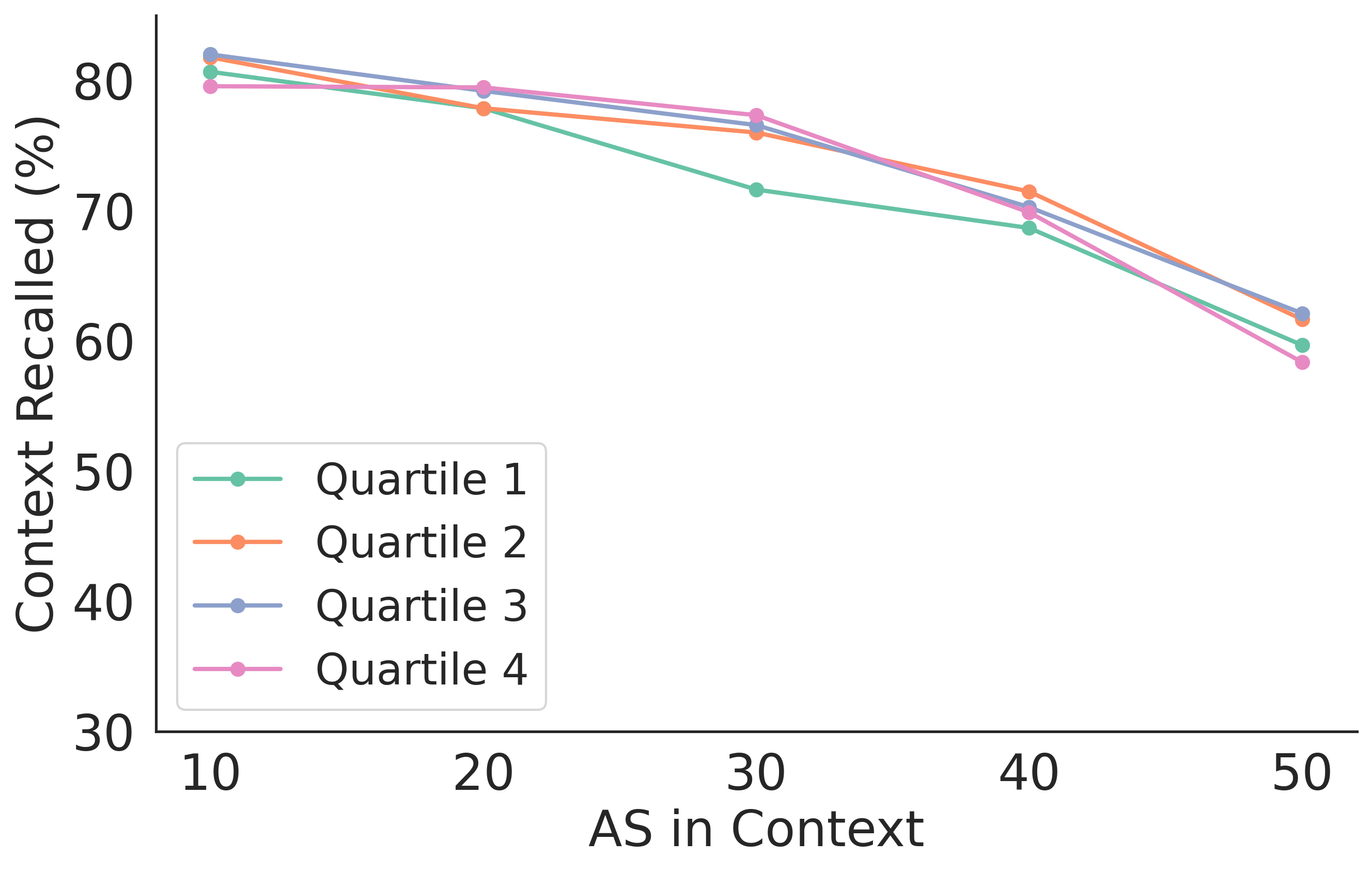}
    \caption{CK Prompt Instruction}
    \end{subfigure}

    \begin{subfigure}[t]{0.32\textwidth}
    \includegraphics[width=\textwidth]{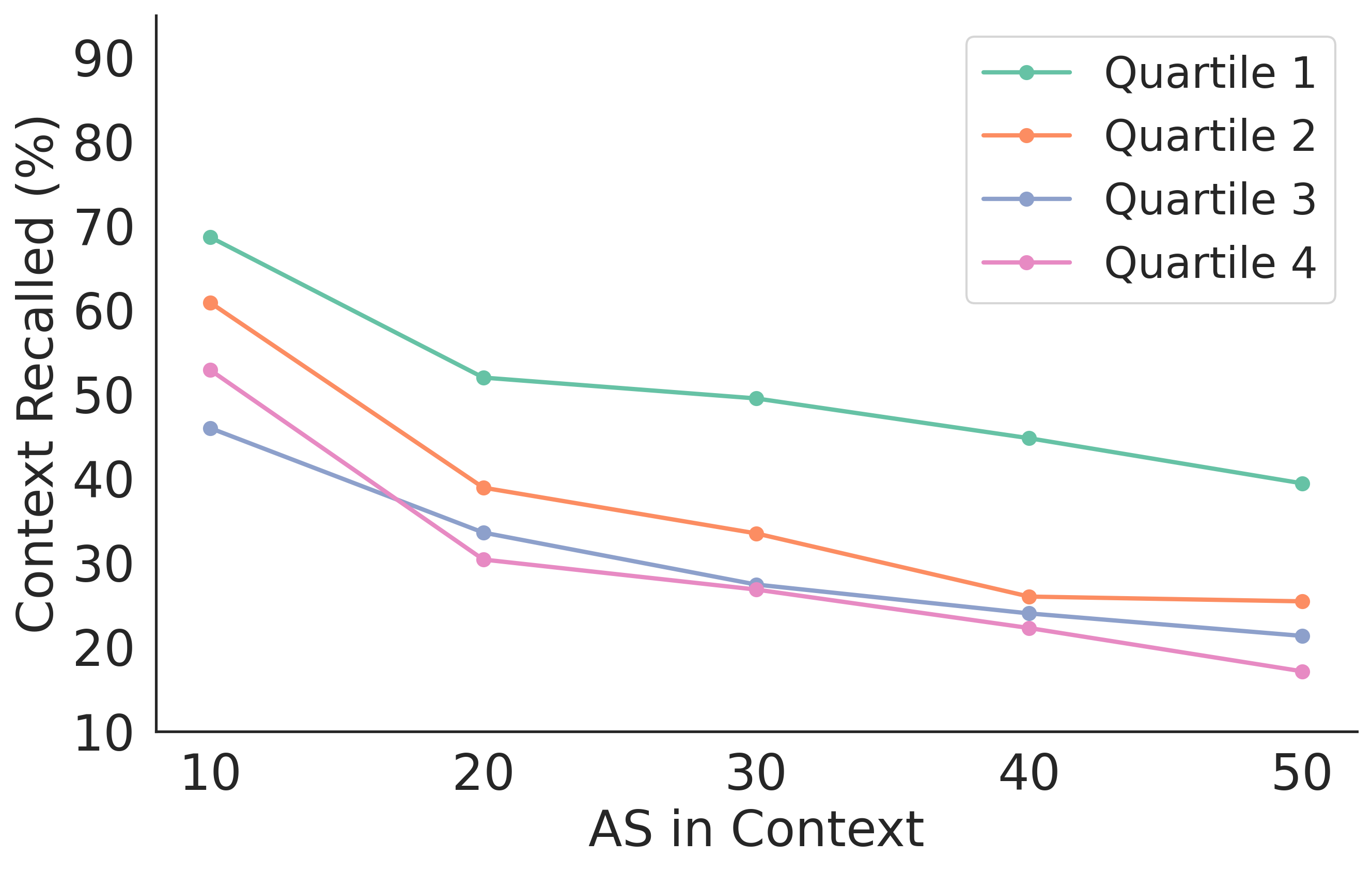}
    \caption{CoT}
    \end{subfigure}
    \begin{subfigure}[t]{0.32\textwidth}
    \includegraphics[width=\textwidth]{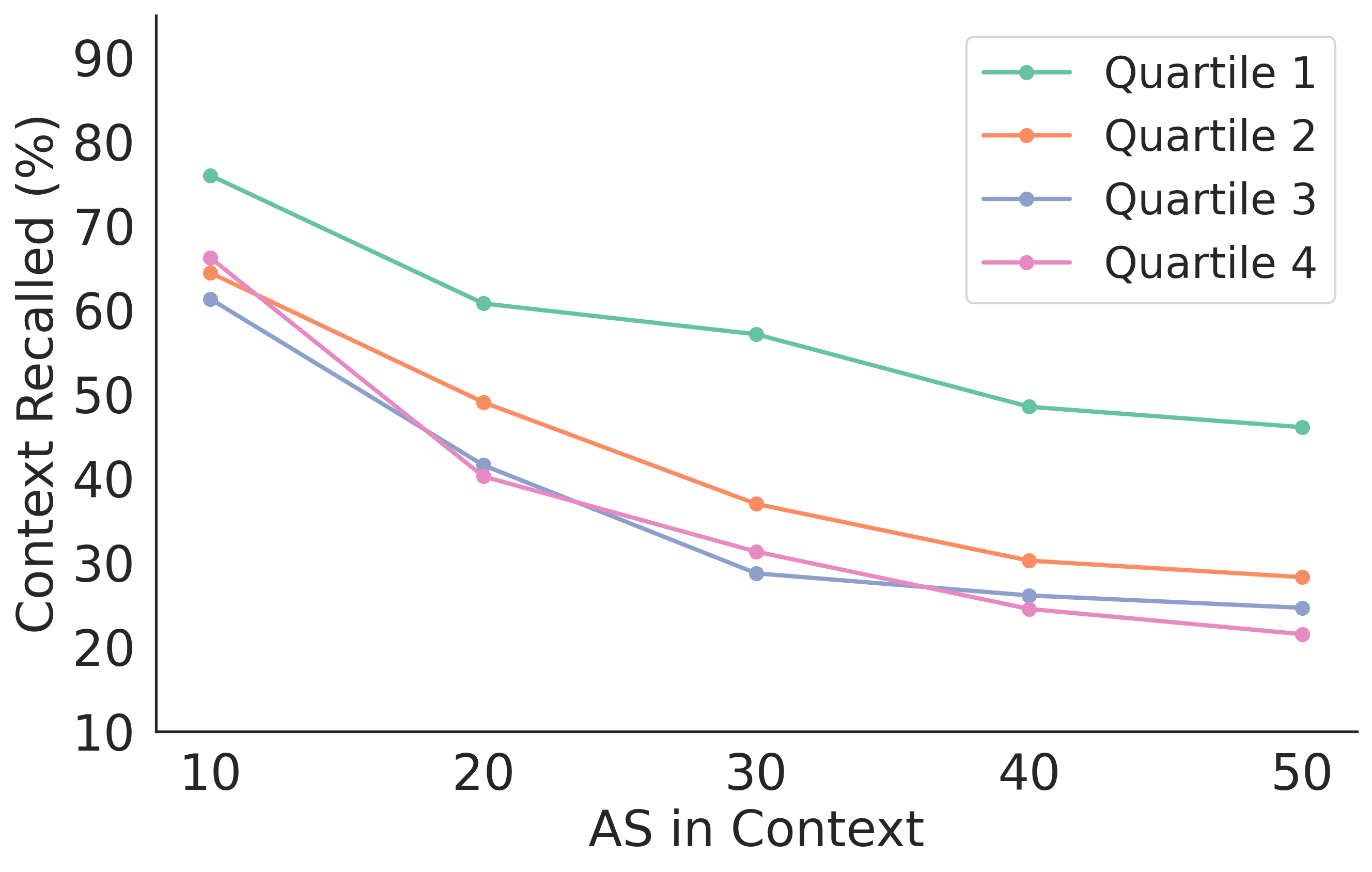}
    \caption{CoT + CK Prompt Instruction}
    \end{subfigure}
    \caption{Llama 3.2 90B average number of CK/PK in responses (Top) and context recall (Bottom) across Different Instructions in Spanish}
    \label{fig:different_instruction_llama3290b_spanish}
\end{figure*}

\begin{figure*}[t!]
    \centering
    \begin{subfigure}[t]{0.32\textwidth}
    \includegraphics[width=1\textwidth]{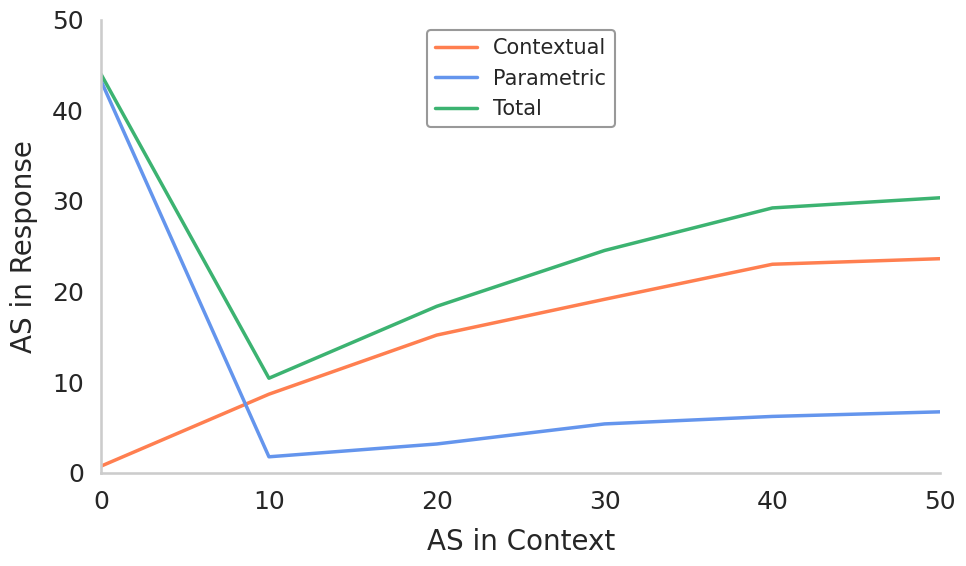}
    \caption{Strict Context Only Instruction}
    \end{subfigure}
    \hfill
    \begin{subfigure}[t]{0.32\textwidth}
    \includegraphics[width=\textwidth]{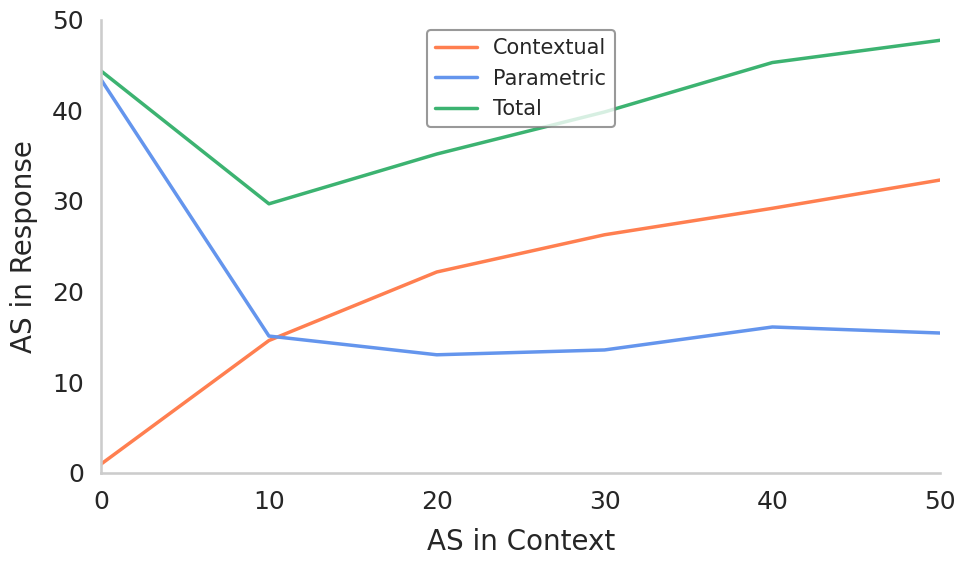}
    \caption{Balanced Use of Context Instruction}
    \end{subfigure}
    \hfill
    \begin{subfigure}[t]{0.32\textwidth}
    \includegraphics[width=\textwidth]{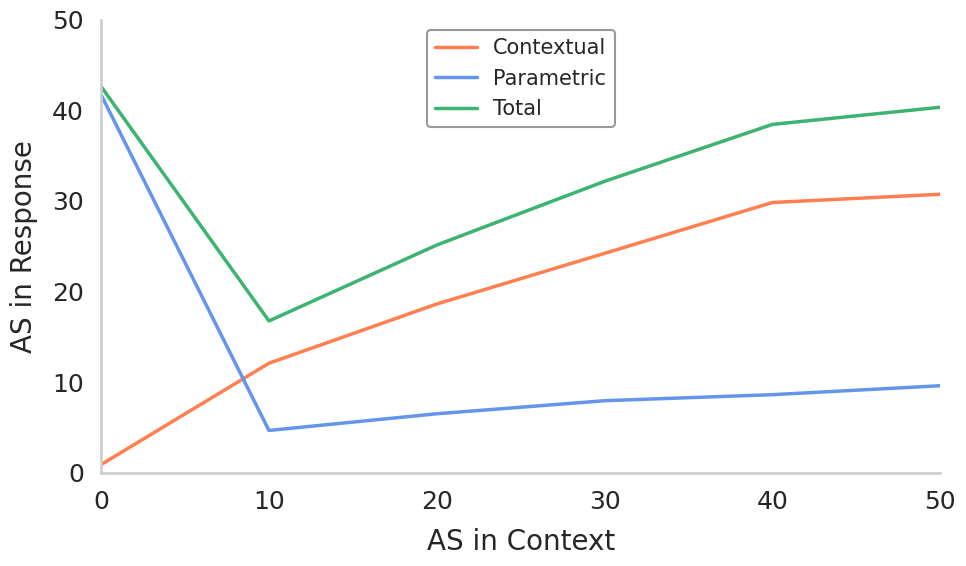}
    \caption{CK Prompt Instruction}
    \end{subfigure}
    \hfill

    \begin{subfigure}[t]{0.32\textwidth}
    \includegraphics[width=\textwidth]{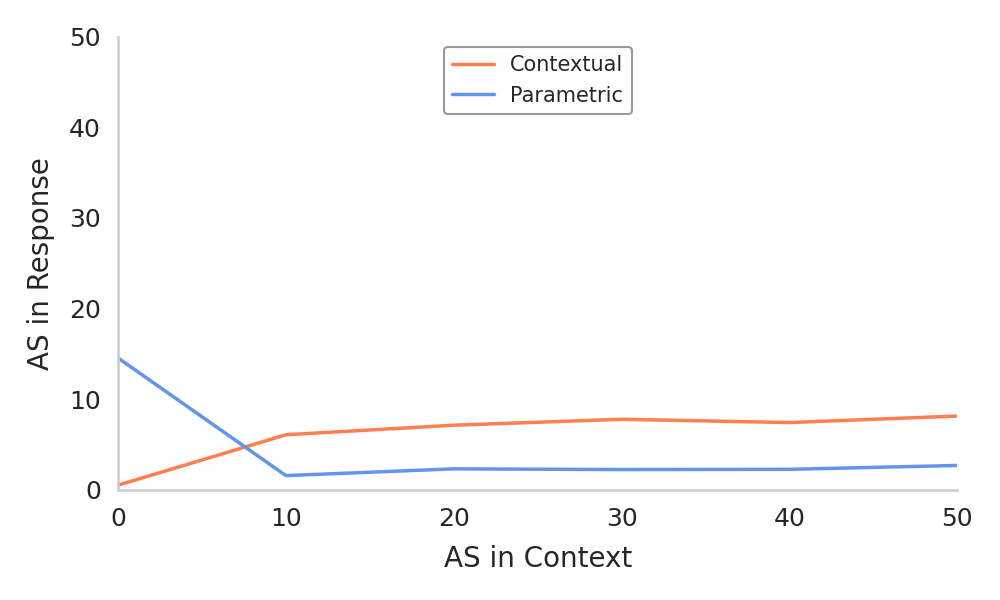}
    \caption{CoT}
    \end{subfigure}
    \begin{subfigure}[t]{0.32\textwidth}
    \includegraphics[width=\textwidth]{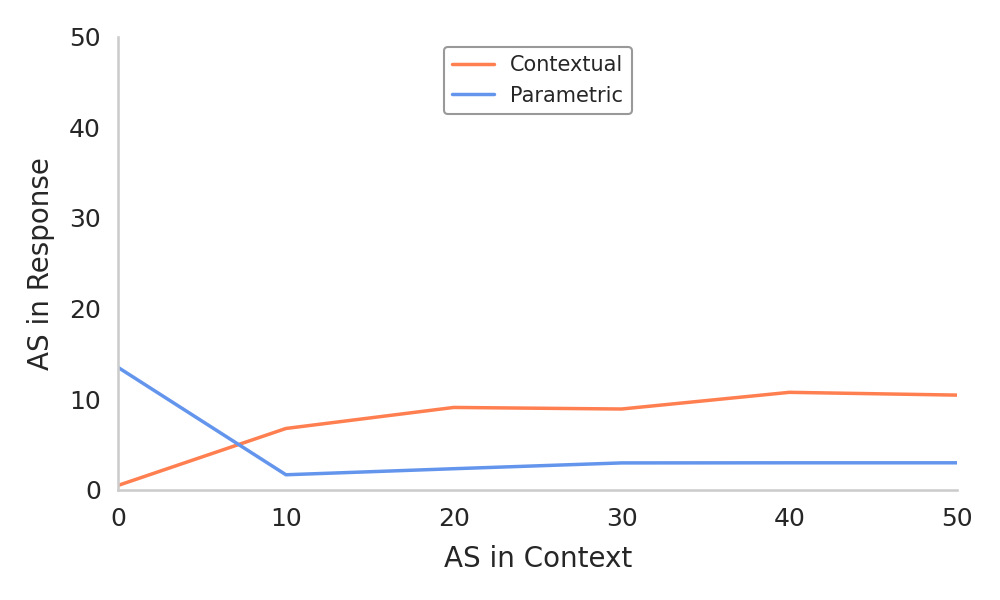}
    \caption{CoT + CK Prompt Instruction}
    \end{subfigure}
    
    \begin{subfigure}[t]{0.32\textwidth}
    \includegraphics[width=\textwidth]{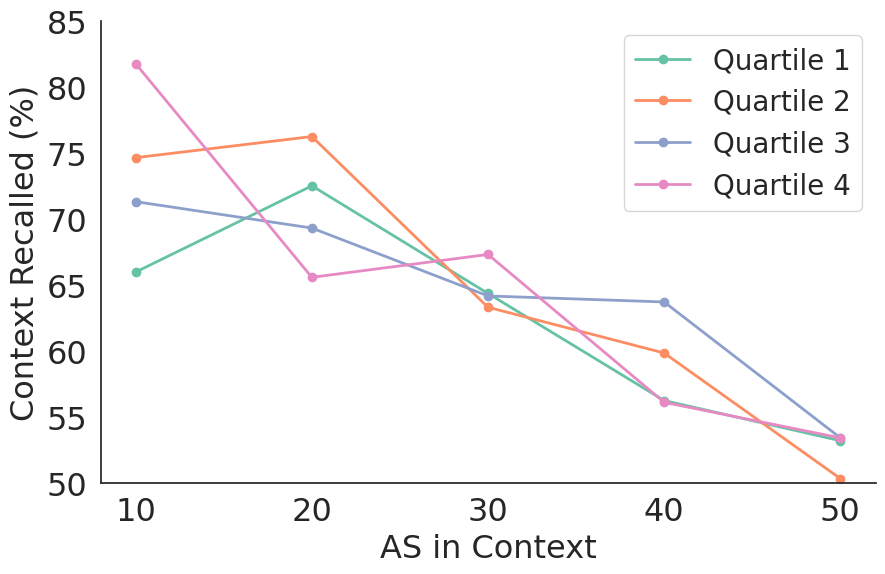}
    \caption{Strict Context Only Instruction}
    \end{subfigure}
    \hfill
    \begin{subfigure}[t]{0.32\textwidth}
    \includegraphics[width=\textwidth]{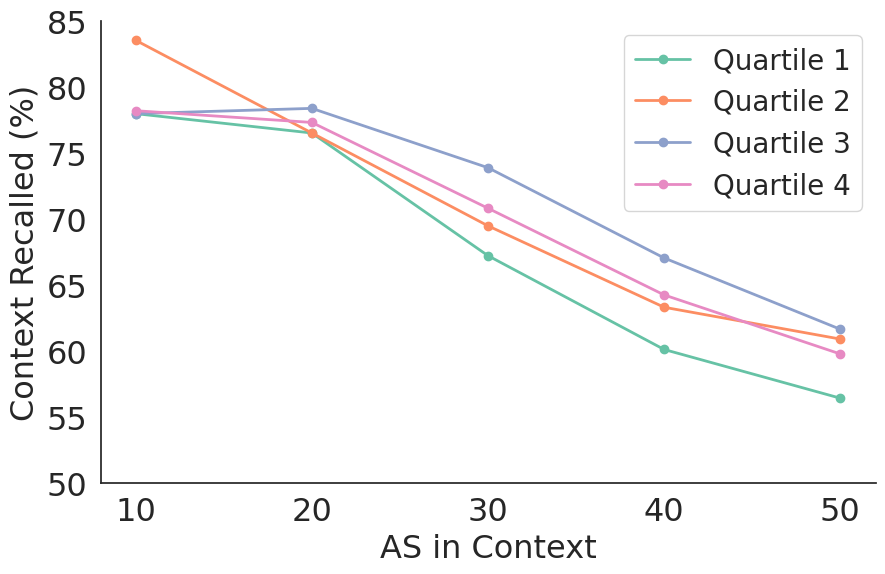}
    \caption{Balanced Use of Context Instruction}
    \end{subfigure}
    \hfill
    \begin{subfigure}[t]{0.32\textwidth}
    \includegraphics[width=\textwidth]{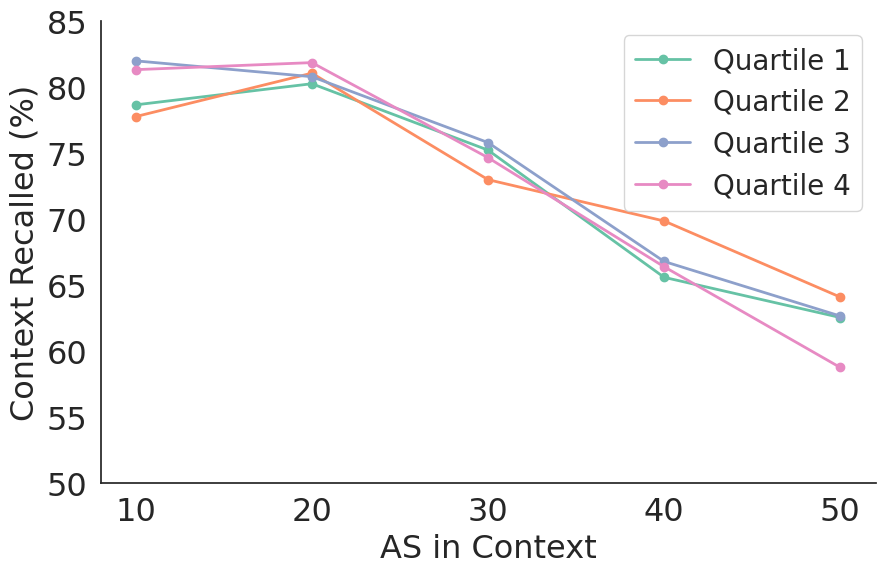}
    \caption{CK Prompt Instruction}
    \end{subfigure}

    \begin{subfigure}[t]{0.32\textwidth}
    \includegraphics[width=\textwidth]{figures/different_instruction/context_recall/llama3290b/da_combined_llama3290b.png}
    \caption{CoT}
    \end{subfigure}
    \begin{subfigure}[t]{0.32\textwidth}
    \includegraphics[width=\textwidth]{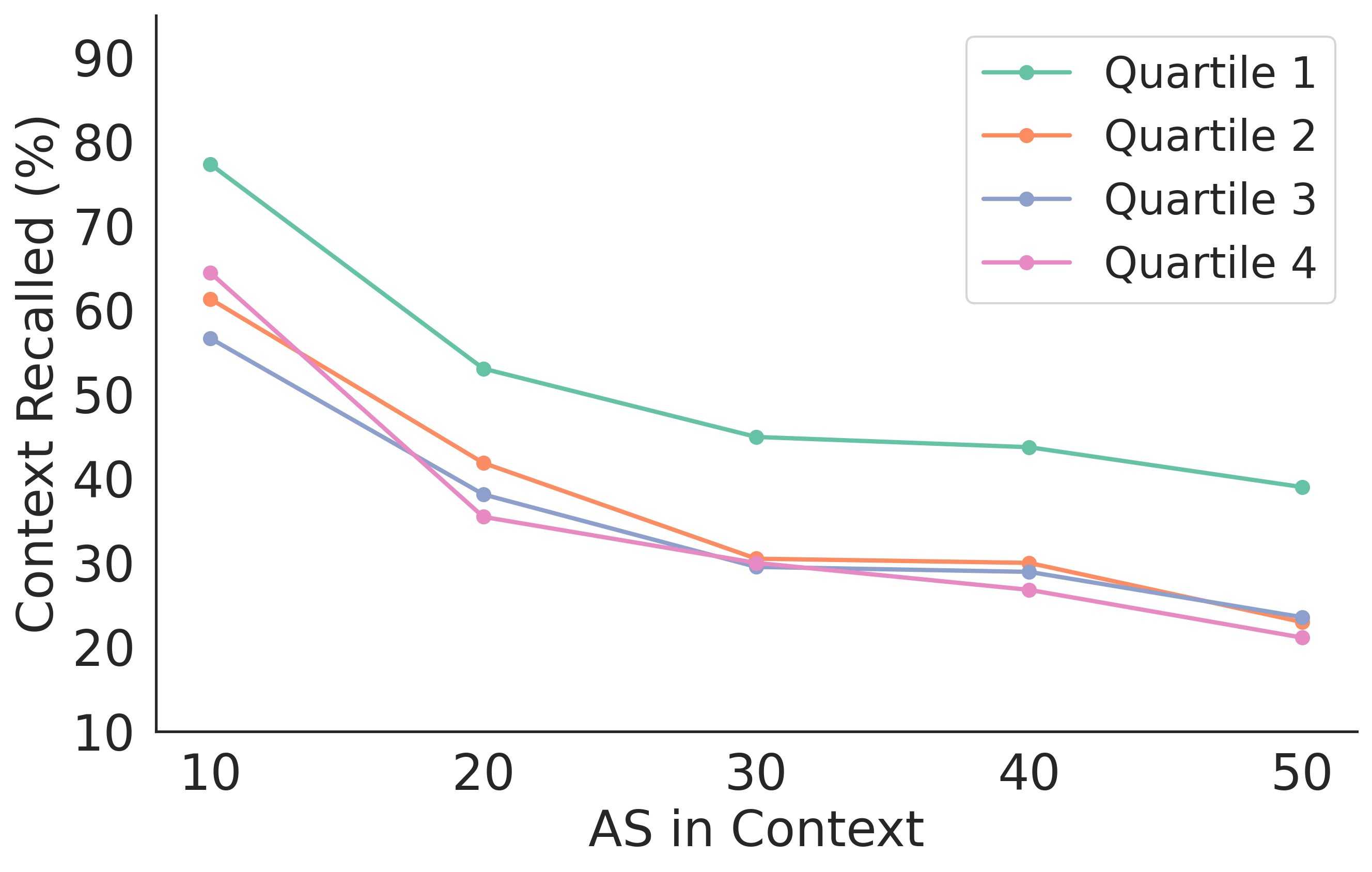}
    \caption{CoT + CK Prompt Instruction}
    \end{subfigure}
    
    \caption{Llama 3.2 90B average number of CK/PK in responses (Top) and context recall (Bottom) across Different Instructions in Danish}
    \label{fig:different_instruction_llama3290b_danish}
\end{figure*}

\begin{figure*}[t!]
    \centering
    \begin{subfigure}[t]{0.32\textwidth}
    \includegraphics[width=1\textwidth]{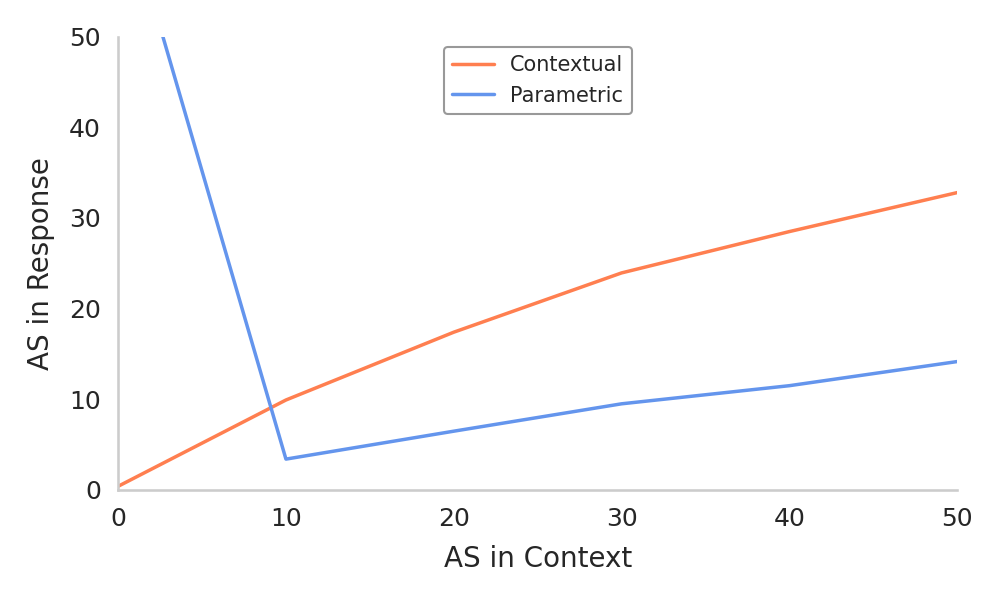}
    \caption{Strict Context Only Instruction}
    \end{subfigure}
    \hfill
    \begin{subfigure}[t]{0.32\textwidth}
    \includegraphics[width=\textwidth]{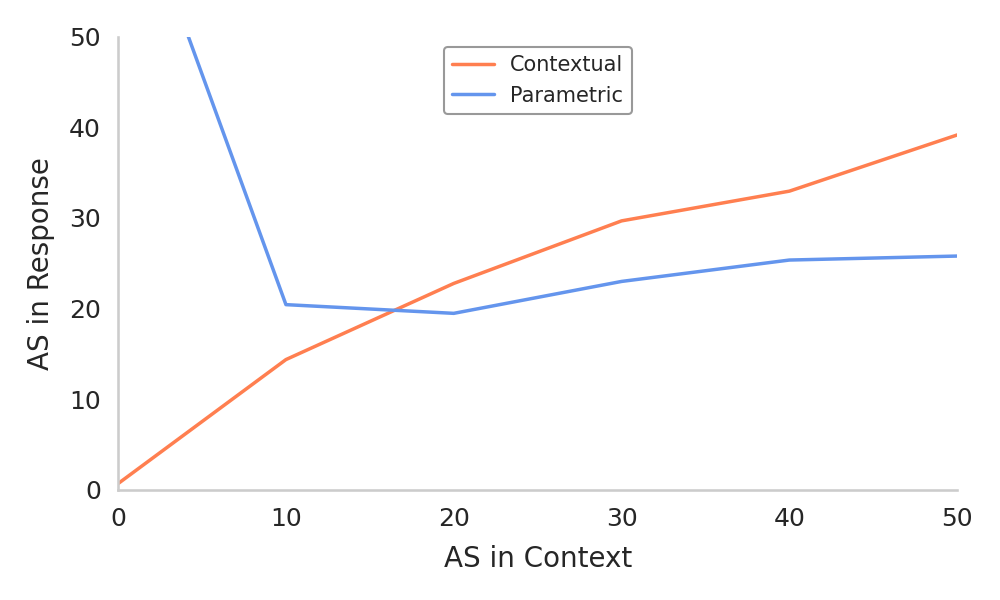}
    \caption{Balanced Use of Context Instruction}
    \end{subfigure}
    \hfill
    \begin{subfigure}[t]{0.32\textwidth}
    \includegraphics[width=\textwidth]{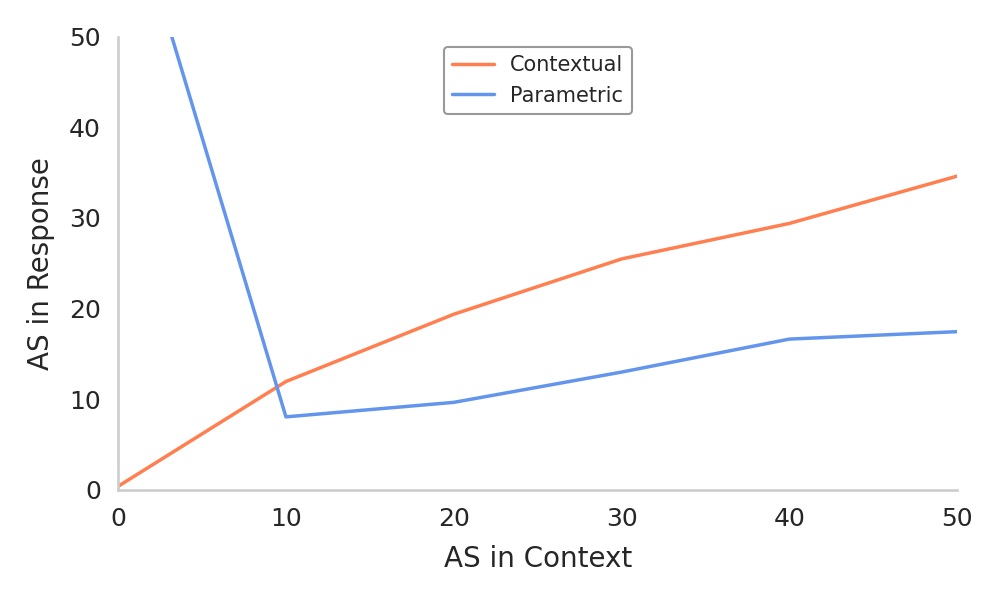}
    \caption{CK Prompt Instruction}
    \end{subfigure}
    \hfill
    \begin{subfigure}[t]{0.32\textwidth}
    \includegraphics[width=\textwidth]{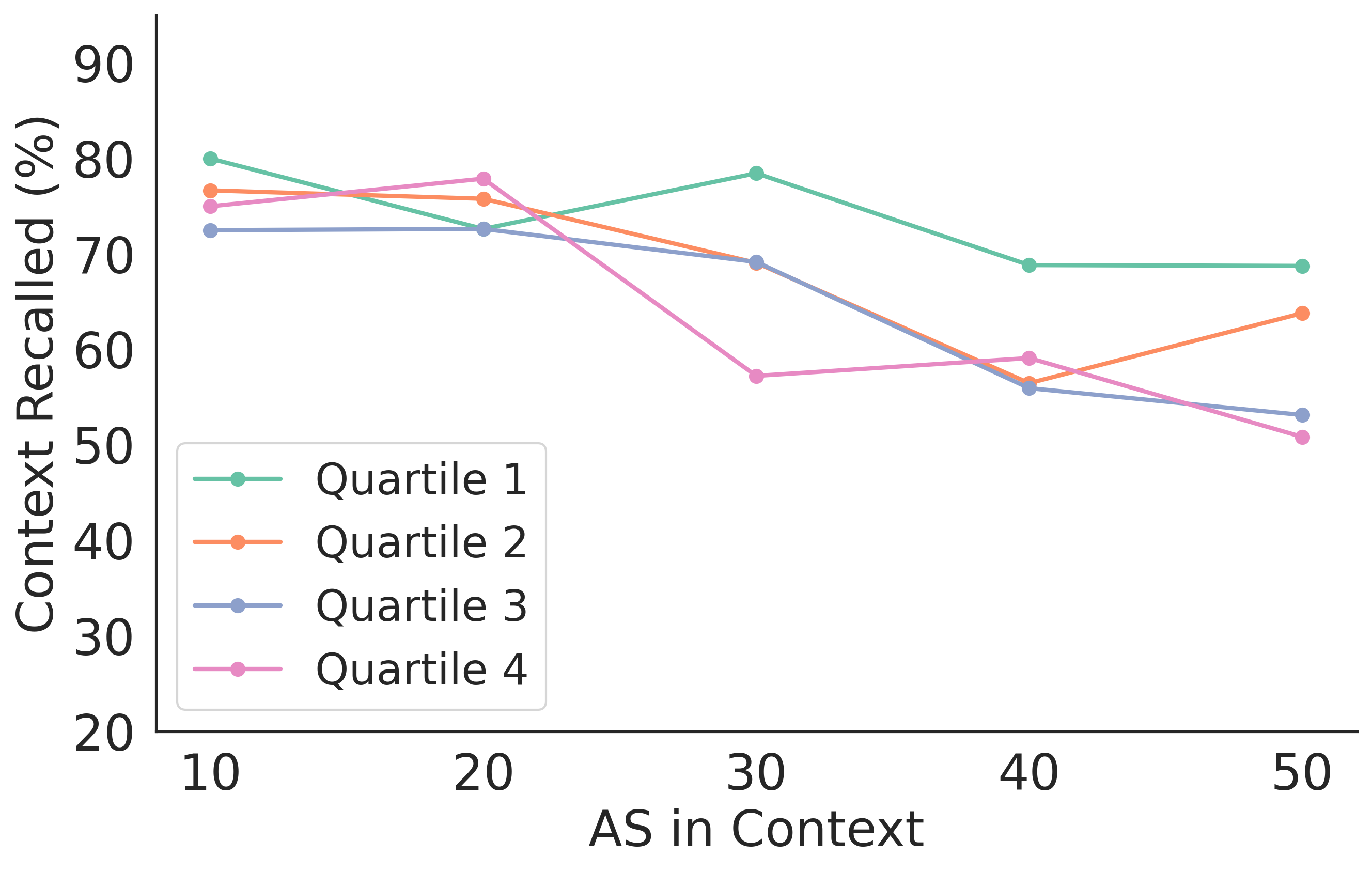}
    \caption{Strict Context Only Instruction}
    \end{subfigure}
    \hfill
    \begin{subfigure}[t]{0.32\textwidth}
    \includegraphics[width=\textwidth]{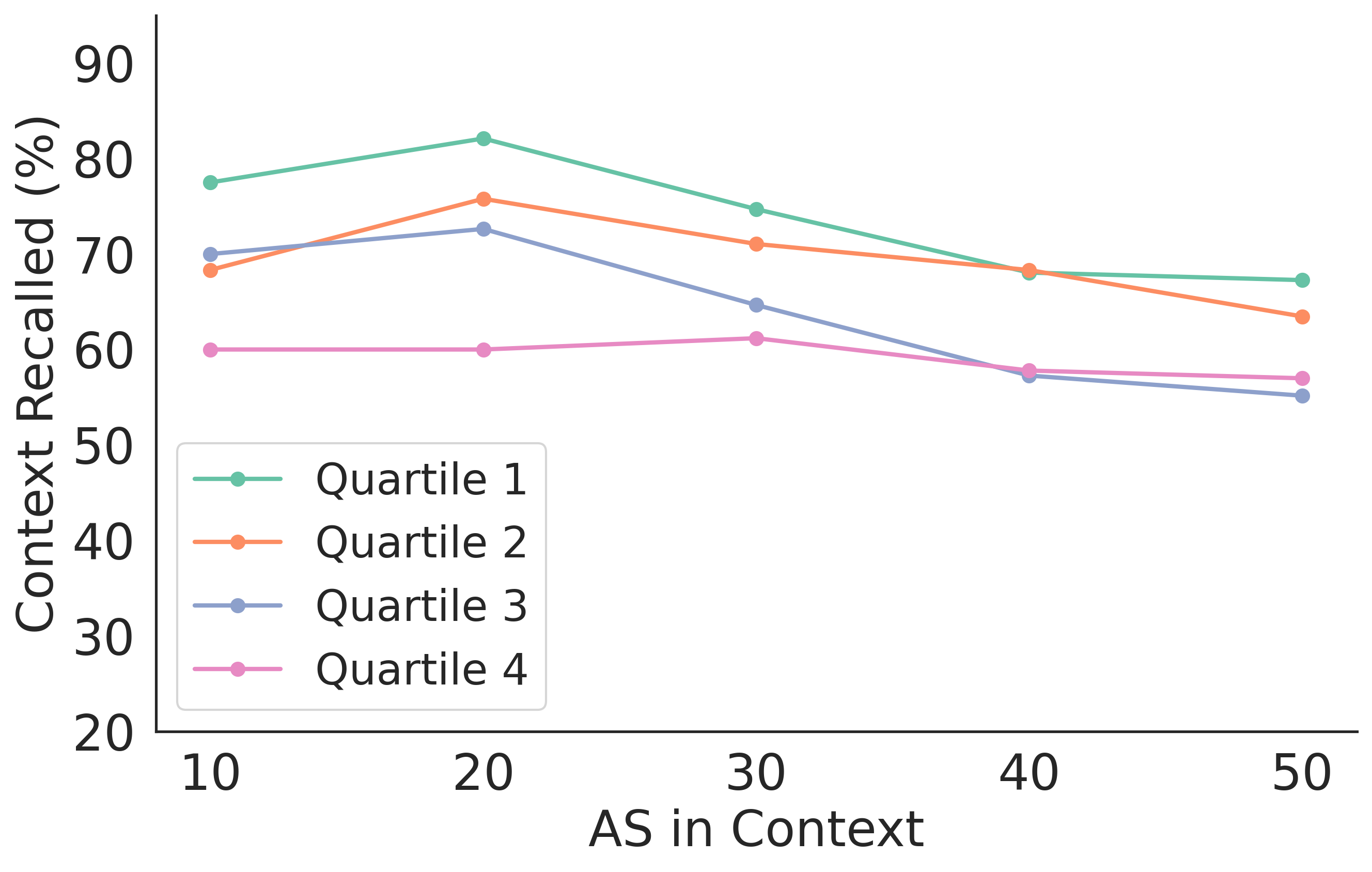}
    \caption{Balanced Use of Context Instruction}
    \end{subfigure}
    \hfill
    \begin{subfigure}[t]{0.32\textwidth}
    \includegraphics[width=\textwidth]{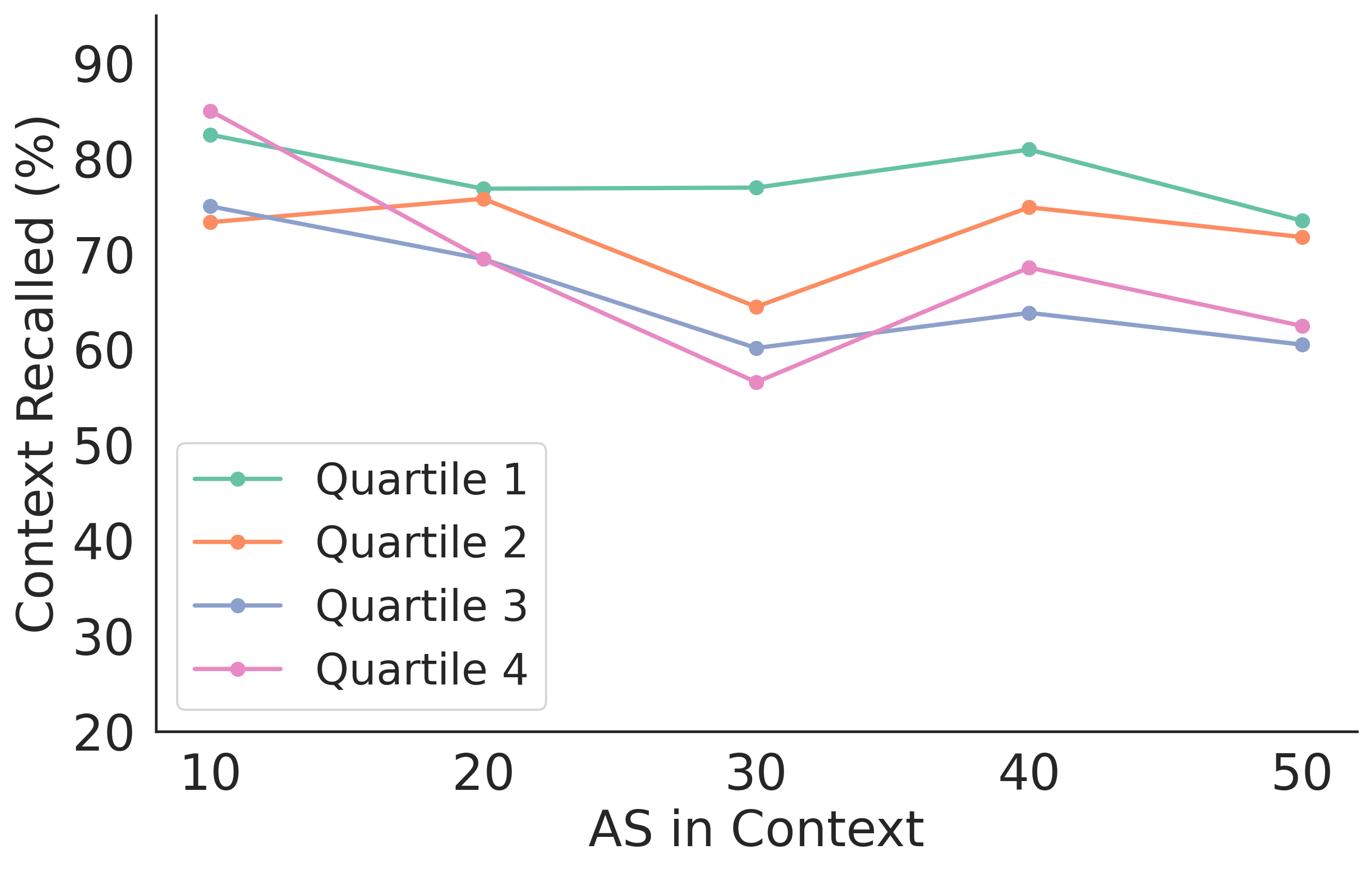}
    \caption{CK Prompt Instruction}
    \end{subfigure}
    \caption{Qwen 3 235B average number of CK/PK in responses (Top) and context recall (Bottom) across Different Instructions in English}
    \label{fig:different_instruction_qwen3235b_english}
\end{figure*}

\begin{figure*}[t!]
    \centering
    \begin{subfigure}[t]{0.32\textwidth}
    \includegraphics[width=1\textwidth]{figures/different_instruction/main/qwen3235b/en_strict_qwen3235b.png}
    \caption{Strict Context Only Instruction}
    \end{subfigure}
    \hfill
    \begin{subfigure}[t]{0.32\textwidth}
    \includegraphics[width=\textwidth]{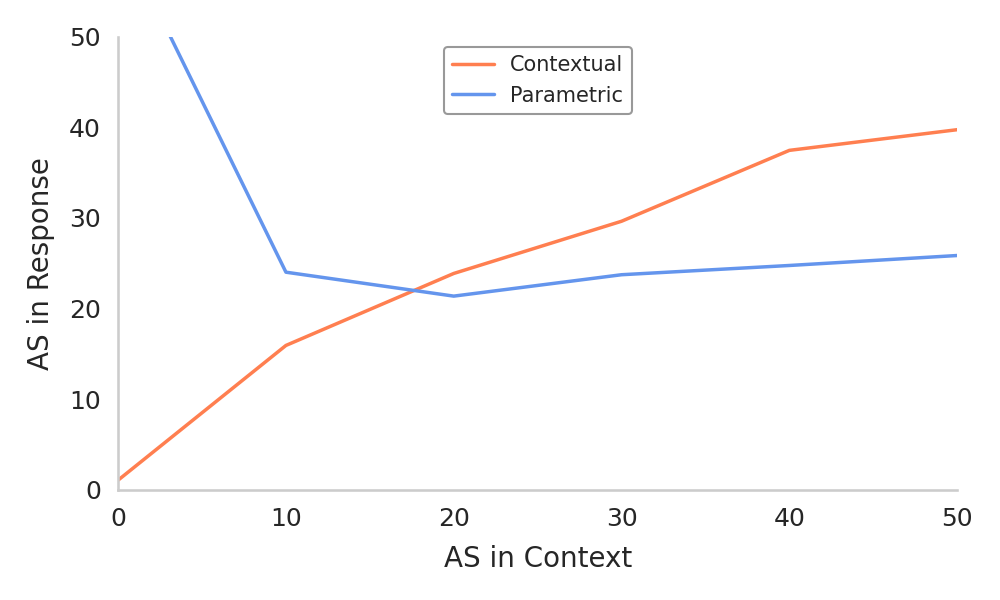}
    \caption{Balanced Use of Context Instruction}
    \end{subfigure}
    \hfill
    \begin{subfigure}[t]{0.32\textwidth}
    \includegraphics[width=\textwidth]{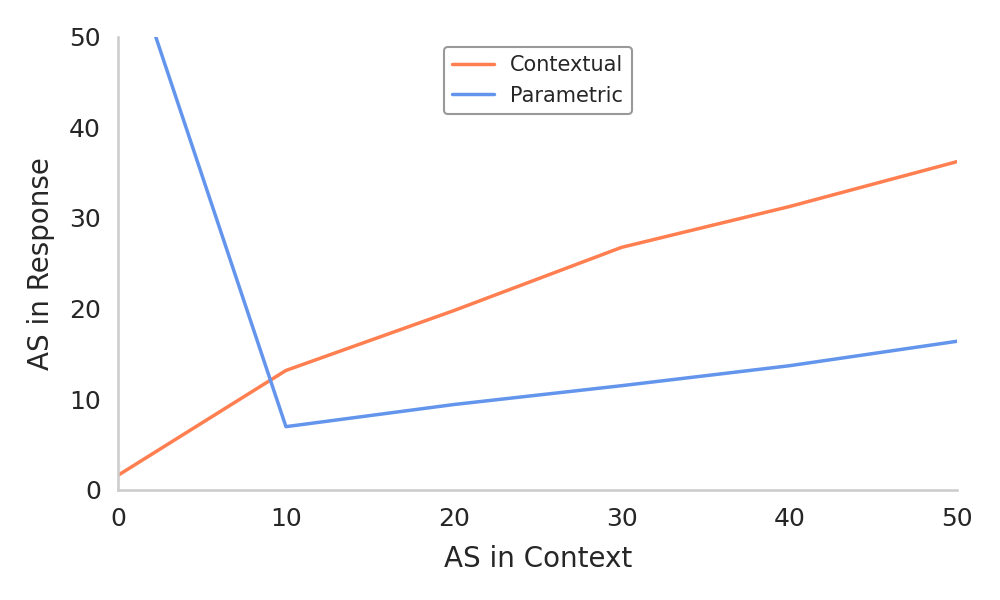}
    \caption{CK Prompt Instruction}
    \end{subfigure}
    \hfill
    \begin{subfigure}[t]{0.32\textwidth}
    \includegraphics[width=\textwidth]{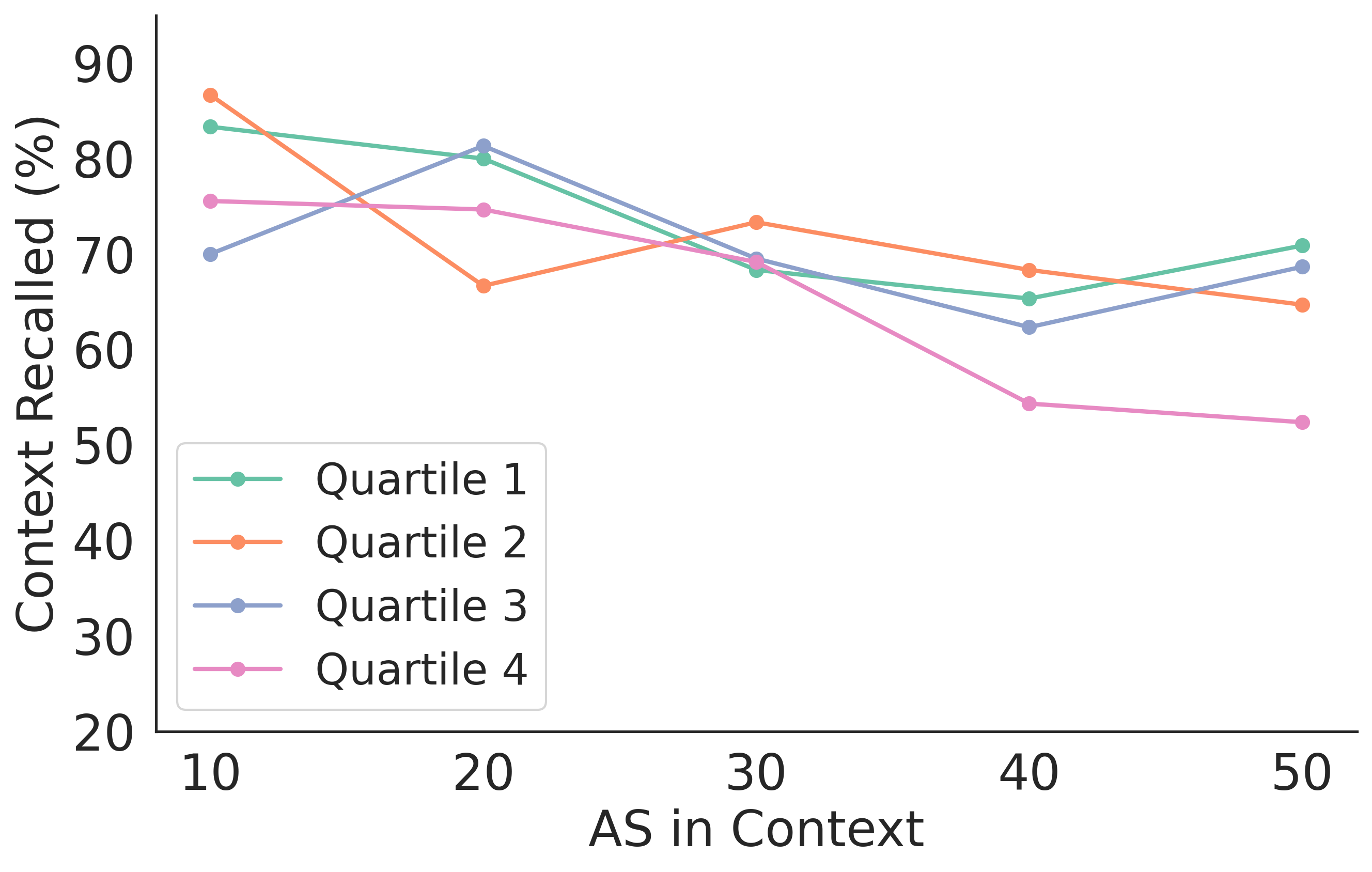}
    \caption{Strict Context Only Instruction}
    \end{subfigure}
    \hfill
    \begin{subfigure}[t]{0.32\textwidth}
    \includegraphics[width=\textwidth]{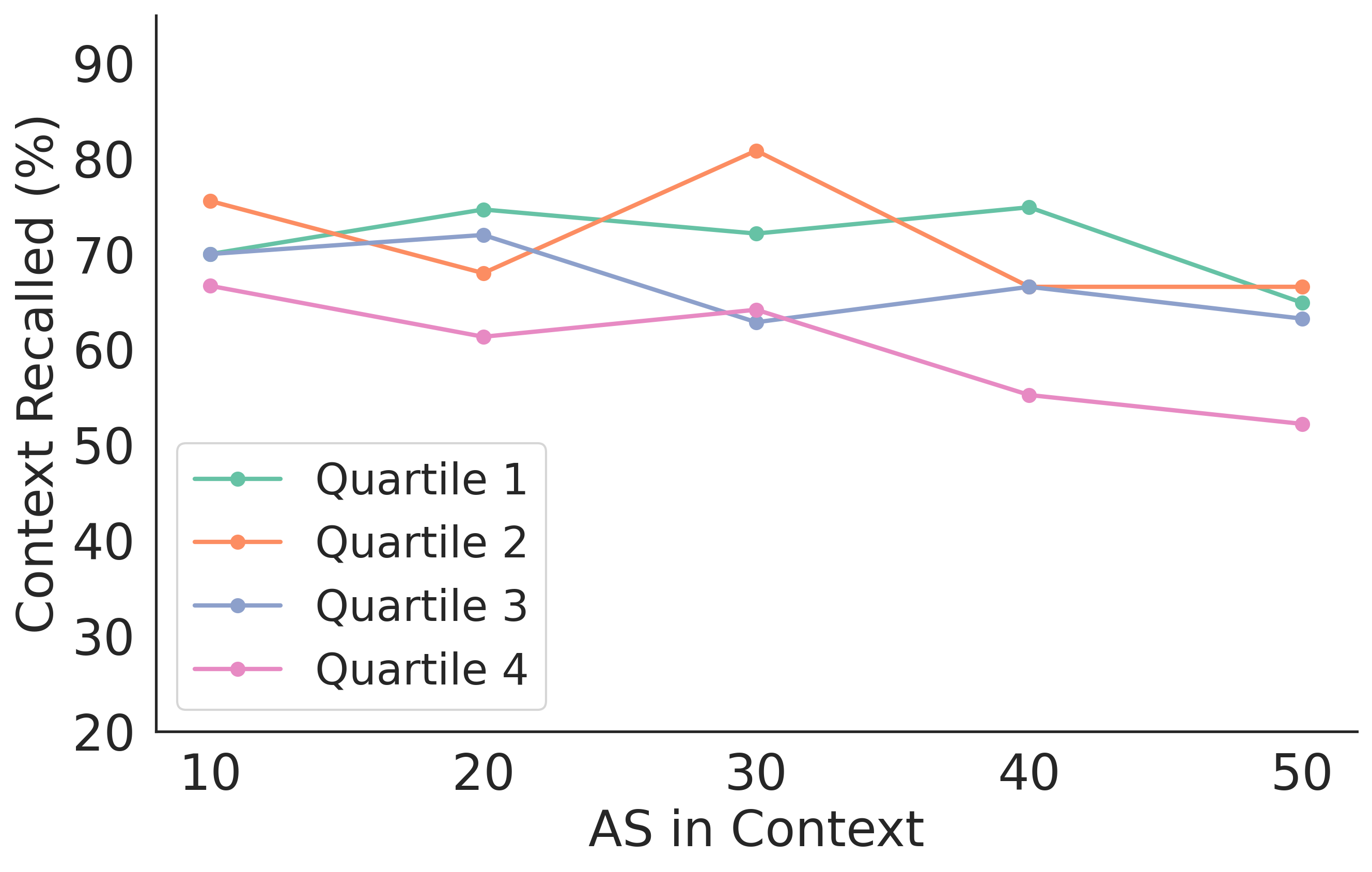}
    \caption{Balanced Use of Context Instruction}
    \end{subfigure}
    \hfill
    \begin{subfigure}[t]{0.32\textwidth}
    \includegraphics[width=\textwidth]{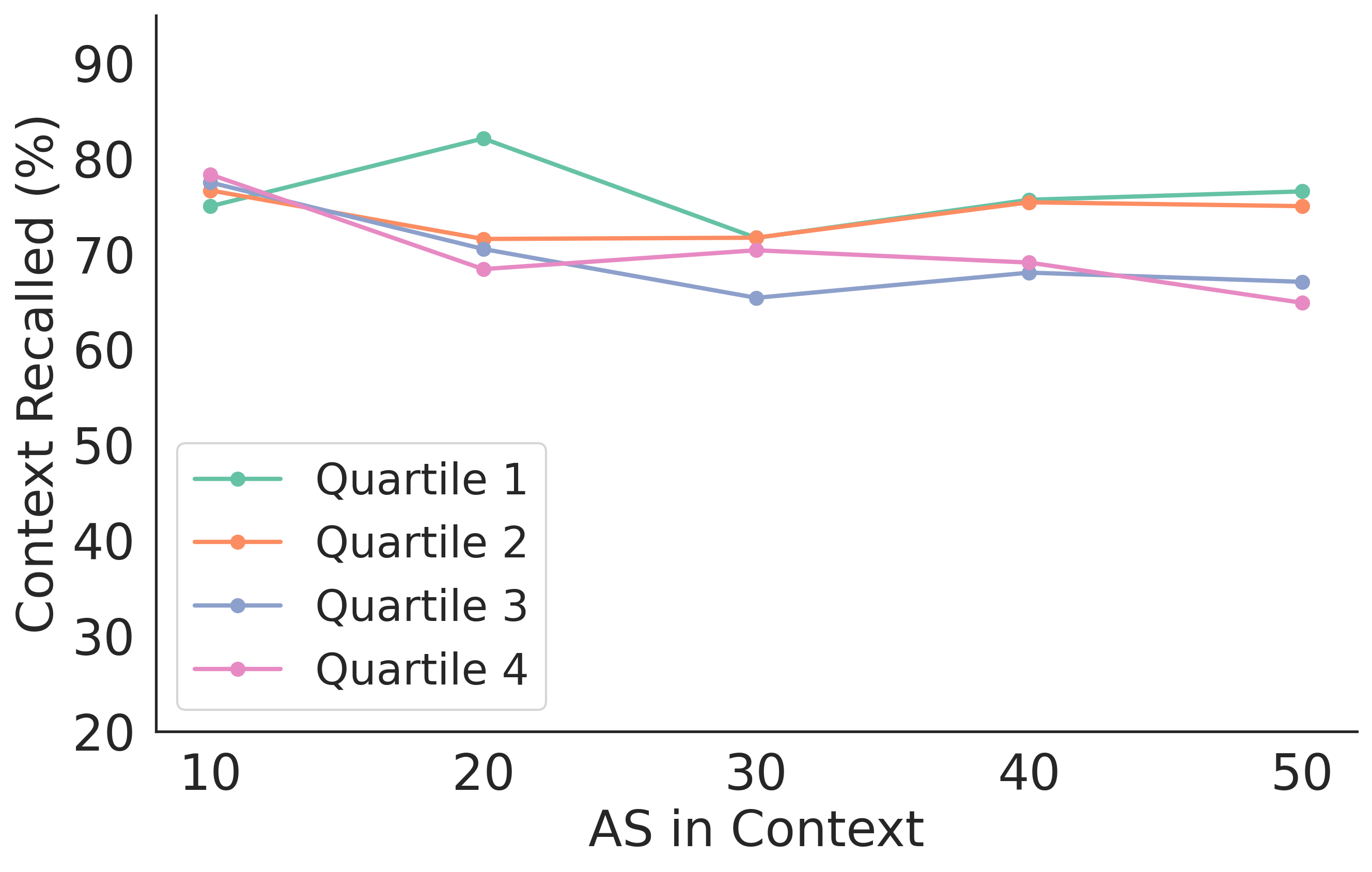}
    \caption{CK Prompt Instruction}
    \end{subfigure}
    \caption{Qwen 3 235B average number of CK/PK in responses (Top) and context recall (Bottom) across Different Instructions in Spanish}
    \label{fig:different_instruction_qwen3235b_spanish}
\end{figure*}

\begin{figure*}[t!]
    \centering
    \begin{subfigure}[t]{0.32\textwidth}
    \includegraphics[width=1\textwidth]{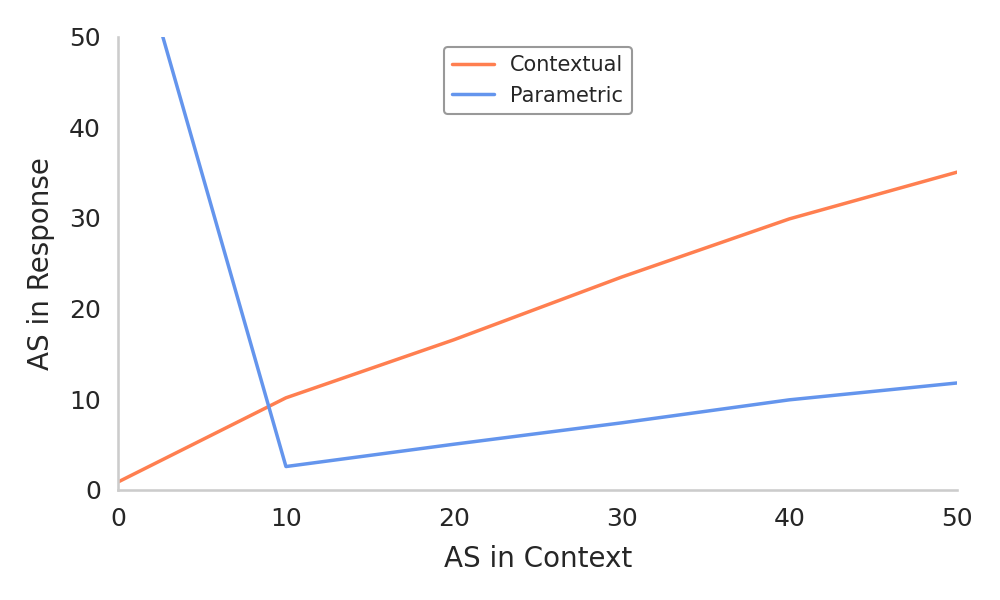}
    \caption{Strict Context Only Instruction}
    \end{subfigure}
    \hfill
    \begin{subfigure}[t]{0.32\textwidth}
    \includegraphics[width=\textwidth]{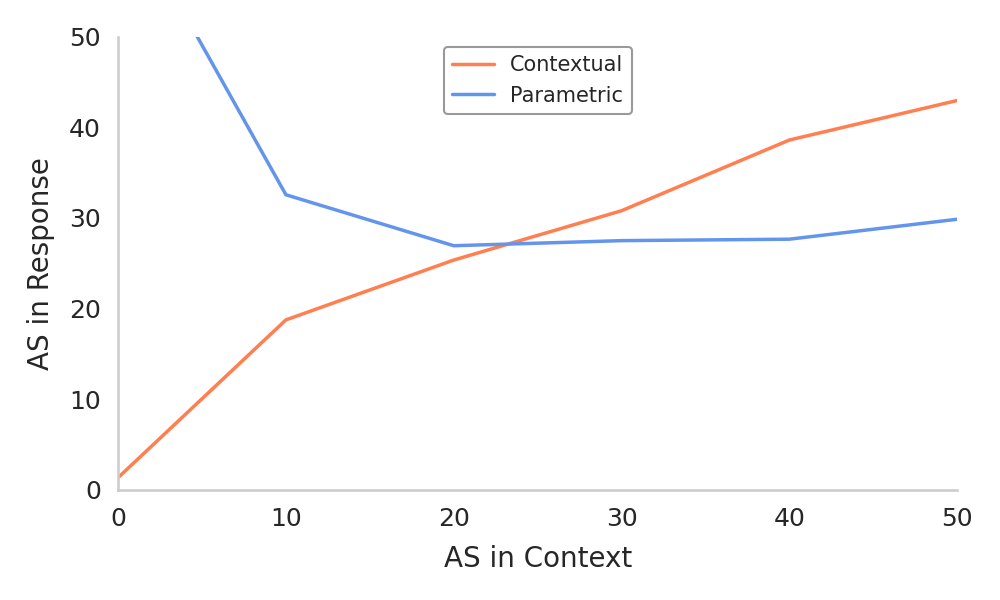}
    \caption{Balanced Use of Context Instruction}
    \end{subfigure}
    \hfill
    \begin{subfigure}[t]{0.32\textwidth}
    \includegraphics[width=\textwidth]{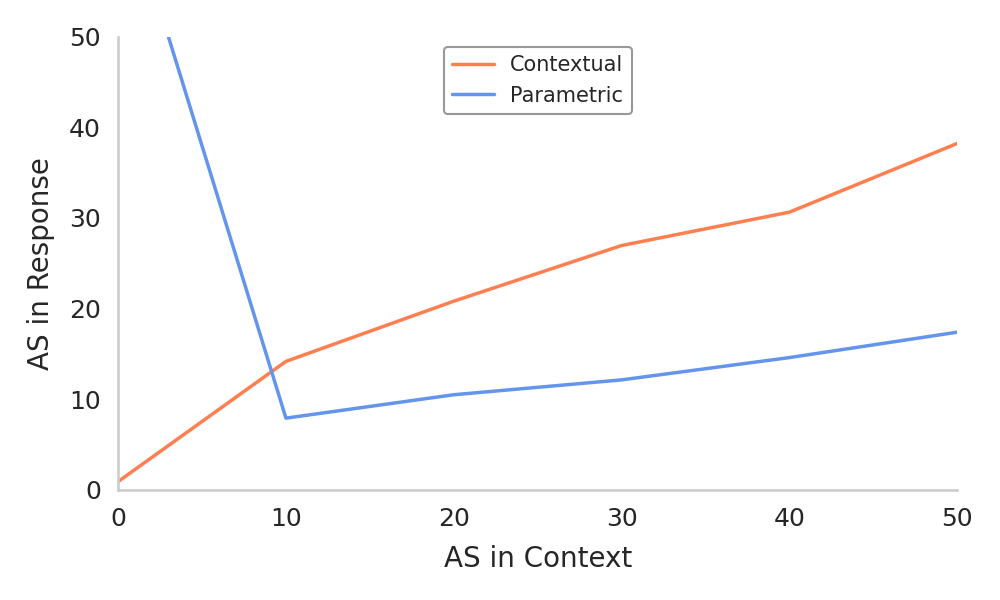}
    \caption{CK Prompt Instruction}
    \end{subfigure}
    \hfill
    \begin{subfigure}[t]{0.32\textwidth}
    \includegraphics[width=\textwidth]{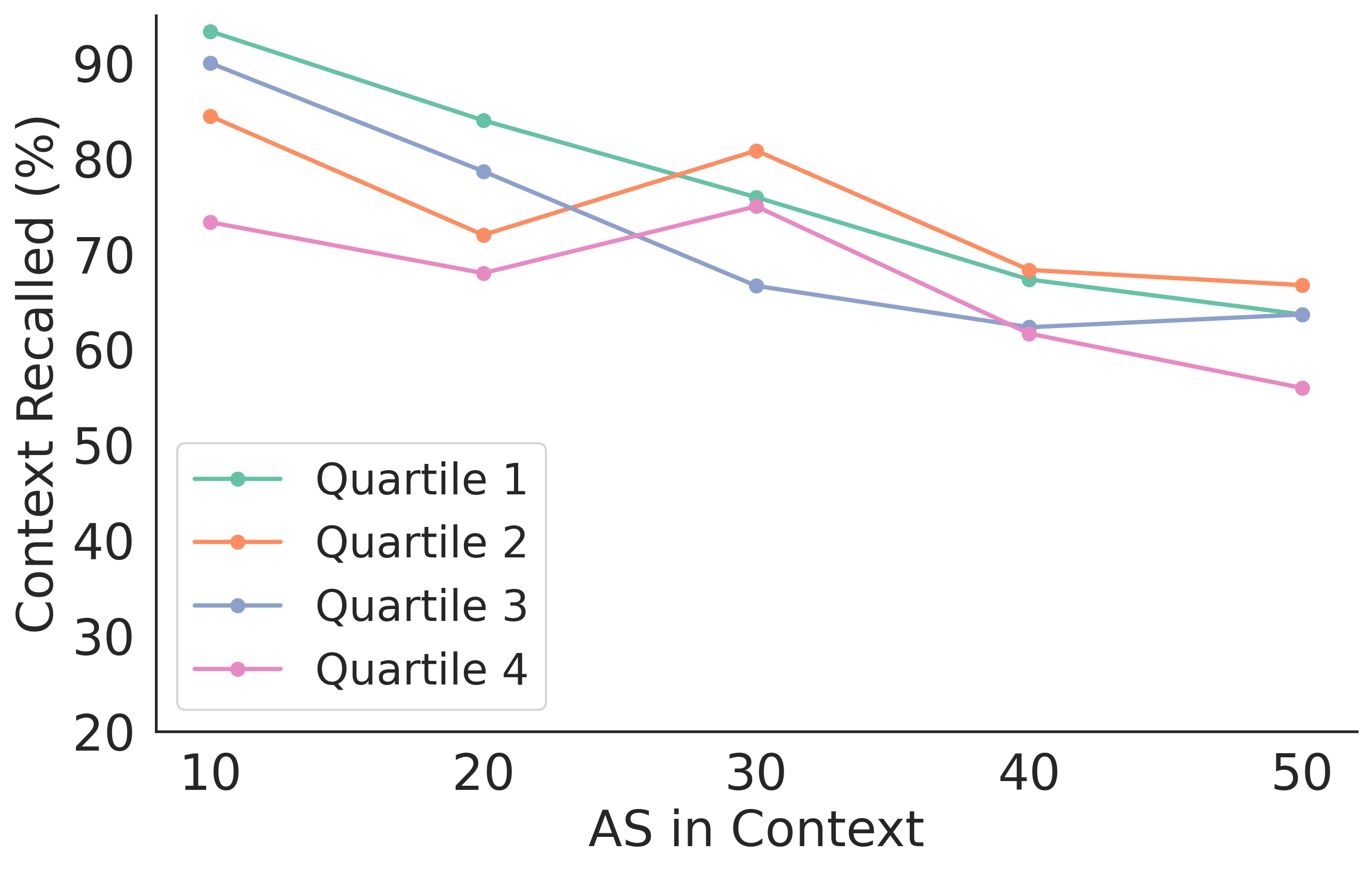}
    \caption{Strict Context Only Instruction}
    \end{subfigure}
    \hfill
    \begin{subfigure}[t]{0.32\textwidth}
    \includegraphics[width=\textwidth]{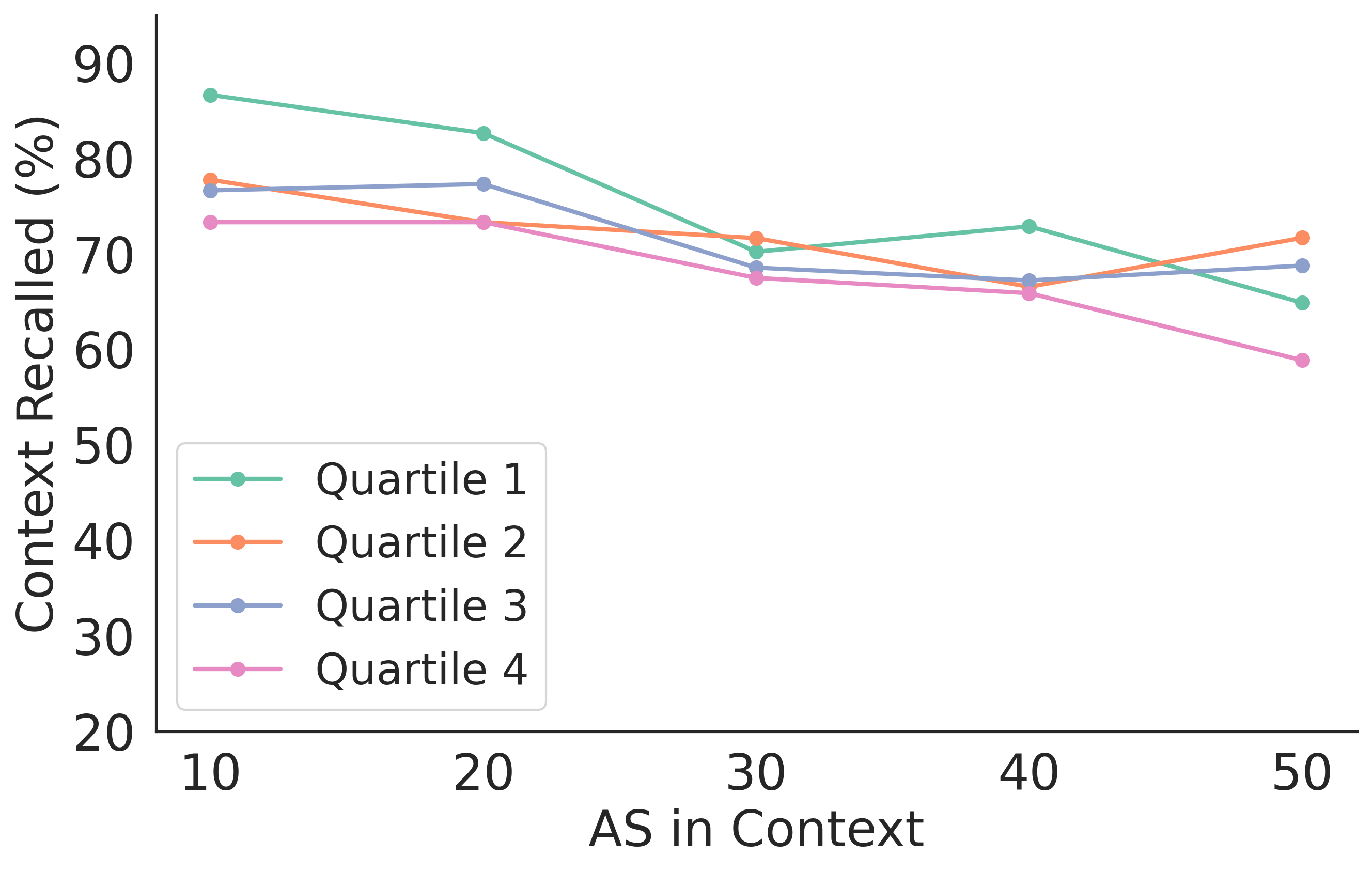}
    \caption{Balanced Use of Context Instruction}
    \end{subfigure}
    \hfill
    \begin{subfigure}[t]{0.32\textwidth}
    \includegraphics[width=\textwidth]{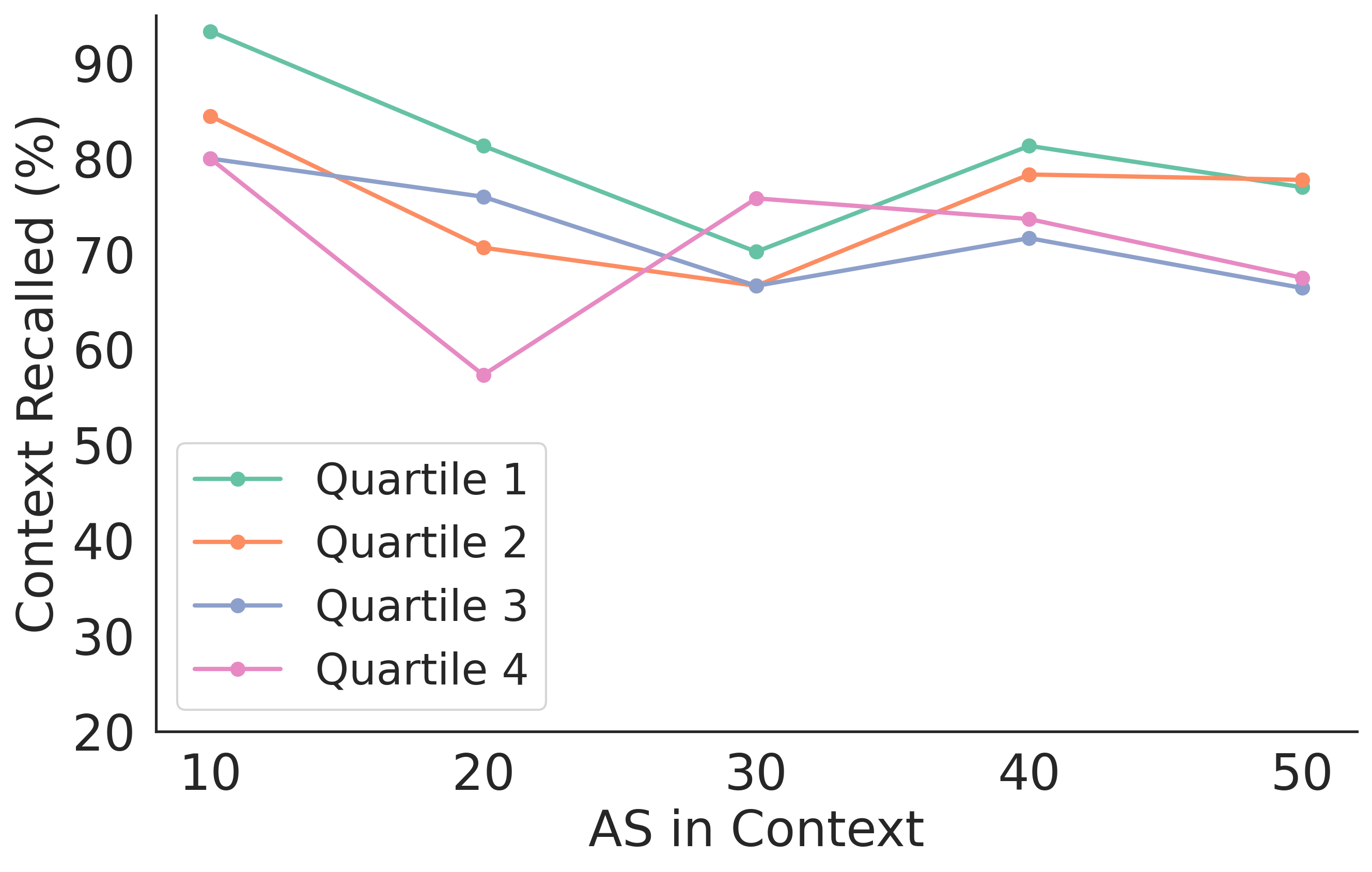}
    \caption{CK Prompt Instruction}
    \end{subfigure}
    \caption{Qwen 3 235B average number of CK/PK in responses (Top) and context recall (Bottom) across Different Instructions in Danish}
    \label{fig:different_instruction_qwen3235b_danish}
\end{figure*}

\section{Length Analysis between CoT and Non-CoT}
\label{app:cot_length}
We report response lengths under different prompting strategies to better understand CoT’s impact on output length. As shown in Table~\ref{tab:response_length_different_prompting}, CoT and CoT+CK prompts consistently produce shorter responses across languages and models, particularly in lower-resource settings.

\begin{table*}[ht]
\centering
\small
\begin{tabular}{|c|c|c|c|c|c|c|}
\hline
Language & Model & Avg Length & 0-25\% & 25-50\% & 50-75\% & 75-100\% \\
\hline
\multirow{6}{*}{English} 
 & \multirow{3}{*}{GPT-4o} 
 & Original & 117.4 & 198.6 & 251.0 & 319.8 \\ \cline{3-7}
 &  & CoT & 41.80 & 62.97 & 80.69 & 113.93 \\ \cline{3-7}
 &  & CoT + CK prompt  & 53.06 & 95.40 & 128.16 & 177.95 \\ \cline{2-7}
 & \multirow{3}{*}{LLaMA 3.2 90B} 
 & Original & 105.7 & 186.8 & 235.8 & 330.6 \\ \cline{3-7}
 &  & CoT & 31.30 & 44.23 &61.81 &108.40 \\ \cline{3-7}
 &  & CoT + CK prompt  & 38.46 &54.53 &79.78 &124.78 \\ \hline
 
\multirow{6}{*}{Spanish}
 & \multirow{3}{*}{GPT-4o} 
 & Original & 145.2 & 222.7 & 281.7 & 248.8 \\ \cline{3-7}
 &  & CoT & 51.64 &77.35 &96.03 &129.55 \\ \cline{3-7}
 &  & CoT + CK prompt  & 60.09 &99.88&133.25 &177.69 \\ \cline{2-7}
 & \multirow{3}{*}{LLaMA 3.2 90B} 
 & Original & 91.6 & 176.1 & 245.2 & 349.6 \\ \cline{3-7}
 &  & CoT & 32.62 &47.58 &63.44 &102.91 \\ \cline{3-7}
 &  & CoT + CK prompt  & 39.41 &55.41 &74.46 &112.62 \\ \hline
 
\multirow{6}{*}{Danish} 
 & \multirow{3}{*}{GPT-4o} 
 & Original & 102.2 & 171.9 & 219.3 & 272.1 \\ \cline{3-7}
 &  & CoT & 32.90 &51.46 &67.21 &98.32 \\ \cline{3-7}
 &  & CoT + CK prompt  & 48.75 &84.89 &113.34 &155.98 \\ \cline{2-7}
 & \multirow{3}{*}{LLaMA 3.2 90B} 
 & Original & 74.1 & 146.3 & 194.1 & 306.1 \\ \cline{3-7}
 &  & CoT & 21.82 &37.04 &51.58  &95.40 \\ \cline{3-7}
 &  & CoT + CK prompt  & 31.36 &47.72 &62.53  &97.78 \\ \hline
\end{tabular}
\caption{Response length with different prompting strategies across models and languages}
\label{tab:response_length_different_prompting}
\end{table*}

\section{Summarization Task Implementation Detail}
\label{app:summ_details}
Here we will go through the implementation details using CoPEval for the summarization task with DivSum dataset. The exact prompt we used to for \textbf{simple (vanilla)} summaries is:

\begin{quote}
\small
    \textit{The following is a collection of tweets about a topic. Summarize the main themes of the tweets in 5 sentences. Do not start your summarization with an introduction such as "Here is a summary in five sentences.}
    
    \textit{Only give your summary without preamble:  Topic: \texttt{[Topic]}}
    
    \textit{!IMPORTANT INSTRUCTIONS:  Remember, keep your summary to exactly five sentences.}
\end{quote}

The \textbf{CK Prompt} prompt is:
\begin{quote}
\small
    \textit{The following is a collection of tweets about a topic. Summarize the main themes of the tweets in 5 sentences.}
    
    \textit{Do not start your summarization with an introduction such as "Here is a summary in five sentences". Only give your summary without preamble.}
    
    \textit{Use only the provided information in the tweets to summarize the tweets. Avoid introducing any additional information not found in the tweets.}
    
    \textit{At the same time, focus on creating a summary that is balanced and draws fairly from relevant parts of the tweets. Your response must reflect balanced usage.}
    
    \textit{!IMPORTANT INSTRUCTIONS:  Remember, keep your summary to exactly five sentences.}
\end{quote}

To generate summaries for QMSum, The simple (vanilla) prompt we used was (from the dataset):

\begin{quote}
\small
    \textit{Summarize the whole meeting in 5 sentences. Do not start your summarization with an introduction such as "Here is a summary in five sentences.}
    
    \textit{Only give your summary without preamble.}
    
    \textit{!IMPORTANT INSTRUCTIONS:  Remember, keep your summary to exactly five sentences.}
\end{quote}

The \textbf{CK Prompt} prompt is:

\begin{quote}
\small
    \textit{The following is a transcript of a meeting. A specific query is provided regarding the meeting content. Summarize the main themes of the transcript in 5 sentences.}

    \textit{Do not begin your summary with an introduction such as "Here is a summary in five sentences." Only give your summary without preamble.}
    
    \textit{Use only the provided information in the meeting to summarize the query. Do not introduce any external information.}
    
    \textit{At the same time, focus on creating a summary that is balanced and draws fairly from relevant parts of the transcript. Your response must reflect balanced usage.}
    
    \textit{!IMPORTANT INSTRUCTIONS: Remember, keep your summary to exactly five sentences.}
\end{quote}

\end{document}